\documentclass[11pt]{article}

\pdfoutput=1

\usepackage[english]{babel}

\usepackage{amsfonts, amssymb, amsmath, amsthm, epsf,cite}

\usepackage[usenames,dvipsnames]{color}
\usepackage{color}
\usepackage{graphicx}
\usepackage{float}
\usepackage{booktabs}

\usepackage{fullpage}

\newtheorem{observation}{Observation}[section]

\newtheorem{interpretation}{Interpretation}[section]
\newtheorem{lemma}{Lemma}[section]

\newtheorem{condition}{Condition}[section]

\theoremstyle{definition}

\newtheorem{example}{Example}[section]

\numberwithin{equation}{section} \numberwithin{figure}{section}
\numberwithin{figure}{section}
\numberwithin{table}{section}

\renewcommand{\phi}{{\varphi}}

\newcommand{\cM}{{\mathcal M}}
\newcommand{\cN}{{\mathcal N}}

\newcommand{\cL}{{\mathcal L}}

\newcommand{\bH}{{\mathbf H}}
\newcommand{\bP}{{\mathbf P}}

\newcommand{\bbR}{{\mathbb R}}
\newcommand{\bbN}{{\mathbb N}}

\newcommand{\bbE}{{\mathbb E}}

\DeclareMathOperator*{\argmin}{arg\,min}

\let\phi=\varphi

\providecommand{\keywords}[1]{\textbf{\textit{Keywords.}} #1}

\title{Pairing an arbitrary regressor with an artificial neural network estimating aleatoric uncertainty}

\author{Pavel  Gurevich\thanks{Free University of Berlin, Berlin, Germany; RUDN University, Moscow, Russia; email: gurevich@math.fu-berlin.de},
Hannes Stuke\thanks{Free University of Berlin, Berlin, Germany; email: h.stuke@fu-berlin.de}\thanks{Equal contribution.}}

\begin{document}

\maketitle

\begin{abstract}
We suggest a general approach to quantification of different forms of aleatoric uncertainty in regression tasks performed by artificial neural networks. It is based on the simultaneous training of two neural networks with a joint loss function and a specific hyperparameter $\lambda>0$ that allows for automatically detecting noisy and clean regions in the input space and controlling their {\em relative contribution} to the loss and its gradients. After the model has been trained, one of the networks performs predictions and the other  quantifies the uncertainty of these predictions by estimating the locally averaged loss of the first one. Unlike in many classical uncertainty quantification methods, we do not assume any a priori knowledge of the ground truth  probability distribution, neither do we, in general, maximize the likelihood of a chosen parametric family of distributions.  We analyze the learning process and the influence of clean and noisy regions of the input space on the loss surface, depending on $\lambda$.  In particular, we show that small   values of $\lambda$ increase the relative contribution of clean regions to the loss and its gradients. This explains why choosing small $\lambda$  allows for better predictions compared with neural networks without uncertainty counterparts and those based on classical likelihood maximization. Finally, we demonstrate that one can naturally form ensembles of pairs of our networks and thus capture both aleatoric and epistemic uncertainty and avoid overfitting.
\end{abstract}


\tableofcontents

\section{Introduction}

Neural networks (NNs) are among the main tools that are used nowadays for solving regression and forecasting problems~\cite{Hussain,Gooijer,Goodfellow2016,Kuhn}.
One theoretical limitation of standard NNs with regression is that they generate  averaged predictions of the target variables, but do not provide information on how certain the predictions are. Obviously, quantification of this uncertainty is crucial from the viewpoint of the real world applications~\cite{Krzywinski13,GalThesis16}. One of the first approaches to learning uncertainty by NNs was the {delta method}~\cite{WildSeber,18Hwang, 19Veaux} originating from nonlinear regression theory. It assumes that the noise is input-independent (homoscedastic), while the prediction error  is proportional to the noise and to the gradient of the output of a NN with respect to the NN's weights. Nowadays, there are two major approaches to learning uncertainty by NNs, both having a probabilistic background. The first exploits Bayesian NNs allowing one to capture epistemic uncertainty (related to lack of data). The second, sometimes called the frequentist approach, treats the weights as deterministic parameters and rather captures aleatoric uncertainty by directly reconstructing probability distributions of the observed data.

In the Bayesian approach to NNs~\cite{MacKay,Hinton93,Neal95}, the weights are treated as random variables.  Based on the likelihood of the target variables, a prior distribution of the weights  and Bayes' theorem, one obtains a posterior distribution of the weights, and hence a predictive distribution for the data. However, in practice, the latter two are computationally intractable, especially for large data sets and large network architectures. Different ways to approximate them form a field of ongoing research~\cite{Neal95,Jylaenki14,Hinton93, Welling2011, Blundell2015, Kingma2015, GalGhahramani2015, HernandezLobato15, GalThesis16, LiTurner2016, HernandezLobatoLi2016, LiuWang2016, LiGal2017, Louizos2017}.  In~\cite{GalGhahramani2015},   a connection between NNs trained with dropout~\cite{Hinton12,Srivastava2014} and Gaussian processes~\cite{Rasmussen2006} was established. The latter are Bayesian, but generally ``non-NN'' regression methods. The typical choice of the likelihood in Bayesian NNs is Gaussian with an input-independent variance. Although this variance is often estimated during fitting, the input-dependence of the variance of the predictive distribution occurs only due to the variance of the weights.

In the frequentist approach, one directly approximates the ground truth input-dependent distribution of the target variables by a parametrically given predictive distribution. The parameters of the latter are approximated by NNs with deterministic weights. The typical choice of the approximating distribution is Gaussian~\cite{Nix}, explicitly yielding the mean and the variance of the data, but other options were also considered, e.g., Laplacian~\cite{KendallGal17}, Students' t~\cite{GurHannesGCP}, and mixture of Gaussians~\cite{Bishop94,BishopBook,ZenSenior}.

To explain both epistemic and aleatoric uncertainty at the same time, a combination of the Bayesian approach (using the dropout variational inference) with the mean-variance estimate was used in~\cite{KendallGal17}. A combination of the mean-variance estimate and ensembles of NNs was suggested in~\cite{Lakshminarayanan16,Lakshminarayanan17}.

We refer the reader to~\cite{Myshkov16, GalThesis16,Lakshminarayanan16,Lakshminarayanan17,GurHannesGCP} for further references and experimental comparison of the above methods.

In this paper, we propose an approach that generalizes the frequentist methods (in which one maximizes the likelihood of the data such as Gaussian or Laplace). However, unlike in the above methods, we do not require a priori knowledge about the probabilistic structure of the noise and do not necessarily maximize a likelihood of the data. Our approach quantifies aleatoric uncertainty by automatically estimating a locally averaged loss of the regression network (called the {\em regressor}) with the help of the second network (called the {\em uncertainty quantifier}). This quantification is applicable to any loss of the regressor and thus allows for estimating uncertainty exactly in terms of the objective one wants to minimize.  We will see that it not only quantifies how certain the predictions are, but also  allows for better predictions (compared with standard NNs and NNs based on classical likelihood maximization), especially in regions of the input space with relatively small noise in the target variables (clean regions). This is achieved due to simultaneously training the regressor and the uncertainty quantifier by minimizing a joint loss.

As we said, our approach does not use any explicit form of probability distribution of the data. However, once this assumption is done, one can give a natural interpretation for ensembles of pairs of our networks in terms of mixture distributions, similarly to~\cite{Lakshminarayanan16,Lakshminarayanan17}. This allows one to capture both aleatoric and epistemic uncertainty and avoid overfitting. Again, we will see that our ensembles typically outperform those in~\cite{Lakshminarayanan16,Lakshminarayanan17}.

The paper is organized as follows. In Sec.~\ref{secMainIdea}, we informally explain our main idea. In Sec.~\ref{subsecGeneralLoss}, we introduce our model and describe the joint loss function. In Sec.~\ref{subsecProbabil}, we give an explicit formula for estimating the expected regressor's loss, introduce ensembles, and discuss similarities and dissimilarities of our method to the classical likelihood maximization approach. In Sec.~\ref{secSigmoidSoftReProba}, we consider different functional forms of the joint loss in the cases where the output of the uncertainty quantifier is implemented as the sigmoid and softplus activation functions, respectively. In Sec.~\ref{secSigmoidSoftReLambda}, we analyze how the hyperparameters of the joint loss affect the overall learning speed and how the loss surface is influenced by clean and noisy regions. In particular, we will see that small values of the hyperparameter reduce the relative contribution of samples from noisy regions to the loss and explain why the usage of the joint loss may improve regressor's predictions.  In Sec.~\ref{secArtificial}, we illustrate our results with synthetic one-dimensional data. In Sec.~\ref{secBoston}, we apply our approach to publicly available data sets and compare it with other NN methods.
Section~\ref{secDiscussion} contains a conclusion. In Appendix~\ref{appendixHyperparameters}, we present the values of hyperparameters of different methods that are compared in Sec.~\ref{secBoston}.

\section{Main idea}\label{secMainIdea}

Suppose we have a standard neural network $\cN_r=\cN_r(x)$ for regression (the {\em regressor}) with a loss function $\cL_r$. We complement it by another neural network $\cN_q=\cN_q(x)$ (the {\em uncertainty quantifier}), and train both networks by minimizing a joint loss of the form
\begin{equation}\label{eqJLFIntro}
\cL_{\rm joint} = \cL_r(\cN_r) f (\cN_q) + \lambda g(\cN_q),
\end{equation}
where $f (z)$ and $g(z)$ are some fixed functions and $\lambda>0$ is a hyperparameter. The main assumption concerning the functions $f $ and $g$ is that the former is positive ($f(z)>0$) and increasing ($f'(z)>0$) and the latter is decreasing ($g'(z)<0$). We will see below that {\it small} values of $\cN_q$ correspond to uncertain predictions and {\it large} values to certain predictions. Intuitively, the more certain the quantifier $\cN_q$ is, the larger $f (\cN_q)$ is and hence the smaller the loss $\cL_r$ tends to be. On the other hand, the less certain the quantifier $\cN_q$ is, the smaller $f (\cN_q)$ is and hence the larger the loss $\cL_r$ can be. The second term $\lambda g(\cN_q)$ penalizes uncertain predictions. Thus, the regressor $\cN_r$ can ``afford'' to fit worse in uncertain noisy regions and use this freedom to fit better in certain clean regions. Moreover, while the regressor $\cN_r$ predicts the target value, it turns out (Interpretation~\ref{interGeneral}) that the quantifier~$\cN_q$ allows for estimating the expected regressor's loss via the formula
\begin{equation}\label{refLrf1f2Intro}
\text{expected regressor's loss} = -\dfrac{\lambda g'(\cN_q)}{f '(\cN_q)}.
\end{equation}
We emphasize that relation~\eqref{refLrf1f2Intro} need not involve any likelihood maximization and directly estimates any regressor's loss (see Sec.~\ref{secNotMaximization} for more details). However, e.g., in the case where the loss $\cL_r$ is chosen as the mean square error (MSE), this yields the empirical variance of the data for {\em any} choice of $f$, $g$, and~$\lambda$ and {\em any} (a priori unknown) ground truth distribution (Example~\ref{exProba1}).

\begin{figure}[!t]
	\begin{minipage}{0.5\textwidth}
       \includegraphics[width=\textwidth]{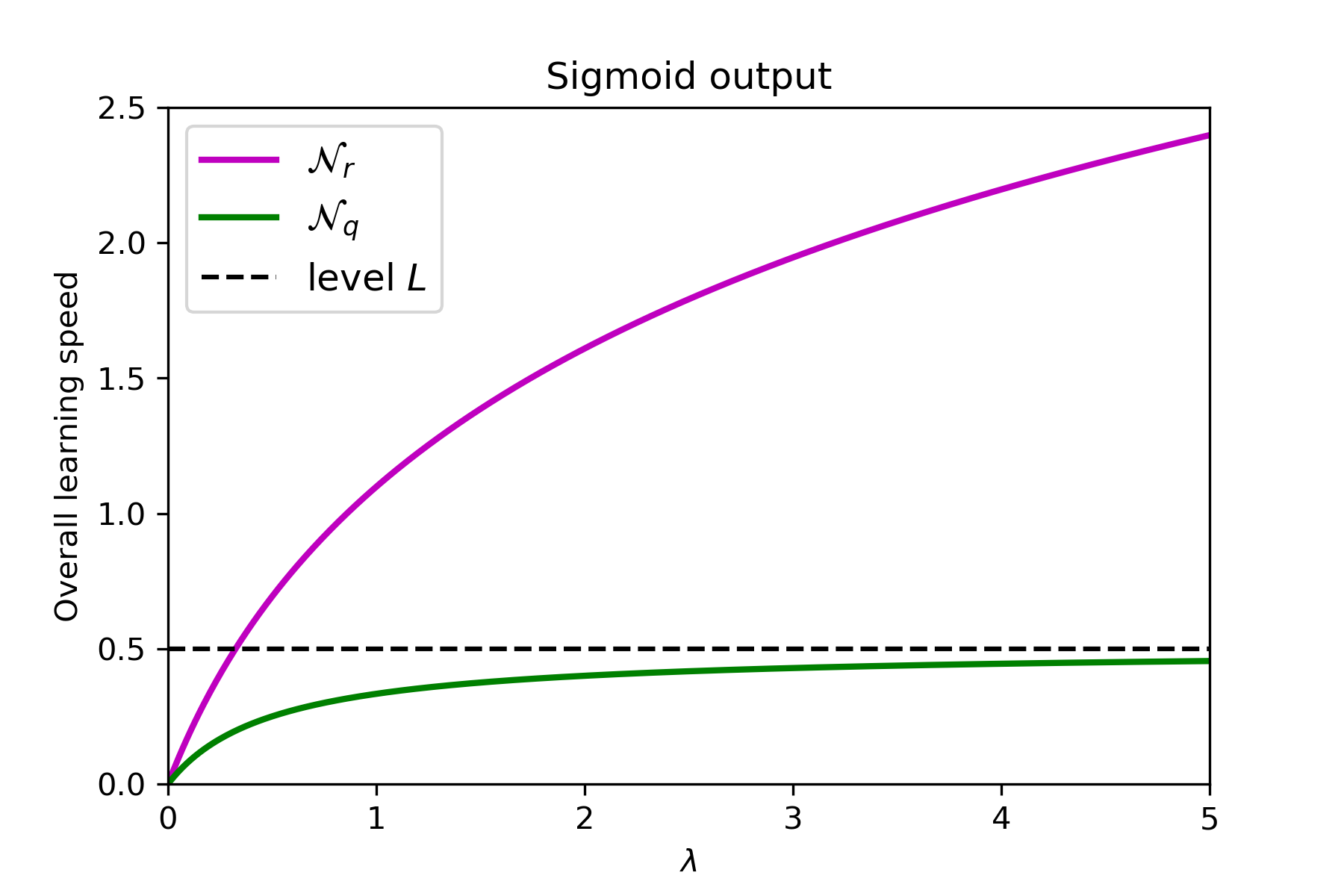}
	\end{minipage}
	\hfill
	\begin{minipage}{0.5\textwidth}
	   \includegraphics[width=\textwidth]{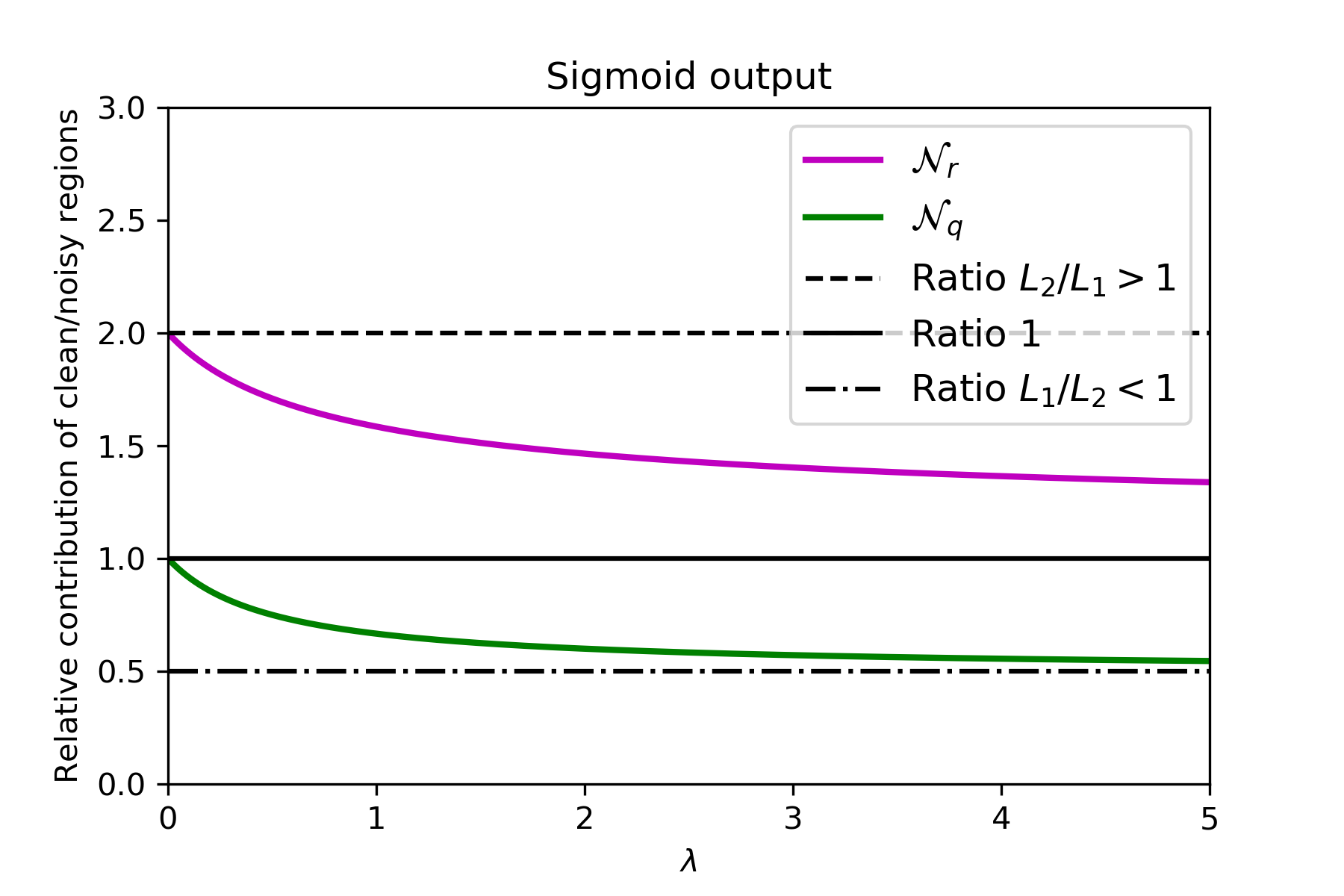}
	\end{minipage}
\caption{The overall learning speed and the relative contribution of clean and noisy regions to fitting the parameters of the regressor $\cN_r$ and the uncertainty quantifier $\cN_q$, depending on  $\lambda$ for the {\em sigmoid output} of $\cN_q$. Left: overall learning speed.
Right: Relative contribution of clean and noisy regions.
}\label{figTrainingSpeedSigmoid}

	\begin{minipage}{0.5\textwidth}
       \includegraphics[width=\textwidth]{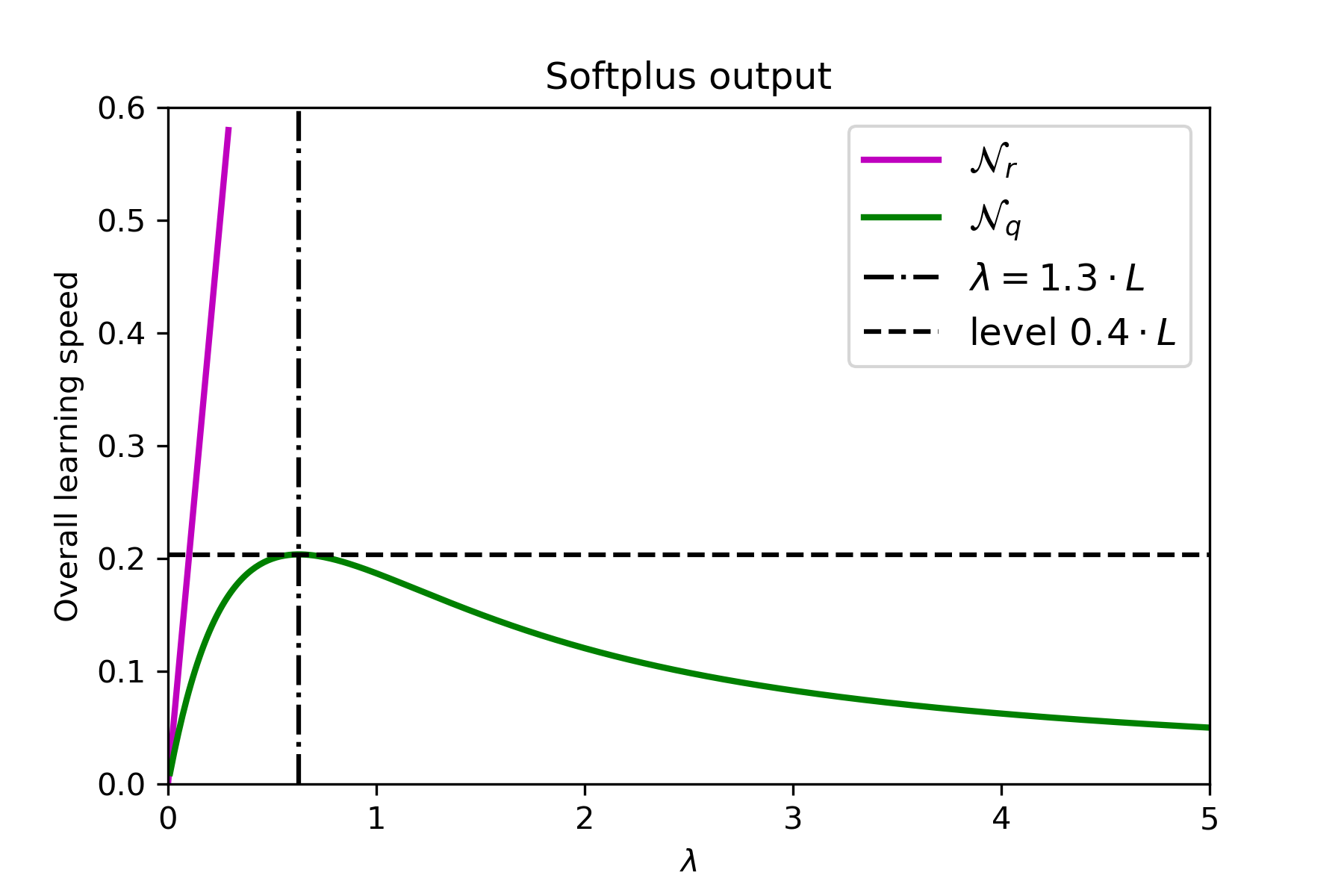}
	\end{minipage}
	\hfill
	\begin{minipage}{0.5\textwidth}
	   \includegraphics[width=\textwidth]{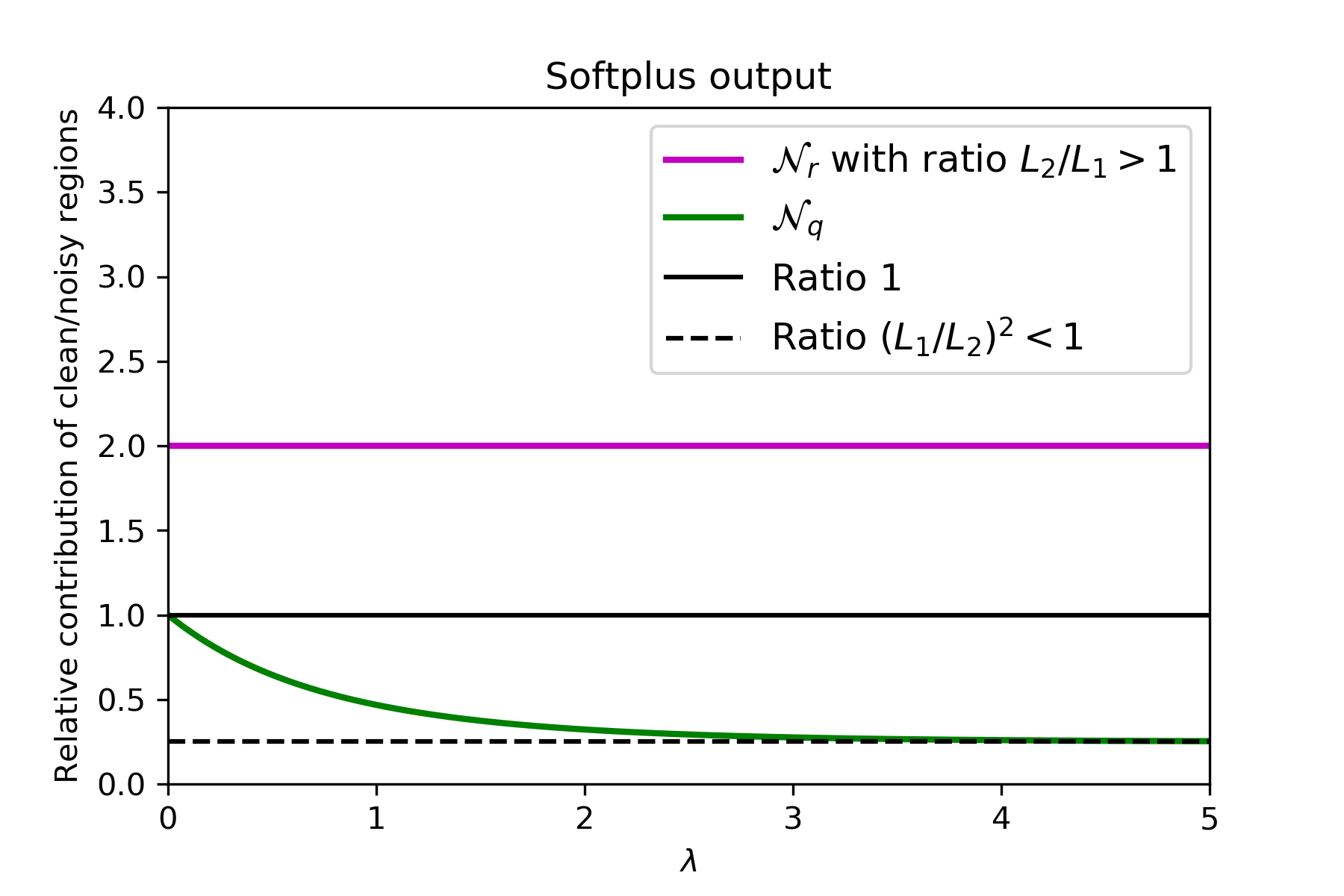}
	\end{minipage}
\caption{The overall learning speed and the relative contribution of clean and noisy regions to fitting the parameters of the regressor $\cN_r$ and the uncertainty quantifier $\cN_q$, depending on  $\lambda$ for the {\em softplus output} of $\cN_q$. Left: overall learning speed.
Right: Relative contribution of clean and noisy regions.
}\label{figTrainingSpeedSoftplus}
\end{figure}

In Sec.~\ref{secSigmoidSoftReLambda}, we explain that choosing small $\lambda$ may significantly facilitate regressor's ability to learn in clean regions (compared with standard NNs and NNs based on classical likelihood maximization). We perform a rigorous analysis that indicates how clean and noisy regions contribute to $\cL_{\rm joint}$ and its gradients, depending on $\lambda$, on the functions $f$ and $g$, and on the activation function of the output node of the quantifier $\cN_q$. In particular, we show that choosing small $\lambda$ decreases the relative contribution of samples from noisy regions to the loss. As a result, the minima of the original loss $\cL_{\rm joint}$ get close to the minima of the loss containing samples from clean regions {\em only}. Table~\ref{table} informally summarizes this influence for the sigmoid and softplus activation functions, while Fig.~\ref{figTrainingSpeedSigmoid} and~\ref{figTrainingSpeedSoftplus} illustrate the qualitative dependence of the overall learning speed and of the relative contribution of clean and noisy regions, depending on $\lambda$.
\begin{table}[h!]
  \centering
    \begin{tabular}{l|c|c||c|c}
    {} & \multicolumn{2}{|c||}{sigmoid} & \multicolumn{2}{|c}{softplus} \\

    \hline

    $\lambda$ & \text{regressor} & \text{quantifier} & \text{regressor} & \text{quantifier} \\

    \hline

    small & \text{clean} $>$ \text{noisy} & \text{clean} $=$ \text{noisy} & \text{clean} $>$ \text{noisy}& \text{clean} $=$ \text{noisy} \\

    \hline

    large & \text{clean} = \text{noisy} & \text{clean} $<$ \text{noisy} & \text{clean} $>$ \text{noisy}& \text{clean} $\ll$ \text{noisy}

  \end{tabular}
  \caption{Relative contribution of clean and noisy regions to fitting the parameters of the regressor and uncertainty quantifier in case of sigmoid and softplus activation functions in the output of the quantifier $\cN_q$. Notation ``$=$'' stands for  ``comparable'' contribution,  ``$<$'' and ``$>$'' for a ``lower'' and ``higher'' contribution, and  ``$\ll$'' for a ``much lower'' contribution.}\label{table}
\end{table}

\section{General model}\label{subsecGeneralLoss}

Let $X=\{x^1,\dots,x^N\}\subset\bbR^n$, $n\in\bbN$, be a (training) data set consisting of $N\in\bbN$ samples, and $Y=\{y^1,\dots,y^N\}\subset\bbR^m$, $m\in\bbN$,  the corresponding target set of observations. The model we propose consists of two NNs that are trained simultaneously. The  first network $\cN_r:\bbR^n\to \bbR^m$ is supposed to perform the regression task, while the     second network $\cN_q:\bbR^n\to \bbR$ is supposed to quantify the uncertainty associated with the predictions of the regression network. Thus, to each sample $x\in\bbR^n$, we assign the pair $\big(\cN_r(x),\cN_q(x)\big)$, where $\cN_r(x)$ is the prediction and $\cN_q(x)$ is its certainty. The networks $\cN_r$ and $\cN_q$ are parametrized by learnable weights~$\theta_r$ and $\theta_q$, respectively, that may be shared but need not be. We omit the dependence of the networks on these weights whenever it does not lead to confusion.  In what follows, we call~$\cN_r$ a {\em regressor} and $\cN_q$ an {\em uncertainty quantifier}. We will see that the smaller $\cN_q(x)$ is, the more uncertain the prediction of $\cN_r(x)$ is.

Let
$$
\cL_r(y^i,y_r^i),\quad y_r^i = \cN_r(x^i),
$$
be a loss of the regressor $\cN_r$.
This can be any loss function used in artificial NNs for regression. We only assume that $\cL_r$ takes positive values. Now we define a joint loss as follows (cf.~\eqref{eqJLFIntro}):
\begin{equation}\label{eqJointLoss}
 \cL_{\rm joint}=\dfrac{1}{N} \sum\limits_{i=1}^N \left( \cL_r(y^i,y_r^i)  f (z^i) + \lambda g(z^i)\right),\quad y_r^i=\cN_r(x^i),\ z^i = \cN_q(x^i),
\end{equation}
where $f ,g:I\mapsto \bbR$ are some fixed functions defined on an open interval $I\subset\bbR$ and $\lambda>0$ is a hyperparameter. We discuss their role in Secs.~\ref{secSigmoidSoftReProba} and~\ref{secSigmoidSoftReLambda}. We will see that the choice of the functions $f (z)$ and $g(z)$ depends on concrete implementations of the uncertainty quantifier~$\cN_q$, while $\lambda$ affects the overall learning speed and the ratio between the learning speeds in clean and noisy regions. To keep notations uncluttered, we do not indicate the arguments of $\cL_{\rm joint}$. Depending on the context, we will treat it as a function of $y_r^i,z^i$ ($i=1,\dots,N$) or $\theta_r,\theta_q$.

We assume throughout the following.
\begin{condition}\label{condPropf1f2}
\begin{enumerate}
  \item\label{condPropf1f2_1} $f(z)>0$, $f '(z)>0,\ g'(z)<0$ for all $z\in I$.
  \item\label{condPropf1f2_2} For any $L>0$, the function $L f (z) + \lambda g(z)$ of variable $z$ achieves its finite minimum on $I$.
\end{enumerate}
\end{condition}

Due to item~\ref{condPropf1f2_1},  the more certain the quantifier $z^i=\cN_q(x^i)$ is, the larger $f (z^i)$ is and hence the smaller the loss $\cL_r$ in~\eqref{eqJointLoss} tends to be. The  term $\lambda g(z^i)$ in~\eqref{eqJointLoss} penalizes uncertain predictions.

\section{Probabilistic aspects and generalizations to ensembles}\label{subsecProbabil}

In this section, we give a probabilistic interpretation of our approach. In Sec.~\ref{secSigmoidSoftReLambda}, we analyze the influence of clean and noisy regions on the loss function and its gradients. In both cases, it is convenient to use an equivalent representation of the loss function~\eqref{eqJointLoss}. We will group the points in $X$ in pairwise disjoint sets $X^1,\dots,X^J$ in such a way that $X=\bigcup\limits_{j=1}^J X^j$ and both $\cN_r(x)$ and $\cN_q(x)$ are almost constant on each $X^j$. We denote these constants by $y_r^j$ and $z^j$, respectively (which have slightly different meaning compared with~\eqref{eqJointLoss}). Denote by $Y^1,\dots,Y^J$ the corresponding subdivision of $Y$. Let $M_j$ be the number of points in $X^j$.
Then
\begin{equation}\label{eqJointLossSumProb}
\begin{aligned}
\cL_{\rm joint} &\approx \sum\limits_{j=1}^J \dfrac{M_j}{N} \left(\dfrac{1}{M_j}  \sum\limits_{y\in Y^j} \cL_r(y,y_r^j)f(z^j) + \lambda g(z^j)\right)\\
& =
\sum\limits_{j=1}^J \dfrac{M_j}{N} \left( \bbE_j[\cL_r(\cdot,y_r^j)] f(z^j) + \lambda g(z^j)\right),
\end{aligned}
\end{equation}
where $\bbE_j[\cL_r(y,y_r^j)]$ stands for the empirical mean of regressor's loss $\cL_r(y,y_r^j)$, $y\in Y^j$,  and we used the relation $M_1+\dots+M_J=N$. For the further theoretical analysis, we will use the right-hand side of~\eqref{eqJointLossSumProb} for $\cL_{\rm joint}$.

\subsection{Similarity to probabilistic models}

We fix the region $X^j$ for some $j$. The first simple observation is as follows.

\begin{observation}\label{oExpectedLossOnePoint}
\begin{enumerate}
\item\label{itemExpectedLoss1} Let a pair
\begin{equation}\label{eqOverlineYZ}
(\overline{y_r}, \overline{z})\in\bbR^m\times\bbR
\end{equation}
be such that $\overline{z}$ is a critical point of the joint loss function
$$
 \bbE_j [\cL_r(\cdot,\overline{y_r})] f (z) + \lambda g(z)
$$
with respect to $z$. Then
$$
 \bbE_j [\cL_r(\cdot,\overline{y_r})] = -\dfrac{\lambda g'(\overline{z})}{f '(\overline{z})}.
$$
\item If, additionally, the pair in~\eqref{eqOverlineYZ} is a critical point of the joint loss function
\begin{equation}\label{eqJointLossProb}
 \bbE_j [\cL_r(\cdot,y_r)] f (z) + \lambda g(z)
\end{equation}
with respect to $(y_r,z)$, then $\overline{y_r}$ is a critical point of $\bbE_j [\cL_r(\cdot,y_r)]$. Moreover, if $\bbE_j [\cL_r(\cdot,y_r)]$ is convex with respect to $y_r$, then $\overline{y_r}$ is its {\em global} minimum.
\end{enumerate}
\end{observation}
Note that, due to Condition~\ref{condPropf1f2}, a critical point $\overline z\in\bbR$ always exists and $f '(\overline z)\ne 0$. We emphasize that $(\overline y_r,\overline z)$ in Observation~\ref{oExpectedLossOnePoint} must be a {\em critical point}, but need not be a local minimum. However, in our concrete implementations in Secs.~\ref{secSigmoidSoftReProba} and~\ref{secSigmoidSoftReLambda}, it is a global minimum.

Observation~\ref{oExpectedLossOnePoint} together with representation of the joint loss function in~\eqref{eqJointLossSumProb} implies the following interpretation in terms of neural networks.
\begin{interpretation}\label{interGeneral}
Let $\cN_r(x)$ be a prediction of the regressor and $y$ the ground truth value. Then the uncertainty quantifier $z=\cN_q(x)$ provides
\begin{equation*}\label{eqExpectedLossR}
\text{expected loss $\cL_r(y,\cN_r(x))$} = -\dfrac{\lambda g'(z)}{f '(z)}.
\end{equation*}
\end{interpretation}

Now we illustrate the relation between our approach and learning probability distributions.

\begin{example}\label{exProba1}
\begin{enumerate}
  \item\label{exProba1-1}   Assume the observations $y$ are scalar sampled from a ground truth probability distribution $\bP(y|x)$. We define regressor's loss as
  \begin{equation}\label{eqLossMSE}
    \cL_r(y,y_r) = |y-y_r|^2.
  \end{equation}
  Assume that $(\overline{y_r}, \overline{z})$ in~\eqref{eqOverlineYZ} is a critical point of the joint loss~\eqref{eqJointLossProb}.
  Then, due to~\eqref{eqLossMSE} and Observation~\ref{oExpectedLossOnePoint},
  \begin{equation}\label{eqEVMean}
  \overline{y_r} = \argmin\limits_{\mu\in\bbR}\bbE_{y\sim \bP(y|x)}\left[|y-\mu|^2\right] =  \bbE_{y\sim \bP(y|x)}[y],
  \end{equation}
  \begin{equation}\label{eqEVVar}
    -\dfrac{\lambda g'(\overline{z})}{f '(\overline{z})} = \bbE_{y\sim \bP(y|x)}\left[|y-\overline{y_r}|^2\right] = {\rm Var}_{y\sim \bP(y|x)}[y].
  \end{equation}
  Thus, the regressor $\cN_r(x)$ learns the expectation of $y$ due to~\eqref{eqEVMean}, and the uncertainty quantifier $\cN_q(x)$ yields the variance of $y$ according to~\eqref{eqEVVar}. We emphasize that we do {\em not} need to know the exact form of the ground truth distribution $\bP(y|x)$ in order to reconstruct its mean and variance.

\item\label{exProba1-2}
  Additionally to the special choice~\eqref{eqLossMSE} of regressor's loss, we consider a particular choice of the functions $f,g$ and the hyperparameter $\lambda$:
  \begin{equation}\label{eqFGLambda1}
  f(z)=z,\quad g(z) = -\ln z, \quad \lambda=1.
  \end{equation}
  Then our method becomes equivalent to maximization of the log-likelihood of an approximating Gaussian distribution
  \begin{equation}\label{eqGaussianDistribution}
    \bP_{\rm approx}(y|x,\mu,\tau) = \sqrt{\frac{\tau}{2\pi}}e^{-|y-\mu|\tau/2}.
  \end{equation}
  In particular,~\eqref{eqEVMean} and~\eqref{eqEVVar} take the form
  $
  \overline{y_r} = \mu$ and $-\dfrac{\lambda g'(\overline{z})}{f '(\overline{z})}=\overline{z} = \tau$. However, we will see in Secs.~\ref{secArtificial} and~\ref{secBoston} that the choice $\lambda=1$ is not optimal from the practical viewpoint.
\end{enumerate}
\end{example}

\begin{example}\label{exLaplaceDistr}
\begin{enumerate}
  \item
  Similarly to Example~\ref{exProba1} (item~\ref{exProba1-1}) it is easy to see that if $\cL_r(y,y_r)=|y-y_r|$, then
    \begin{equation}\label{eqEVMeanLaplace}
  \overline{y_r} =  \bbE_{y\sim \bP(y|x)}[y],\qquad
    -\dfrac{\lambda g'(\overline{z})}{f '(\overline{z})} = \bbE_{y\sim \bP(y|x)}\left[|y-\overline{y_r}|\right]
  \end{equation}
  for {\em any} ground truth distribution $\bP(y|x)$.
  \item For the particular choice~\eqref{eqFGLambda1} of the functions $f,g$ and the hyperparameter $\lambda$, our method becomes equivalent to maximization of the log-likelihood of the approximating Laplace distribution
      \begin{equation}\label{eqLaplaceDistribution}
      \bP_{\rm approx}(y|x,\mu,\tau)=\frac{\tau}{2}e^{-|y-\mu|\tau}
      \end{equation}
       and yields $\overline{y_r}=\mu$ and $-\dfrac{\lambda g'(\overline{z})}{f '(\overline{z})}=\overline{z}=\tau$.
  \end{enumerate}
\end{example}

\begin{figure}[!t]
\centering
	\begin{minipage}{0.9\textwidth}
       \includegraphics[width=\textwidth]{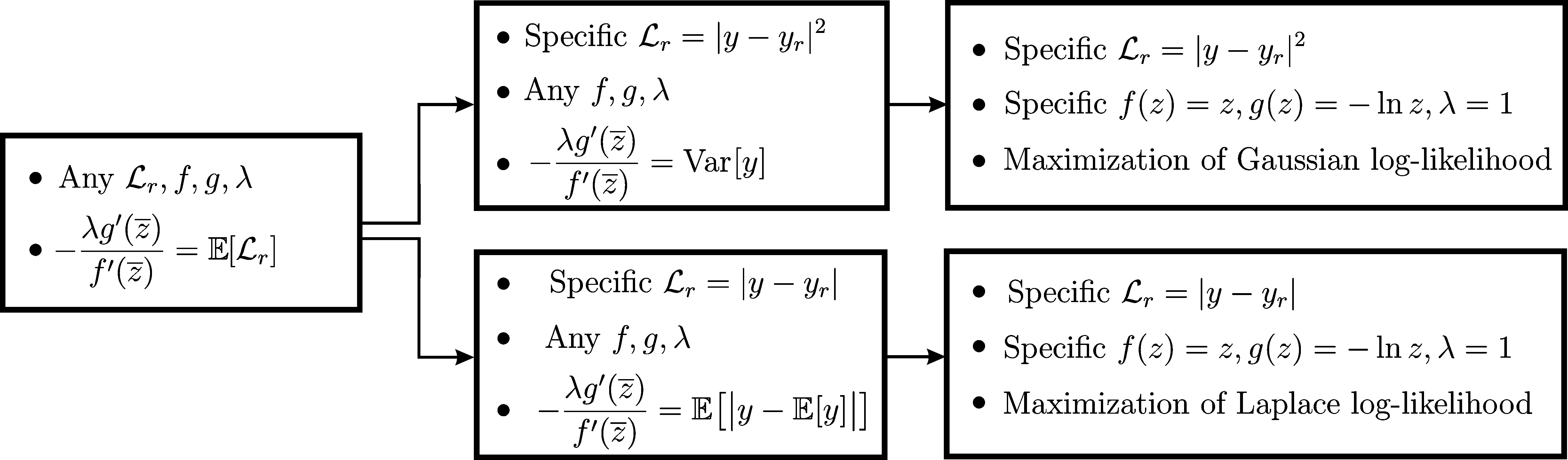}
	\end{minipage}
\caption{A flowchart illustrating particular cases of our method. Expectation and variance are taken with respect to (a priori unknown) ground truth distribution $\bP(y|x)$.}\label{figFlowchart}
\end{figure}

Examples~\ref{exProba1} and~\ref{exLaplaceDistr} are illustrated with a flowchart in Fig.~\ref{figFlowchart}.

\subsection{Ensembles}

Similarly to~\cite{Lakshminarayanan16,Lakshminarayanan17}, one can fit an ensemble of $K$ pairs of our regressor-quantifier networks and treat their predictions as components of a mixture model. Suppose we are interested in predicting means and variance of observed data. We assume that the data is sampled from an unknown ground truth distribution $\bP(y|x,\mu,V)$ with unknown mean $\mu$ and variance $V$. Each individual pair of networks estimates the values $(\mu_j,V_j)$, $j=1,\dots,K$. If we interpret them as admissible values for the mean and variance, respectively, we can marginalize over them and, similarly to~\cite{Lakshminarayanan16,Lakshminarayanan17}, obtain the predictive (approximating) distribution
$$
\bP_{\rm approx}(y|x)=\frac{1}{K}\sum\limits_{j=1}^K \bP(y|x,\mu_j,V_j)
$$
of the ensemble. One can easily check that the mean and the variance of $\bP_{\rm approx}(y|x)$ are
\begin{equation}\label{eqMeanVarianceEnsemble}
\mu=\frac{1}{K}\sum\limits_{j=1}^K \mu_j,\qquad V=\frac{1}{K}\sum\limits_{j=1}^K \left(V_j + (\mu_j-\mu)^2\right),
\end{equation}
{\em independently} of the concrete form of the unknown ground truth distribution $\bP(y|x,\mu,V)$.

As another example, suppose we want to predict means and expected absolute errors (EAE) of the observed data. To give a probabilistic interpretation to the ensemble in this case, we assume that the data is sampled from a ground truth Laplace distribution $\bP(y|x,\mu,\tau)$ given by the right-hand side in~\eqref{eqLaplaceDistribution} with unknown $\mu$ and $\tau$. Due to Example~\ref{exLaplaceDistr}, each individual pair of networks (for any choice of $f,g,\lambda$) estimates the values $(\mu_j,\tau_j)$, which can be interpreted as admissible values for the parameters of the Laplace distribution~\eqref{eqLaplaceDistribution}. Marginalizing over them, we obtain the
predictive (approximating) distribution
$$
\bP_{\rm approx}(y|x)=\frac{1}{K}\sum\limits_{j=1}^K \bP(y|x,\mu_j,\tau_j).
$$
One can easily check that, for this distribution,
\begin{equation}\label{eqMeanEAEEnsemble}
\mu=\bbE[y]=\frac{1}{K}\sum\limits_{j=1}^K \mu_j,\qquad \bbE[|y-\mu|]=\frac{1}{K}\sum\limits_{j=1}^K \left(|\mu-\mu_j| + \frac{1}{\tau_j}e^{-|\mu-\mu_j|\tau_j}\right).
\end{equation}

\subsection{Dissimilarity to the classical likelihood maximization approach}\label{secNotMaximization}

Let us explain in detail why our approach is different from the classical likelihood maximization. For each fixed $z$, one can consider the (approximating) probability distribution
\begin{equation}\label{eqP}
  \bP_{\rm approx}(y|x,y_r,z) = \frac{e^{-\left[\cL_r(y,y_r)f(z) + \lambda g(z)\right]}}{Z(z)}
\end{equation}
associated with regressor's loss $\cL_r(y,y_r)$, cf.~\cite{Tishby89}. Assuming that regressor's loss depends only on $y-y_r$, the normalization constant
\begin{equation}\label{eqZ}
Z(z) = e^{-\lambda g(z)}\int_{-\infty}^{\infty} e^{-\cL_r(y,y_r)f(z)}dy
\end{equation}
does not depend on $y_r$, but depends on $z$. For a fixed $z$, the minimization of the joint loss $\cL_r(y,y_r)f(z) + \lambda g(z)$ with respect to $y_r$ is obviously equivalent to the maximization of the log-likelihood of $\bP_{\rm approx}(y|x,y_r,z)$ with respect to $y_r$, in which case neither the functions $f(z),g(z)$, nor the normalization constant $Z(z)$ play any role. However, we minimize the joint loss with respect to {\em both} $y_r$ and $z$  in our approach. One could try  to maximize the log-likelihood of $\bP_{\rm approx}(y|x,y_r,z)$ with respect to both $y_r$ and $z$ as well. This is also a valid approach, but then two issues would arise.

First, the normalization constant $Z(z)$ is in general not explicitly given or the integral in~\eqref{eqZ} may even diverge, which makes it impossible in practice  to maximize the log-likelihood, nor to find expected regressor's loss\footnote{In this formula, we assume that we managed to maximize the log-likelihood and that the found approximating distribution $\bP_{\rm approx}(y|x,y_r,z)$ coincides with the ground truth distribution. In general, the family of distributions $\bP_{\rm approx}(y|x,y_r,z)$ may even be not rich enough to contain $\bP(y|x)$. Then formula~\eqref{eqExpectedLossML} would provide only an approximation for $\bbE_{y\sim \bP(y|x)}[\cL_r(y,y_r)]$.}
\begin{equation}\label{eqExpectedLossML}
\bbE_{y\sim \bP_{\rm approx}(y|x,y_r,z)}[\cL_r(y,y_r)] = \int_{-\infty}^{\infty} \cL_r(y,y_r) \bP_{\rm approx}(y|x,y_r,z) \,dy.
\end{equation}
On the other hand, the minimization of the joint loss $\cL_r(y,y_r)f(z) + \lambda g(z)$ is still straightforward.

Second, even if $Z(z)$ can be explicitly calculated, it follows from~\eqref{eqP} and~\eqref{eqZ} that $\bP_{\rm approx}(y|x,y_r,z)$ is actually independent of $\lambda$ and $g(z)$. Hence, the value $z_{\rm max}$ for which the maximum of log-likelihood is achieved  is also independent of $\lambda$ and of the choice of the function $g$. This is obviously not the case for the minimization of $\cL_r(y,y_r)f(z) + \lambda g(z)$. The value $z_{\rm min}(\lambda)$ that minimizes the joint loss depends on $\lambda$ and $f,g$. This value $z_{\rm min}(\lambda)$ need {\em not} maximize the log-likelihood of $\bP_{\rm approx}(y|x,y_r,z)$, but still allows one to estimate the expected loss according to Interpretation~\ref{interGeneral}. Thus, the learning process in our case is completely different. Moreover, we will see in Sec.~\ref{secSigmoidSoftReLambda} that the freedom to choose $\lambda$ together with the fact that $z_{\rm min}(\lambda)$ and hence $f(z_{\rm min}(\lambda))$ depend on $\lambda$ allow one to vary the relative contribution of clean and noisy regions to the joint loss and to effectively downweight samples from noisy regions.


\section{Sigmoid and softplus activation functions in the output of $\cN_q$}\label{secSigmoidSoftReProba}

In this section, we   discuss different choices of the functions $f$ and $g$ for   concrete activation functions and rewrite the general Interpretation~\ref{interGeneral} accordingly.

\subsection{Sigmoid output of $\cN_q$}\label{subsecSigmoidProba}

Assume the output $z$ of the uncertainty quantifier $\cN_q$ is implemented as the sigmoid activation function
\begin{equation}\label{eqZSigmoid}
z = Z(\xi) = \dfrac{1}{1+e^{-\xi}},\quad \xi = (\text{output of the last hidden layer})\cdot w,
\end{equation}
where $w$ is a column of weights connecting the last hidden layer with the sigmoid activation function. We define the functions $f $ and $g$ as follows:
\begin{equation}\label{eqf1f2Sigmoid}
f (z) = - \ln(1-z),\quad g(z) = -  \ln z,\quad z\in I=(0,1).
\end{equation}
It is easy to see that Condition~\ref{condPropf1f2} is fulfilled.
The general Interpretation~\ref{interGeneral} takes the following form.

\begin{interpretation}\label{interSigmoid}
Let $\cN_r(x)$ be a prediction of the regressor and $y$ the ground truth value. Then the uncertainty quantifier $z=\cN_q(x)$ provides
$$
\text{expected loss $\cL_r(y,\cN_r(x))$} =   \lambda\left(\dfrac{1}{z}-1\right).
$$
\end{interpretation}

\subsection{Softplus output of $\cN_q$}\label{subsecSoftplusProba}

Assume the output $z$ of the uncertainty quantifier $\cN_q$ is implemented as the softplus nonlinearity
\begin{equation}\label{eqZSoftplus}
z = Z(\xi) = \ln(1+e^\xi),\quad \xi = (\text{output of the last hidden layer})\cdot w,
\end{equation}
where $w$ has the same meaning as in~\eqref{eqZSigmoid}. We define the functions $f $ and $g$ as follows:
\begin{equation}\label{eqf1f2Softplus}
f (z) = z,\quad g(z) = -  \ln z,\quad z\in I=(0,\infty),
\end{equation}
Condition~\ref{condPropf1f2} is again fulfilled. The general Interpretation~\ref{interGeneral} takes the following form.

\begin{interpretation}\label{interSoftplus}
Let $\cN_r(x)$ be a prediction of the regressor and $y$ the ground truth value. Then the uncertainty quantifier $z=\cN_q(x)$ provides
$$
\text{expected loss $\cL_r(y,\cN_r(x))$} = \dfrac{\lambda}{z}.
$$
\end{interpretation}

\section{Contribution of clean and noisy regions to the joint loss, depending on  $\lambda$}\label{secSigmoidSoftReLambda}

In this section, we clarify the role of the hyperparameter $\lambda$. We study in detail the case of the sigmoid activation function in the output of $\cN_q$. The analysis of the softplus output is analogous, and hence we will formulate only  final conclusions.

We distinguish {\em clean} and {\em noisy} regions in the data set $X$. At each learning stage, we say that a region is {\em clean} if the loss of the regressor $\cN_r(x)$ is small for $x$ in this region. Otherwise, we call a region {\em noisy}. The notions clean and noisy are understood relative to each other. It turns out that $\lambda$ affects  to what extent the samples from clean and noisy regions contribute to the loss. The influence is illustrated in Fig.~\ref{figTrainingSpeedSigmoid} and \ref{figTrainingSpeedSoftplus} and in Table~\ref{table}, which we justify  below.

We discuss two different (but closely related) aspects. First, we analyze the contribution of different regions to the gradients of the loss (see Sec.~\ref{subsecOverallLearningSigmoid} and~\ref{subsecRelativeLearningSigmoid} for the sigmoid output and Sec.~\ref{subsecOverallLearningSoftplus} and~\ref{subsecRelativeLearningSoftplus} for the softplus output). Second, we analyze the influence of different regions on the loss surface itself and its minima (see  Sec.~\ref{subsubsecImportanceSigmoid} for the sigmoid output and Sec.~\ref{subsubsecImportanceSoftplus} for the softplus output).

In both cases, we are interested in the structure of the joint loss $\cL_{\rm joint}=\cL_{\rm joint}(\theta_r,\theta_q)$ in the $\theta_r$-space of the  regressor's weights and in the $\theta_q$-space of the uncertainty quantifier's weights. Due to~\eqref{eqJointLossSumProb}, we can write the joint loss as
\begin{equation}\label{eqJointLossM}
\cL_{\rm joint} \approx \sum\limits_{j=1}^J C_j \cM(L_j,\xi^j),
\end{equation}
where
\begin{equation}\label{eqMSigmoid}
\cM(L,\xi) = L f (Z(\xi))+ \lambda g(Z(\xi)), \qquad C_j = \dfrac{M_j}{N},
\end{equation}
$L_j=\bbE_j [\cL_r(\cdot,y_r^j)]$, the function $Z(\xi)$ is given either by~\eqref{eqZSigmoid} or~\eqref{eqZSoftplus},  $\xi^j = (\text{output of the last hidden layer})\cdot w$, the ``output of the last hidden layer'' is determined by the inputs from $X^j$ and all the weights of $\cN_q$ except for the last layer, and $w$ is a column of weights connecting the last hidden layer with the corresponding activation function of the output of $\cN_q$; finally, $f$ and $g$ are given by~\eqref{eqf1f2Sigmoid}.

In this section, we assume
that the weights $\theta_q$ of the uncertainty quantifier satisfy the following.

\begin{condition}\label{condThetaQ}
The weights $\theta_q$ of the uncertainty quantifier are in a neighborhood of the global minimum $\overline\theta_q $ of $\cL_{\rm joint}$ as a function of $\theta_q$. Moreover, this global minimum corresponds to the values $\overline{\xi^j}=\xi^j(\overline\theta_q)$ that minimize $\cM(L_j,\xi^j)$ for all $j=1,\dots,J$.
\end{condition}

Since the gradient of $\cL_{\rm joint}$ equals the sum of the gradients of the individual terms in~\eqref{eqJointLossM}, we will concentrate on these individual terms, corresponding to different regions $X_j$.

\subsection{Sigmoid output of $\cN_q$}\label{subsecSigmoidLambda}
In this subsection, we assume that the output $z$ of the uncertainty quantifier $\cN_q$ is implemented as the sigmoid activation function, see Sec.~\ref{subsecSigmoidProba}. We will justify the column ``sigmoid'' in Table~\ref{table}.

We begin with the following simple lemma, which shows in particular that, for any $L>0$, the function $\cM(L,\xi)$ has a unique global minimum with respect to $\xi$.

\begin{lemma}\label{lPropertiesMSigmoid}
  \begin{enumerate}
    \item\label{lPropertiesMSigmoid1}
    The function $\cM(L,\xi)$ in~\eqref{eqMSigmoid} is convex with respect to $\xi$. For each $L>0$, it achieves a global minimum with respect to $\xi$ at the point
    $
    \overline\xi =\overline\xi(L) = \ln({\lambda}/{L}).
    $

    \item\label{lPropertiesMSigmoid2}
    $
    \dfrac{\partial^2 \cM(L,\xi)}{\partial\xi^2}\Big|_{\xi=\overline\xi}=\dfrac{L\lambda}{L+\lambda}.
    $
    \item\label{lPropertiesMSigmoid3}  $f (Z(\overline\xi))= \ln\left(1+ {\lambda}/{L}\right)$.
  \end{enumerate}
\end{lemma}
\proof
Due to~\eqref{eqMSigmoid}, \eqref{eqZSigmoid}, and~\eqref{eqf1f2Sigmoid},
$
\cM(L,\xi) = -L \ln\left(\dfrac{e^{-\xi}}{1+e^{-\xi}}\right) - \lambda \ln\left(\dfrac{1}{1+e^{-\xi}}\right).
$
Hence, the result follows from the formulas
    $$
    \dfrac{\partial \cM(L,\xi)}{\partial\xi}=\dfrac{L-\lambda e^{-\xi}}{1+e^{-\xi}},\qquad
    \dfrac{\partial^2 \cM(L,\xi)}{\partial\xi^2}=(L+\lambda)\dfrac{e^{-\xi}}{(1+e^{-\xi})^2}.
    $$
\endproof

\subsubsection{Impact of $\lambda$ on the overall learning speed}\label{subsecOverallLearningSigmoid}

To analyze the overall learning speed of $\cN_r$, we check how the gradients of $\cM(L,\xi)$ (where $L=L(\theta_r)$ and $\xi=\xi(\theta_q)$) with respect to $\theta_r$ and $\theta_q$ depend on $\lambda$.

{\bf Regressor.} Due to Condition~\ref{condThetaQ},
\begin{equation}\label{eqf1SigmoidNabla}
\nabla_{\theta_r}\cM(L,\xi) = \nabla_{\theta_r} L(\theta_r)\cdot f(Z(\xi))\approx \nabla_{\theta_r} L(\theta_r) \cdot f(Z(\overline\xi)).
\end{equation}
Hence, by Lemma~\ref{lPropertiesMSigmoid}, item~\ref{lPropertiesMSigmoid3},
\begin{equation}\label{eqf1Sigmoid}
\left|\nabla_{\theta_r}\cM(L,\xi)\right|   \propto \ln\left(1+\dfrac{\lambda}{L}\right)\quad \text{(the magenta line in Fig.~\ref{figTrainingSpeedSigmoid}, left),}
\end{equation}
where $\propto$ stands for ``approximately proportional''.
Thus, for a fixed $L$, the smaller $\lambda$ is, the closer to $0$ the learning speed of $\cN_r$ is (asymptotically proportionally to $\lambda/L$). On the other hand, the larger $\lambda$ is, the larger the learning speed of $\cN_r$ is.

{\bf Uncertainty quantifier.} By Condition~\ref{condThetaQ},
\begin{equation}\label{eqHessian}
\nabla_{\theta_q}\cM(L,\xi(\theta_q)) \approx \bH(L,\overline\theta_q )\cdot (\theta_q - \overline\theta_q ) =
\dfrac{\partial^2 \cM(L,\overline\xi)}{\partial\xi^2} (\nabla_{\theta_q}\xi)\cdot (\nabla_{\theta_q}\xi)^T \cdot (\theta_q-\overline\theta_q ),
\end{equation}
where $\bH(L,\overline\theta_q)$ is the Hessian of $\cM(L,\xi(\theta_q))$ with respect to $\theta_q$ evaluated at $\overline\theta_q$, and $\overline\xi$ is defined in Lemma~\ref{lPropertiesMSigmoid}.
%
%
%
%
%
Therefore, by  Lemma~\ref{lPropertiesMSigmoid}, item~\ref{lPropertiesMSigmoid2},
\begin{equation}\label{eqHessianSigmoid}
\left|\nabla_{\theta_q}\cM(L,\xi(\theta_q))\right| \propto \dfrac{L\lambda}{L+\lambda}\quad \text{(the green line in Fig.~\ref{figTrainingSpeedSigmoid}, left)}.
\end{equation}
Hence, for a fixed $L$, the smaller $\lambda$ is, the closer to $0$ the learning speed of $\cN_q$ is (asymptotically proportionally to $\lambda$, independently of $L$). On the other hand, the larger $\lambda$ is, the closer to $L$ the learning speed of $\cN_q$ is.



\subsubsection{Relative contribution of clean and noisy regions to the gradients of the loss, depending on $\lambda$}\label{subsecRelativeLearningSigmoid}

Now we analyze the  {\em relative} amplitudes of the gradients in clean and noisy regions. Their ratio determines which regions are downweighted or upweighted, respectively, during training. 
Consider two regions $X^{j_1}$ and $X^{j_2}$ with the corresponding values of regressor's loss~$L_1$ and~$L_2$. Assume that $X^{j_1}$ is  a clean region and $X^{j_2}$ is a noisy region in the sense that
$$
L_2\gg L_1.
$$

{\bf Regressor.} Due to~\eqref{eqJointLossM}  and~\eqref{eqf1Sigmoid}, the relative contribution of the clean and noisy regions $X^{j_1}$ and $X^{j_2}$ to $\nabla_{\theta_r}\cL_{\rm joint}$ is determined by the value
\begin{equation}\label{eqf1RatioSigmoid}
R_{\rm sigmoid}(\lambda)=\dfrac{\ln\left(1+\dfrac{\lambda}{L_1}\right)}{\ln\left(1+\dfrac{\lambda}{L_2}\right)}\quad\text{(the magenta line in Fig.~\ref{figTrainingSpeedSigmoid}, right).}
\end{equation}
The smaller $\lambda$ is, the closer to $L_2/L_1$ ($\gg 1$) the ratio $R_{\rm sigmoid}(\lambda)$ is and the lower the relative contribution of the noisy region to $\nabla_{\theta_r}\cL_{\rm joint}$ is. {\em This is the regime in which $\cN_r$ can learn  more efficient in clean regions compared with the usual regression network without the uncertainty quantifier.} On the other hand, the larger $\lambda$ is, the closer to~$1$ the ratio $R_{\rm sigmoid}(\lambda)$ is and the higher the relative contribution of the noisy region to $\nabla_{\theta_r}\cL_{\rm joint}$ is.

{\bf Uncertainty quantifier.} Due to~\eqref{eqJointLossM} and~\eqref{eqHessianSigmoid}, the relative contribution of the clean and noisy regions $X^{j_1}$ and $X^{j_2}$ to $\nabla_{\theta_q}\cL_{\rm joint}$ is determined by the value
\begin{equation}\label{eqHessianRatioSigmoid}
Q_{\rm sigmoid}(\lambda) = \dfrac{L_1\lambda}{L_1+\lambda} \left(\dfrac{L_2\lambda}{L_2+\lambda} \right)^{-1}= \dfrac{L_1(L_2+\lambda)}{L_2(L_1+\lambda)} \quad\text{(the green line in Fig.~\ref{figTrainingSpeedSigmoid}, right).}
\end{equation}
The larger $\lambda$ is, the closer to $L_1/L_2$ ($\ll 1$) the ratio $Q_{\rm sigmoid}(\lambda)$ is and the higher the relative contribution of the noisy region to  $\nabla_{\theta_q}\cL_{\rm joint}$ is. On the other hand, the smaller $\lambda$ is, the closer to $1$ the ratio $Q_{\rm sigmoid}(\lambda)$ is and the more balanced the relative contributions of clean and noisy regions to $\nabla_{\theta_r}\cL_{\rm joint}$ is.

\subsubsection{Relative contribution of clean and noisy regions to the loss surface and its minima, depending on $\lambda$}\label{subsubsecImportanceSigmoid}
For clarity, assume we have only two regions with regressor's losses $L_1=L_1(\theta_r)$ and $L_2=L_2(\theta_r)$, respectively. As before, let $L_1\ll L_2$. 

\begin{figure}[!t]
\centering
	\begin{minipage}{0.95\textwidth}
       \includegraphics[width=\textwidth]{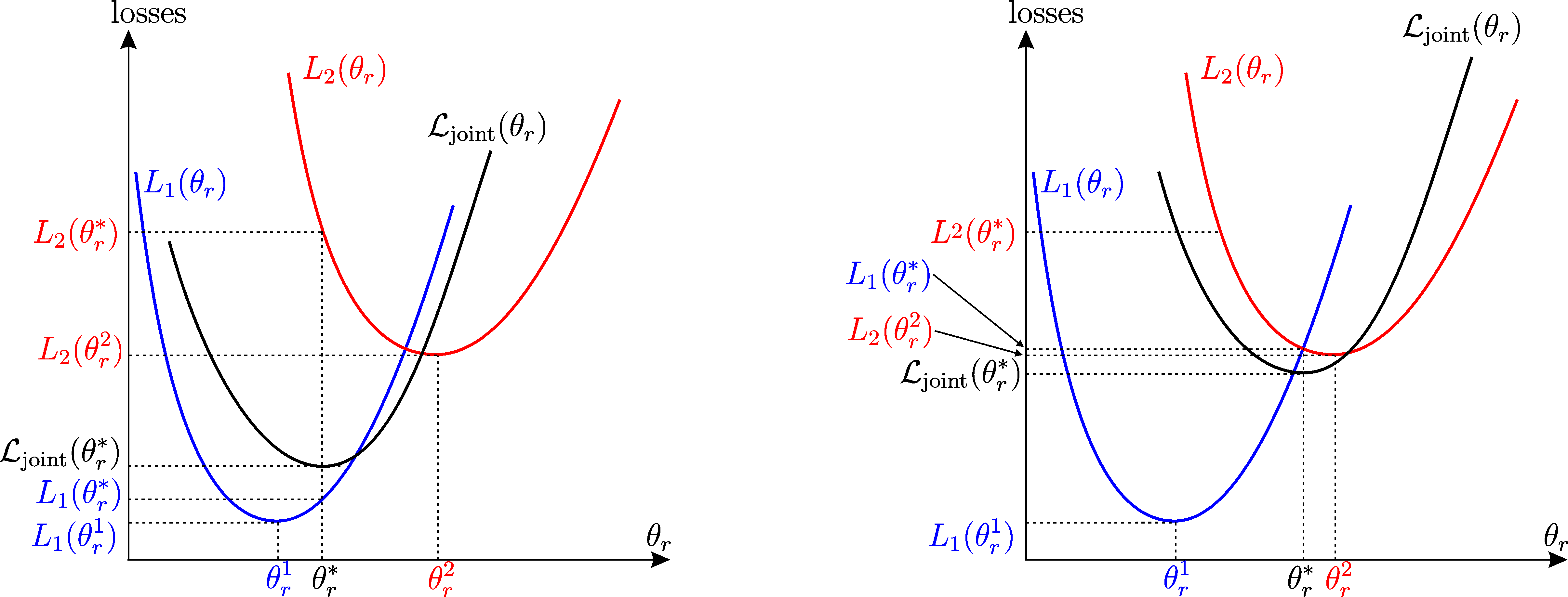}
	\end{minipage}
\caption{Regressor's losses $L_1(\theta_r)$ (blue) and $L_2(\theta_r)$ (red) defined with samples from clean and noisy regions, respectively, and the joint loss $\cL_{\rm joint}(\theta_r)$ defined with all samples as functions of regressor's weights $\theta_r$. The values $\theta_r^1$ and $\theta_r^2$ are the minima of $L_1(\theta_r)$  and $L_2(\theta_r)$, while $\theta_r^*$ is the minimum of $\cL_{\rm joint}(\theta_r)$. Left: Small values of $\lambda$ imply that $\theta_r^*$ is close to $\theta_r^1$; hence the loss $L_1(\theta_r^*)$ in the clean region is close to its optimal value $L_1(\theta_r^1)$. Right:  Large values of $\lambda$ imply that $\theta_r^*$ is far away from $\theta_r^1$; hence the loss $L_1(\theta_r^*)$ in the clean region is far away from its optimal value $L_1(\theta_r^1)$.}\label{figLosses}
\end{figure}

{\bf Regressor.} Due to~\eqref{eqJointLossM}, \eqref{eqMSigmoid} and Lemma~\ref{lPropertiesMSigmoid} (item~\ref{lPropertiesMSigmoid3}), 
\begin{equation}\label{eqNablaJointLossM}
\cL_{\rm joint} \approx \ln\left(1+\frac{\lambda}{L_1}\right)\left(C_1 L_1(\theta_r) + \frac{C_2}{R_{\rm sigmoid}(\lambda)} L_2(\theta_r) + C\right),
\end{equation}
where $C$ does not depend on $\theta_r$ and $R_{\rm sigmoid}(\lambda)$ is given by~\eqref{eqf1RatioSigmoid}. As~$\lambda\to 0$, we have $1/R_{\rm sigmoid}(\lambda)\to L_1/L_2\ll 1$, see the magenta line in~Fig.~\ref{figTrainingSpeedSigmoid} (right). Thus, for small~$\lambda$, the minimum $\theta_r^*$ of $\cL_{\rm joint}$ is generically close to the minimum $\theta_r^1$ of the loss $L_1(\theta_r)$ defined {\em only} with samples from the clean region. Figure~\ref{figLosses} schematically shows the loss surfaces of the loss $\cL_{\rm joint}(\theta_r)$ defined with all samples and  the loss surfaces defined with samples only from clean regions $L_1(\theta_r)$ and only from noisy regions $L_2(\theta_r)$. We see that, for small values of $\lambda$, the loss surface $\cL_{\rm joint}$ is close\footnote{After an appropriate vertical shift and rescaling, which do not affect its minima.} to that given by $L_1(\theta_r)$.

{\bf Uncertainty quantifier.} Now we  analyze the loss surface $\cL_{\rm joint}$ in the $\theta_q$-space for fixed values $L_1$ and $L_2$ of regressor's loss. Analogously to~\eqref{eqHessian}, we have
$$
\cL_{\rm joint} \approx C_1\tilde \cM_1(\theta_q) + C_2 \tilde \cM_2(\theta_q) + C,
$$
where $C$ does not depend on $\theta_q$,
$$
\begin{aligned}
\tilde \cM_j(\theta_q) = \frac{1}{2}
\dfrac{\partial^2 \cM(L,\xi^j(\theta_q^j))}{\partial\xi^j(\theta_q^j)} \nabla_{\theta_q}\xi^j(\theta_q^j)\cdot \nabla_{\theta_q}\xi^j(\theta_q^j)^T\cdot (\theta_q-\theta_q^j)^2 \\
=  \frac{1}{2}
\dfrac{L_j\lambda}{L_j+\lambda} \nabla_{\theta_q}\xi^j(\theta_q^j)\cdot \nabla_{\theta_q}\xi^j(\theta_q^j)^T\cdot (\theta_q-\theta_q^j)^2,
\end{aligned}
$$
and $\theta_q^j$ is the minimum of $\cM(L_j,\xi^j)$ (which is the loss taking into account only the samples from the $j$th region, see~\eqref{eqJointLossM} and~\eqref{eqMSigmoid}). Now we see that the distance between the minimum $\theta_q^*$ of the joint loss $\cL_{\rm joint}$ and the minima $\theta_q^j$ depends on the ratio $Q_{\rm sigmoid}(\lambda)$ in~\eqref{eqHessianRatioSigmoid}. In particular, the larger $\lambda$ is the closer $\theta_q^*$ to $\theta_q^2$ (the minimum corresponding to the noisy region) is.

Note that, due to the assumption in Condition~\ref{condThetaQ}, the above argument is valid only if the uncertainty quantifier correctly estimates both values $L_1$ and $L_2$ of regressor's losses in clean and noisy regions. These estimates dynamically change during training.
Again, due to~\eqref{eqHessianRatioSigmoid}, if $\lambda$ is large, the uncertainty quantifier ``ignores'' clean regions and rather learns from samples in noisy regions. As a result, it may wrongly quantify clean regions as noisy. Hence, it erroneously  suppresses their contribution to the gradient $\nabla_{\theta_r}\cL_{\rm joint}$ with respect to the regressor's weights $\theta_r$. On the other hand, if $\lambda$ is small, clean and noisy regions equally contribute to the learning of the quantifier's weights $\theta_q$, which facilitates the fit of the regressor in clean regions.

\subsection{Softplus output of $\cN_q$}\label{subsecSoftplusLambda}
Assume the output $z$ of the uncertainty quantifier $\cN_q$ is implemented as the softplus nonlinearity, see Sec.~\ref{subsecSoftplusProba}.  We will justify the column ``softplus'' in Table~\ref{table}. Similarly to Lemma~\ref{lPropertiesMSigmoid}, one can show that the function $\cM(L,\xi)$ is convex with respect to $\xi$, and, for each $L>0$, it achieves a global minimum with respect to $\xi$ at the point
    $
    \overline\xi =\overline\xi(L) = \ln\left(e^{\lambda/L}-1\right)
    $
and
\begin{equation}\label{eqFZXiSoftplus}
  f(Z(\overline\xi))=\frac{\lambda}{L}.
\end{equation}

\subsubsection{Impact of $\lambda$ on the overall learning speed}\label{subsecOverallLearningSoftplus}

{\bf Regressor.} Similarly to Sec.~\ref{subsecOverallLearningSigmoid}, due to~\eqref{eqFZXiSoftplus} the overall learning speed of $\cN_r$ is determined by
\begin{equation}\label{eqf1Softplus}
\left|\nabla_{\theta_r}\cM(L,\xi)\right| \propto \dfrac{\lambda}{L} \quad \text{(the magenta line in Fig.~\ref{figTrainingSpeedSoftplus}, left).}
\end{equation}

{\bf Uncertainty quantifier.} The overall learning speed of $\cN_q$ is determined by
\begin{equation}\label{eqHessianSoftplus}
\left|\nabla_{\theta_q}\cM(L,\xi(\theta_q))\right| \propto \dfrac{L^2}{\lambda}\left(1-e^{-\lambda/L}\right)^2\quad \text{(the green line in Fig.~\ref{figTrainingSpeedSoftplus}, left).}
\end{equation}
We note that the function in~\eqref{eqHessianSoftplus} asymptotically equals $\lambda$ as $\lambda\to0$ and $L^2/\lambda$ as $\lambda\to\infty$. In particular, it tends to zero both as $\lambda\to 0$ and $\lambda\to\infty$. It is easy to calculate that it achieves its maximum
$$
\dfrac{4\mu_0}{(1+2\mu_0)^2} L \approx 0.4 \cdot L \quad \text{(the dashed line in Fig.~\ref{figTrainingSpeedSoftplus}, left).}
$$
at $\lambda = \mu_0 L$, where $\mu_0 \approx 1.3$ is a positive root of the equation $e^{\mu_0} = 1+2\mu_0$.

\subsubsection{Relative contribution of clean and noisy regions to the gradients of the loss, depending on $\lambda$}\label{subsecRelativeLearningSoftplus}
As in Sec.~\ref{subsecRelativeLearningSigmoid}, assume that $X^{j_1}$ is a clean region and $X^{j_2}$ is a noisy region in the sense that
$
L_2\gg L_1.
$

{\bf Regressor.} Due to~\eqref{eqJointLossM}  and~\eqref{eqf1Softplus}, the relative contribution of the clean and noisy regions to $\nabla_{\theta_r}\cL_{\rm joint}$ is determined by the value
\begin{equation}\label{eqf1RatioReLU}
R_{\rm soft}=\dfrac{L_2}{L_1}\quad \text{(the magenta line in Fig.~\ref{figTrainingSpeedSoftplus}, right).}
\end{equation}
In particular, it does not depend on $\lambda$, and clean regions always dominate (provided $L_1$ and $L_2$ are correctly estimated by the uncertainty quantifier). 

{\bf Uncertainty quantifier.}  However, the relative contribution of the clean and noisy regions to $\nabla_{\theta_q}\cL_{\rm joint}$ does depend on $\lambda$. Due to~\eqref{eqJointLossM}  and~\eqref{eqHessianSoftplus}, it is determined by the value
\begin{equation}\label{eqHessianRatioReLU}
Q_{\rm soft}(\lambda) = \left(\dfrac{L_1\left(1-e^{-\lambda/L_1}\right)}{L_2\left(1-e^{-\lambda/L_2}\right)}\right)^2\quad \text{(the green line in Fig.~\ref{figTrainingSpeedSoftplus}, right).}
\end{equation}
As $\lambda$ increases, $Q_{\rm soft}(\lambda)$ decays to $(L_1/L_2)^2\ll 1$. Note that this limit is even smaller than the limit  $L_1/L_2$ of $Q_{\rm sigmoid}(\lambda)$ in the case of the sigmoid output. Hence, in this case the main contribution to the gradient $\nabla_{\theta_q}\cL_{\rm joint}$  is due to noisy regions. On the other hand, for small $\lambda$, $Q_{\rm soft}(\lambda)$ is close to $1$, i.e., the contributions of clean and noisy regions get balanced.

\subsubsection{Relative contribution of clean and noisy regions to the loss surface and its minima, depending on $\lambda$}\label{subsubsecImportanceSoftplus}

{\bf Regressor.} Repeating the argument in Sec.~\ref{subsubsecImportanceSigmoid} and using~\eqref{eqFZXiSoftplus}, we see that the distance between the minimum $\theta_r^*$ of $\cL_{\rm joint}(\theta_r)$ and the minimum $\theta_r^1$ of $L_1(\theta_r)$ depends on the value $R_{\rm soft}$ in~\eqref{eqf1RatioReLU}. In particular, it is small if $R_{\rm soft}\ll 1$, independently of whether $\lambda$ is small or large.
 
{\bf Uncertainty quantifier.} However, the contribution of the clean region to the loss surface $\cL_{\rm joint}$ in the $\theta_q$-space is governed by the ratio $Q_{\rm soft}(\lambda)$ in~\eqref{eqHessianRatioReLU}, which does depend on $\lambda$.  As $\lambda$ increases, $Q_{\rm soft}(\lambda)$ decays to $(L_1/L_2)^2\ll 1$, see the green line in Fig.~\ref{figTrainingSpeedSoftplus} (right). Saying differently, the minimum $\theta_q^*$ of the joint loss $\cL_{\rm joint}$ with respect to $\theta_q$ gets close to the minimum $\theta_q^2$ of the loss defined with the samples from the noisy region only. As a result, clean regions can be misidentified as noisy. On the other hand, as in Sec.~\ref{subsubsecImportanceSigmoid}, we see that this does not happen for small $\lambda$, and, therefore, the fit of the regressor in clean regions gets facilitated.

\section{Synthetic data} \label{secArtificial}

In this section, we generate data with $X\subset[0,1]$ and $Y\subset\bbR$ and implement the uncertainty quantifiers with the sigmoid output, see Sec.~\ref{subsecSigmoidProba}. We choose regressor's loss to be the MSE and generate Gaussian noise. We predict the uncertainty in terms of the standard deviation, using~\eqref{eqEVVar}.

\subsection{Smooth data}\label{subsecArtSmooth}
\begin{figure}[!t]
	\begin{minipage}{0.5\textwidth}
	   \includegraphics[width=\textwidth]{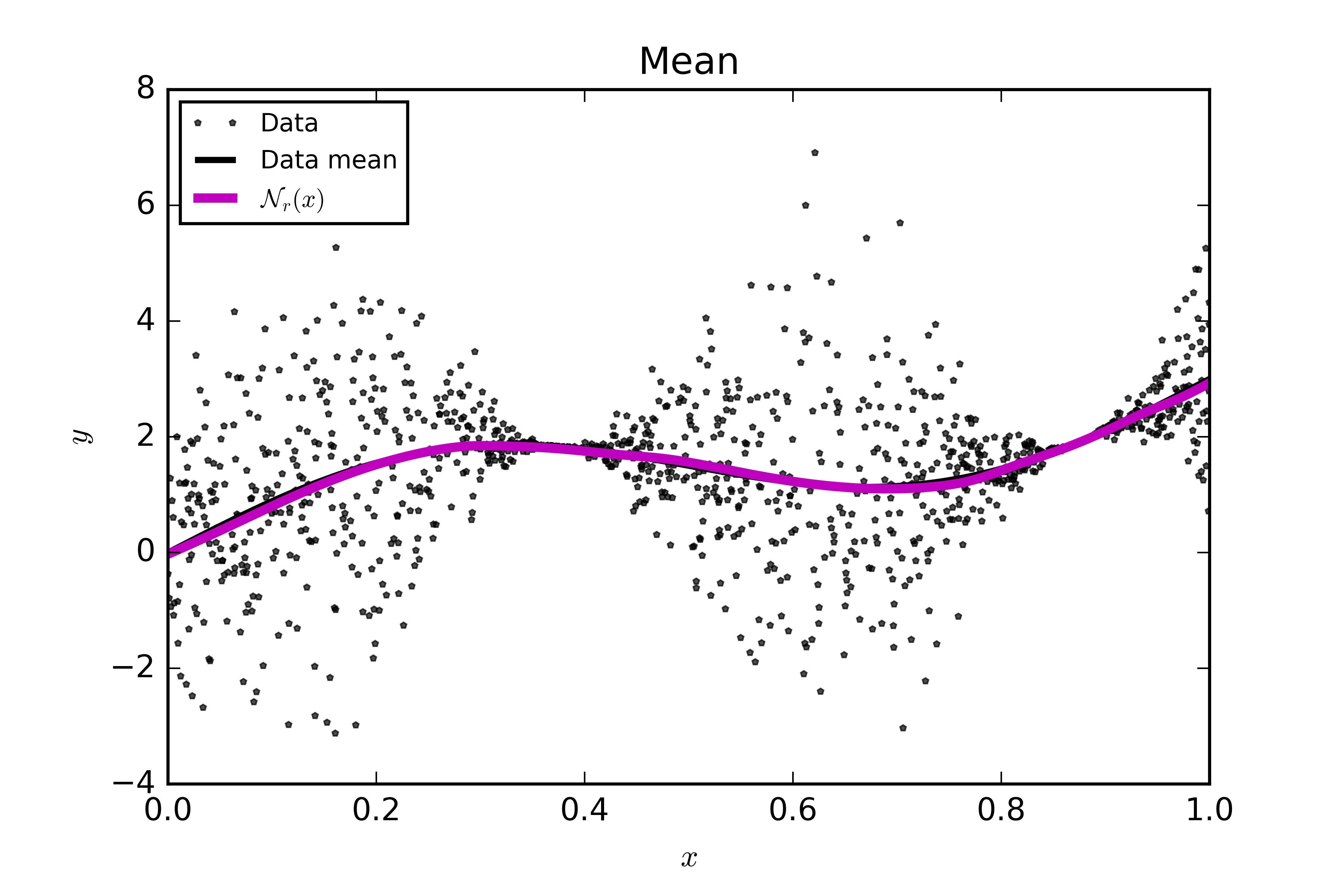}
	\end{minipage}
\hfill
	\begin{minipage}{0.5\textwidth}
       \includegraphics[width=\textwidth]{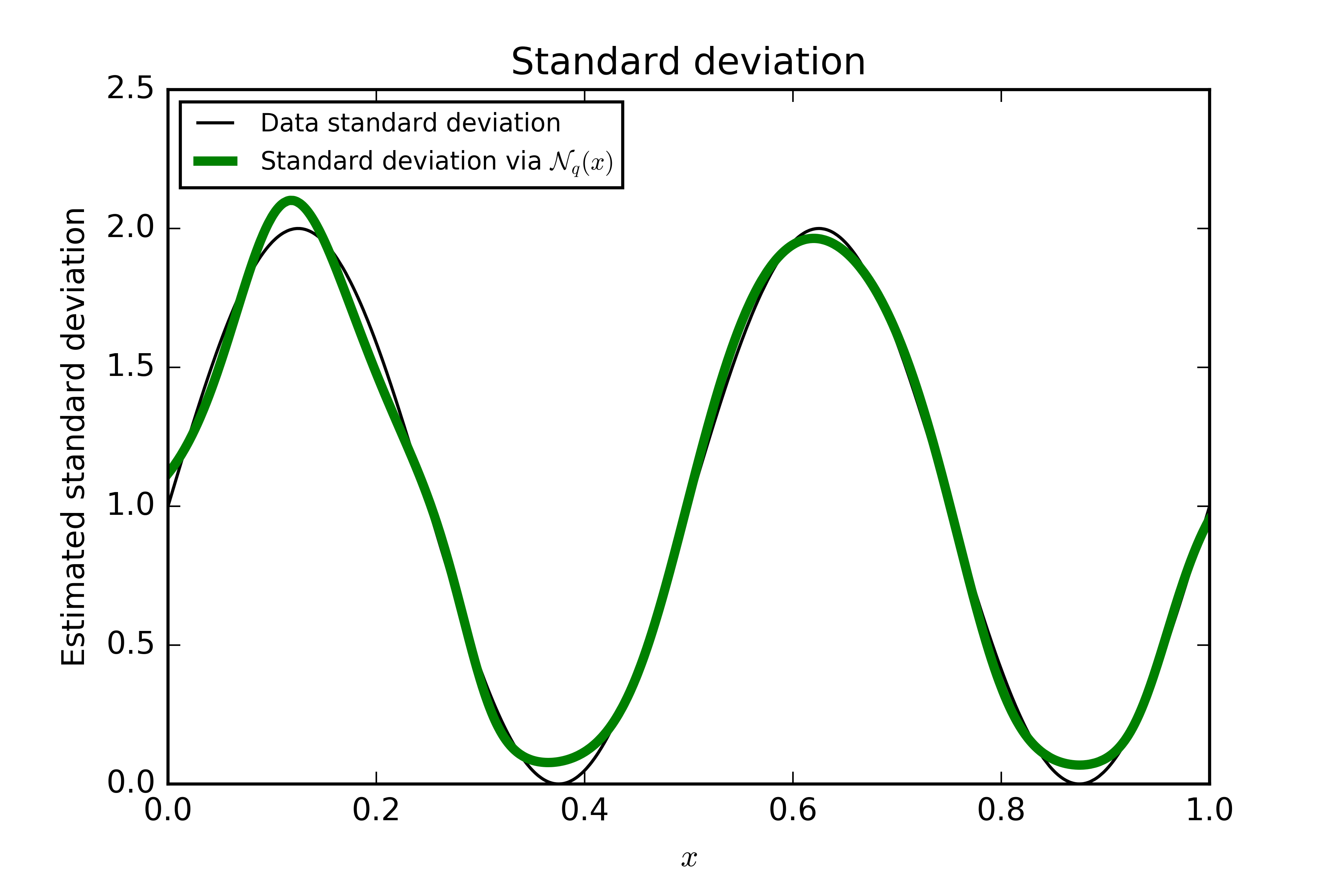}
	\end{minipage}
\caption{Fit of the networks $(\cN_r,\cN_q)$ in~\eqref{eqImplemArtifSigmoid} with $\lambda=0.1$, MSE regressor's loss for the data $y$ sampled from the normal distribution with mean $3x + \sin(2 \pi x)$ and standard deviation $1 + \sin(4\pi x)$. Left: $\cN_r(x)$ (the curve is almost indistinguishable from the black line indicating the mean of the data). Right: standard deviation via $\cN_q(x)$ according to~\eqref{refLrf1f2Intro}.}\label{figArtSmooth}
\end{figure}

\begin{figure}[!t]
	\begin{minipage}{0.5\textwidth}
       \includegraphics[width=\textwidth]{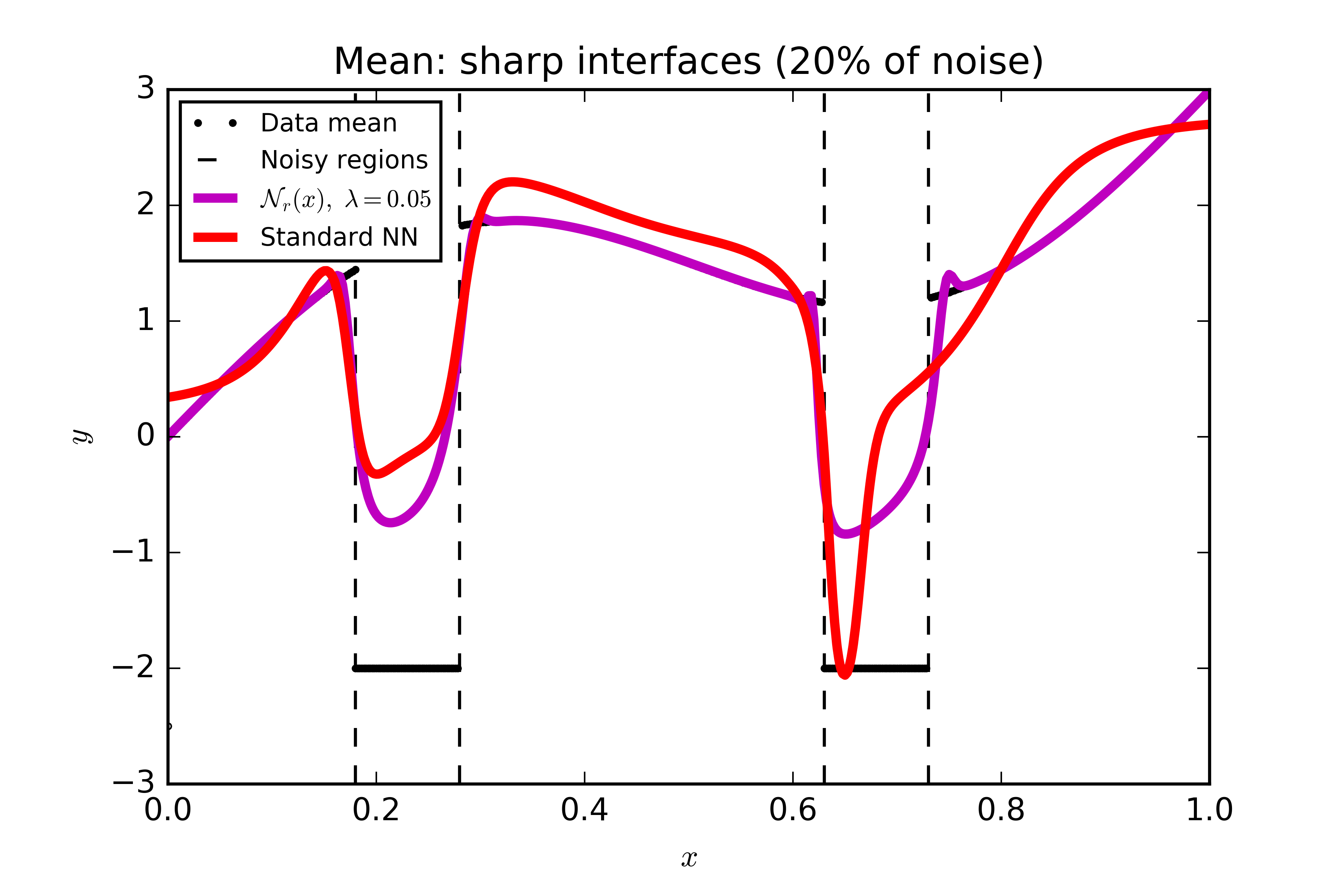}
	\end{minipage}
	\hfill
    \begin{minipage}{0.5\textwidth}
       \includegraphics[width=\textwidth]{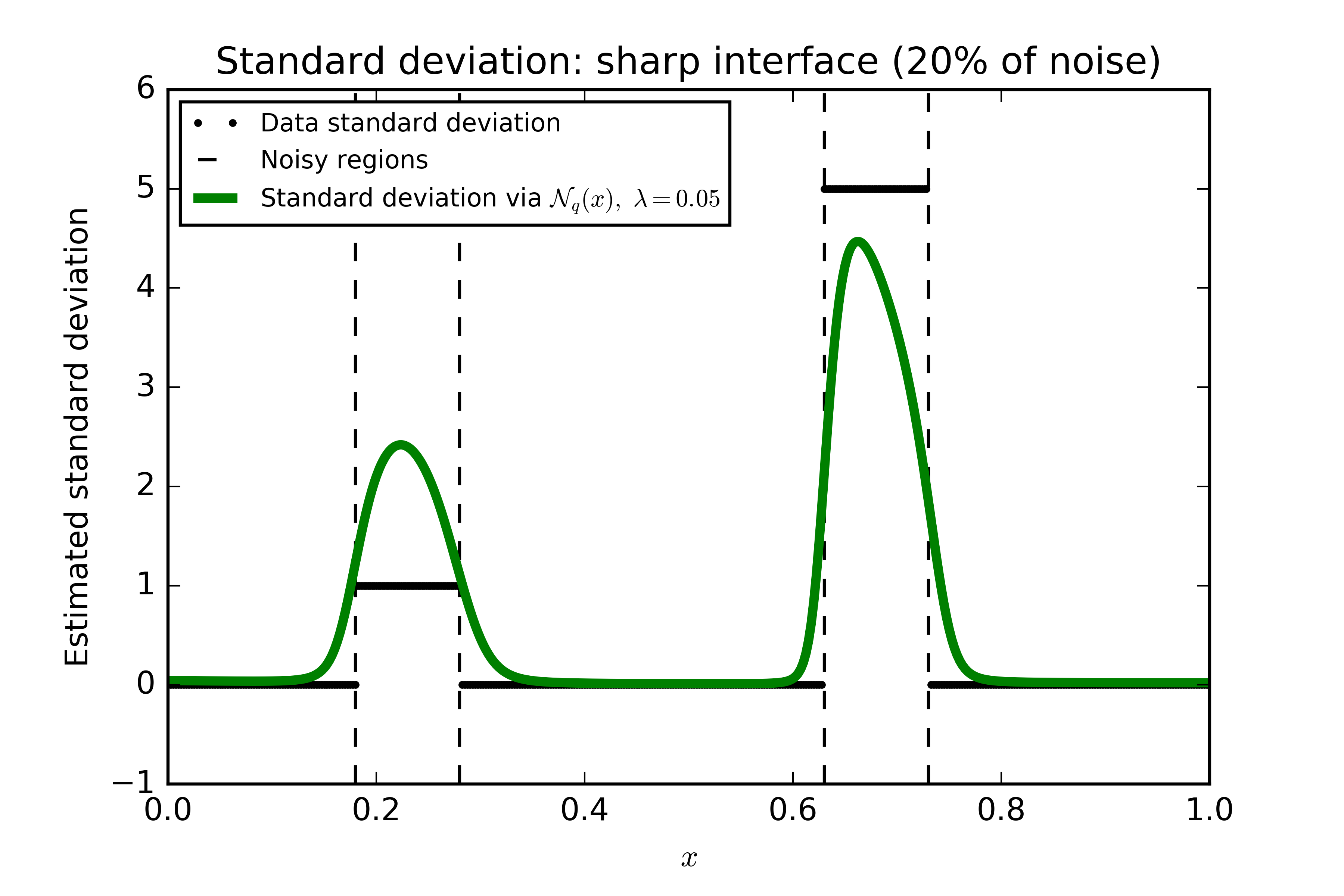}
	\end{minipage}

    \begin{minipage}{0.5\textwidth}
       \includegraphics[width=\textwidth]{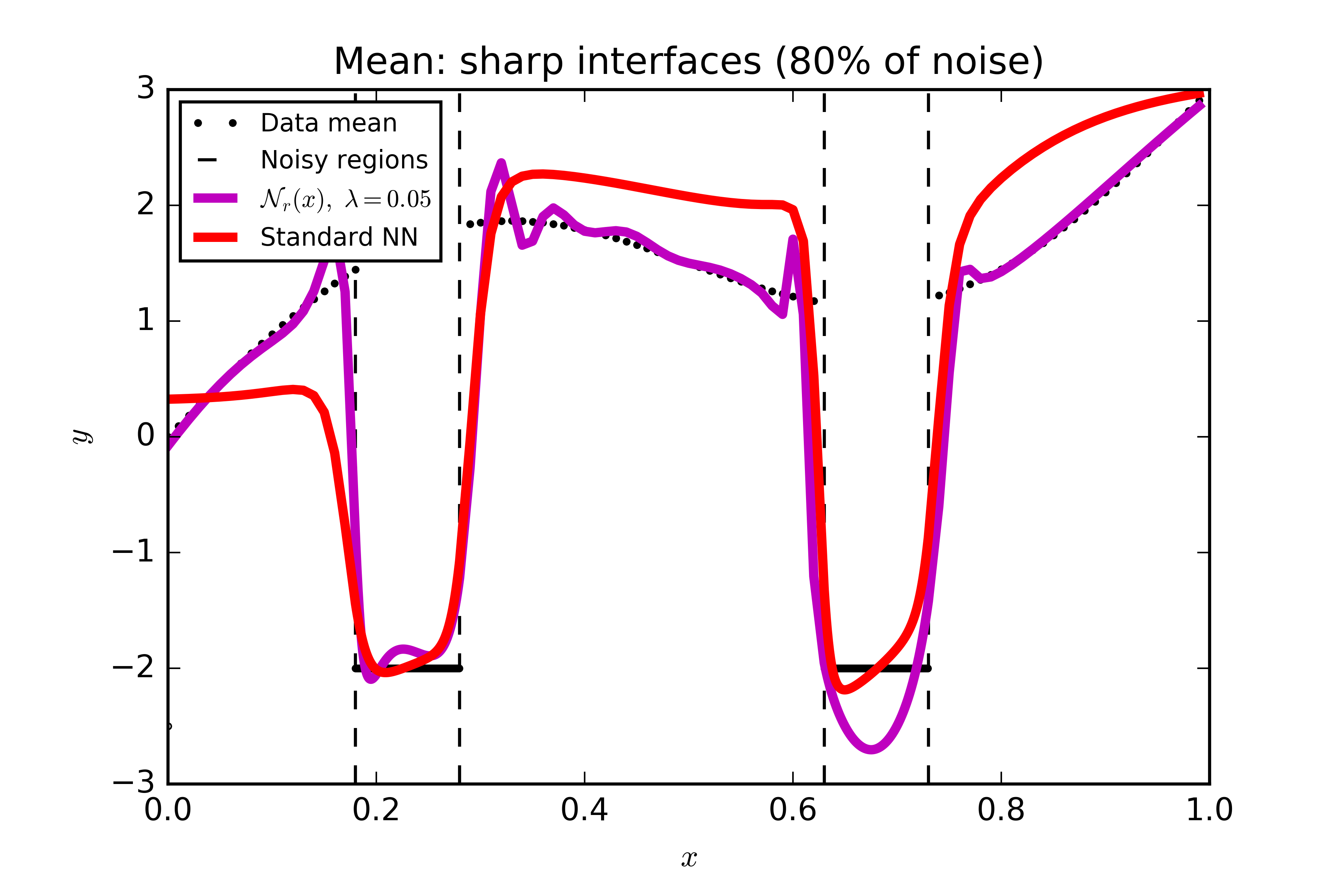}
	\end{minipage}
	\hfill
    \begin{minipage}{0.5\textwidth}
       \includegraphics[width=\textwidth]{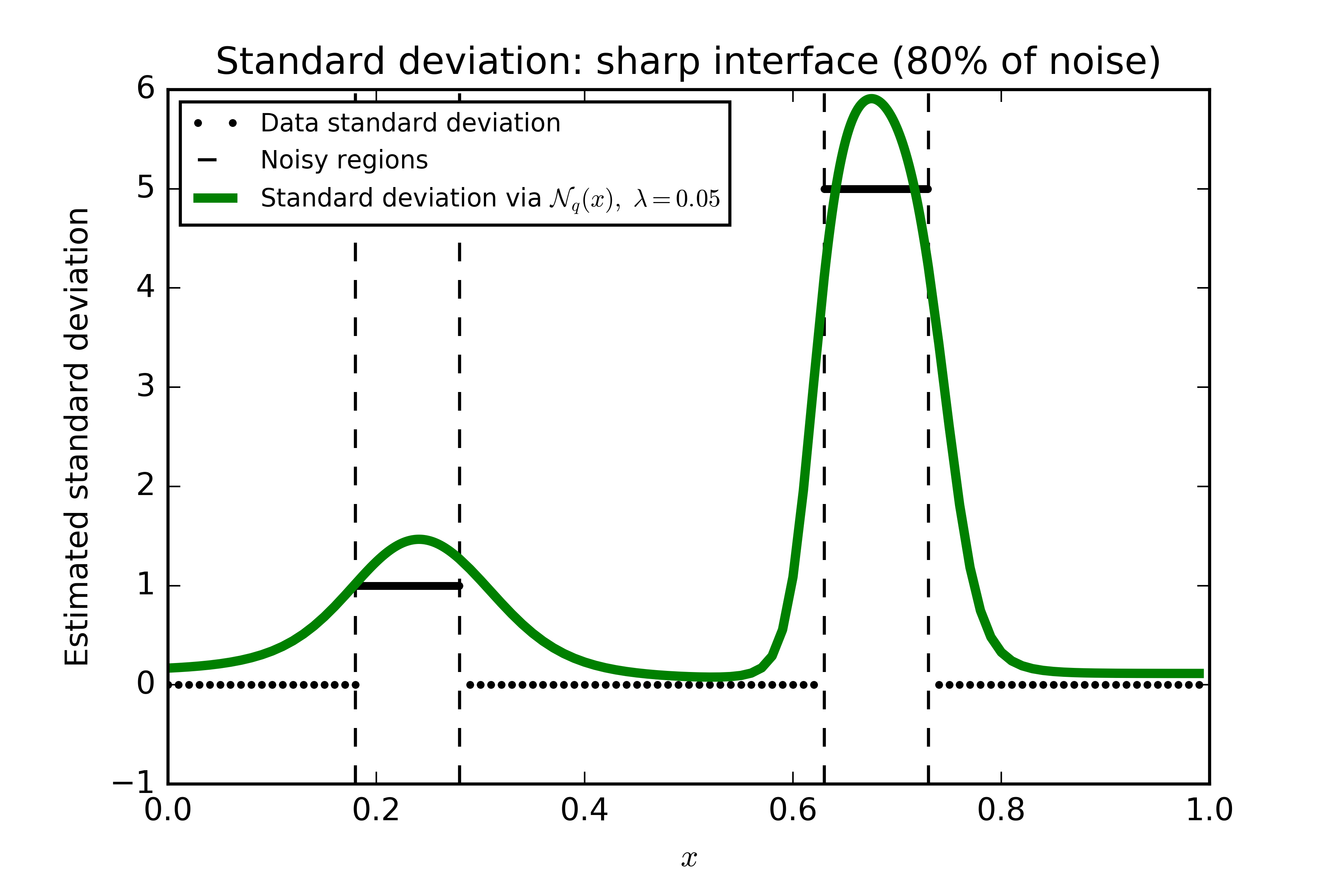}
	\end{minipage}
\caption{Fit of the networks $(\cN_r,\cN_q)$ in~\eqref{eqImplemArtifSigmoid} with MSE regressor's loss in comparison with a standard NN (in red). The dashed vertical lines indicate the two regions of width $0.1$ each with Gaussian noise with mean $-2$ and standard deviations $1$ and $5$, respectively. Top: noisy regions contain $20\%$ of the data. Bottom: noisy regions contain $80\%$ of the data. In both cases, $\lambda = 0.05$.}\label{figArtSharp}
\end{figure}

First, we consider smoothly varying mean and variance. Namely, we sample $y$ from the normal distribution with mean $3x + \sin(2 \pi x)$ and standard deviation $1 + \sin(4\pi x)$, $x\in[0,1]$. We implement the network pair $(\cN_r,\cN_q)$ as follows:
\begin{equation}\label{eqImplemArtifSigmoid}
\left\{
\begin{aligned}
&\cN_r:\ \text{input(1),\ 2 x hidden(10,\ tanh),\ output(1,\ linear)},\\
&\cN_q:\ \text{input(1),\ 2 x hidden(10,\ tanh),\ output(1,\ sigmoid)}.
\end{aligned}
\right.
\end{equation}
We take $\lambda = 0.1$.
The fit of the regressor $\cN_r(x)$ is illustrated by Fig.~\ref{figArtSmooth} (left) and the predictions of the standard deviation via~\eqref{refLrf1f2Intro} are shown in Fig.~\ref{figArtSmooth} (right). The corresponding choice of $\lambda$, $f$, and $g$ is discussed in Sec.~\ref{subsecArtSmooth}.
Figure~\ref{figArtSmooth} represents the situation where both the mean of the data and its variance vary smoothly. In this case, the choice of $\lambda$ is not too important from the practical point of view. However, it becomes crucial once the data exhibits rather sharp interfaces. Figure~\ref{figArtSharp} illustrates  ``clean'' data given by $y=3x + \sin (2\pi x)$, $x\in[0,1]$, complemented by two vertical strips of width $0.1$ with the Gaussian noise (mean is $-2$ in each strip and standard deviation is $1$ in the first strip and $5$ in the second). One can see that the standard NN (red line) estimates the mean in the clean regions outside the two noisy strips much worse than the regressor $\cN_r$ (blue line). This is especially evident in the lower picture, where the noisy regions contain $80\%$ of the data.  These and other simulations are discussed in more detail in Sec.~\ref{secArtificial}. We refer to Sec.~\ref{secBoston} for the analysis of real world data sets.


\subsection{Data with sharp interfaces}\label{subsecSharpInterface}
In our second example, we generate ``clean'' data given by $y=3x + \sin (2\pi x)$, $x\in[0,1]$, and complement them by two vertical strips of width $0.1$ with Gaussian noise (mean is $-2$ in each strip and standard deviation is $1$ in the first strip and $5$ in the second).
The fit of the network~\eqref{eqImplemArtifSigmoid} is illustrated in Fig.~\ref{figArtSharp}.
One can see that even if the data contains $80\%$ of noise, the uncertainty quantifier allows the regressor to fit well enough for the remaining~$20\%$. This can be explained by formula~\eqref{eqJointLossSumProb} for the loss and by the results in Sec.~\ref{secSigmoidSoftReLambda}. Indeed, small coefficients $M_j/N$ in~\eqref{eqJointLossSumProb} corresponding to regions $X_j$ with low density would also be present in the loss function of a standard NN for regression. As a consequence, the samples from these regions would contribute little to the gradient of the loss, and the gradient descent would be mostly governed by the samples from the noisy regions. On the other hand, as we saw in Sec.~\ref{secSigmoidSoftReLambda} (cf. also Table~\ref{table} and Fig.~\ref{figTrainingSpeedSigmoid} and \ref{figTrainingSpeedSoftplus} (right)), in case of the joint loss~\eqref{eqJointLossSumProb}, small values of $\lambda$ yield larger gradients of the terms $\bbE_j[\cL_r(\cdot,y_r^j)] f(z^j) + \lambda g(z^j)$ corresponding to clean regions $X_j$, which compensate the small values of  $M_j/N$.

\begin{figure}[!t]
	\begin{minipage}{0.5\textwidth}
       \includegraphics[width=\textwidth]{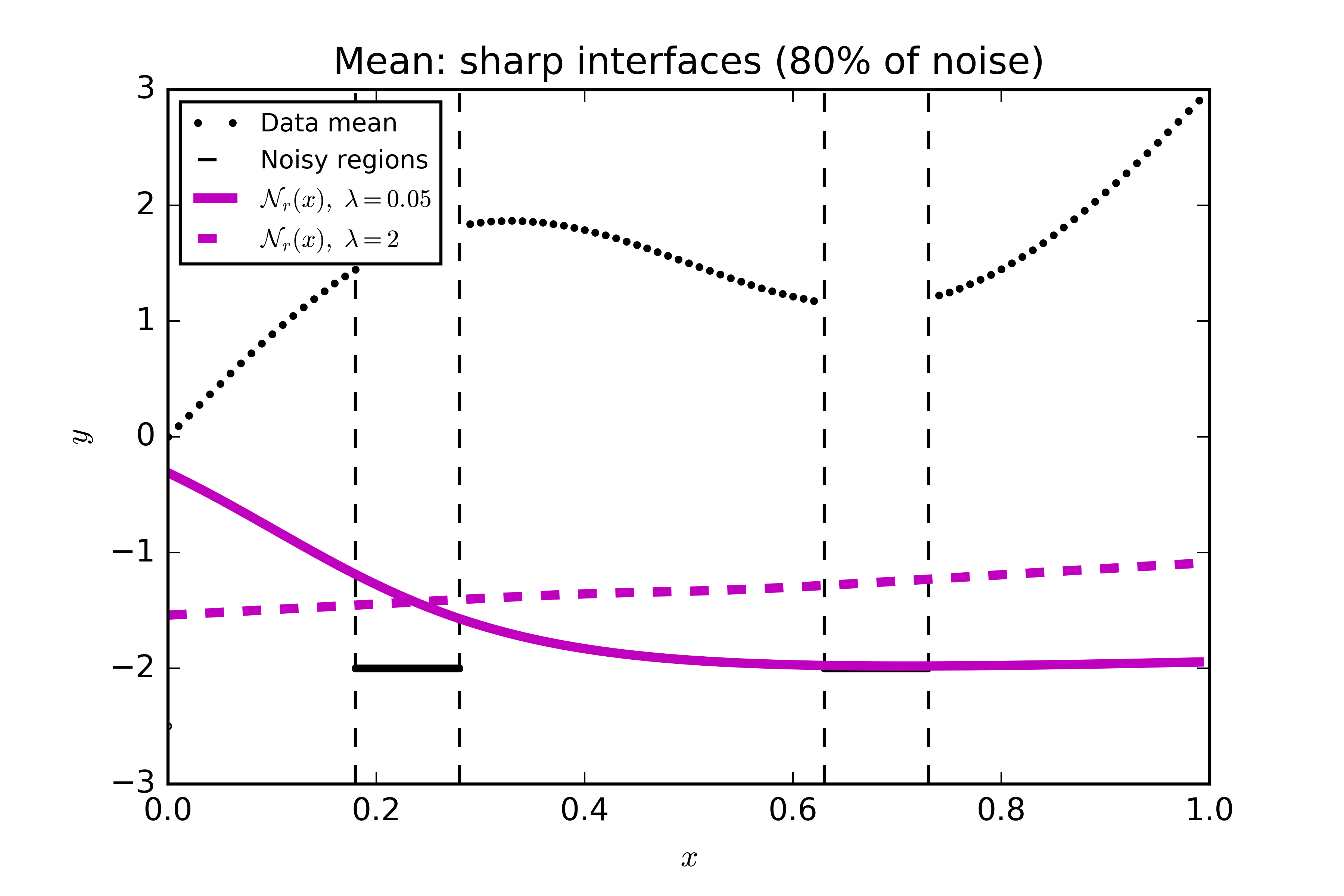}
	\end{minipage}
	\hfill
    \begin{minipage}{0.5\textwidth}
       \includegraphics[width=\textwidth]{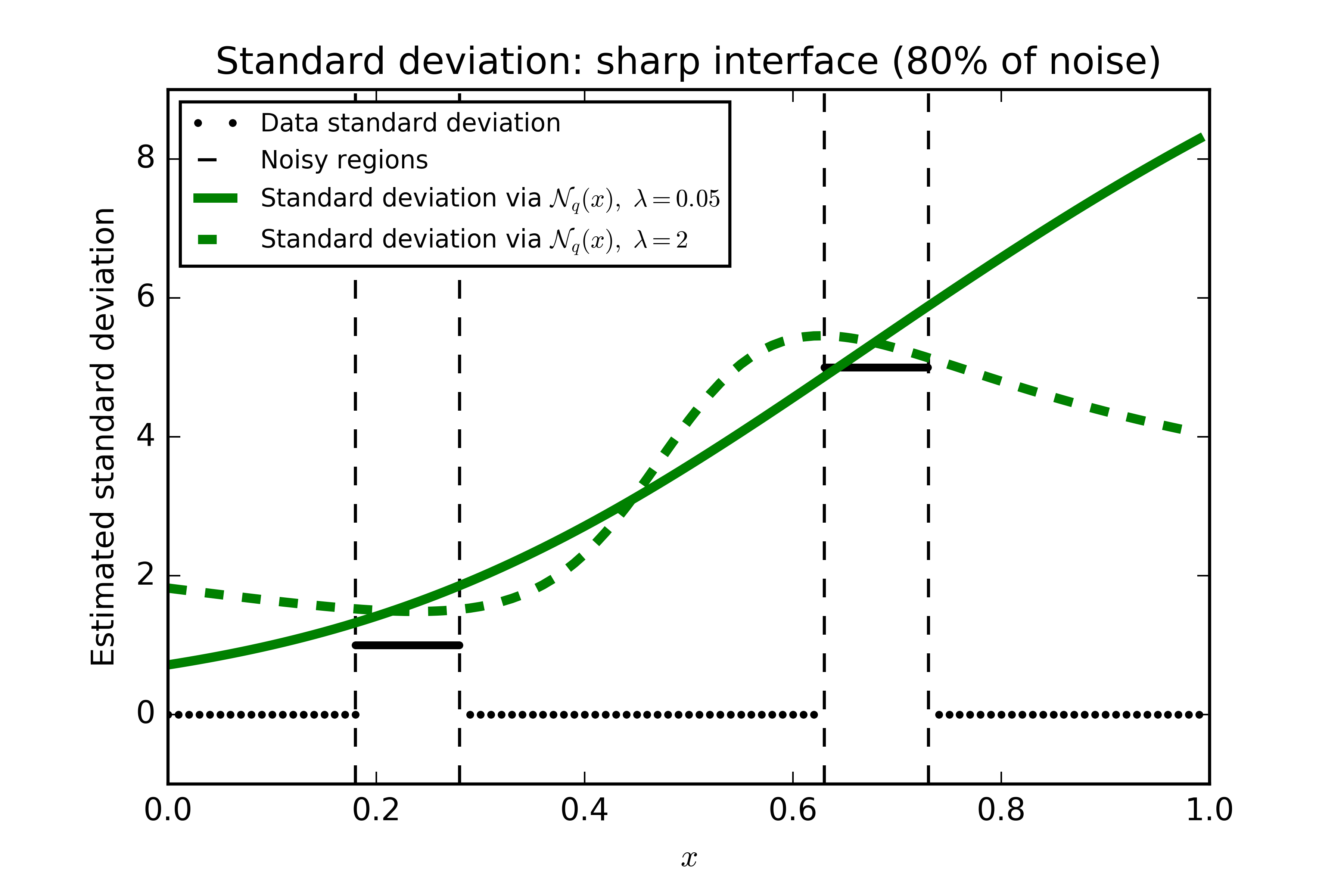}
	\end{minipage}

	\begin{minipage}{0.5\textwidth}
       \includegraphics[width=\textwidth]{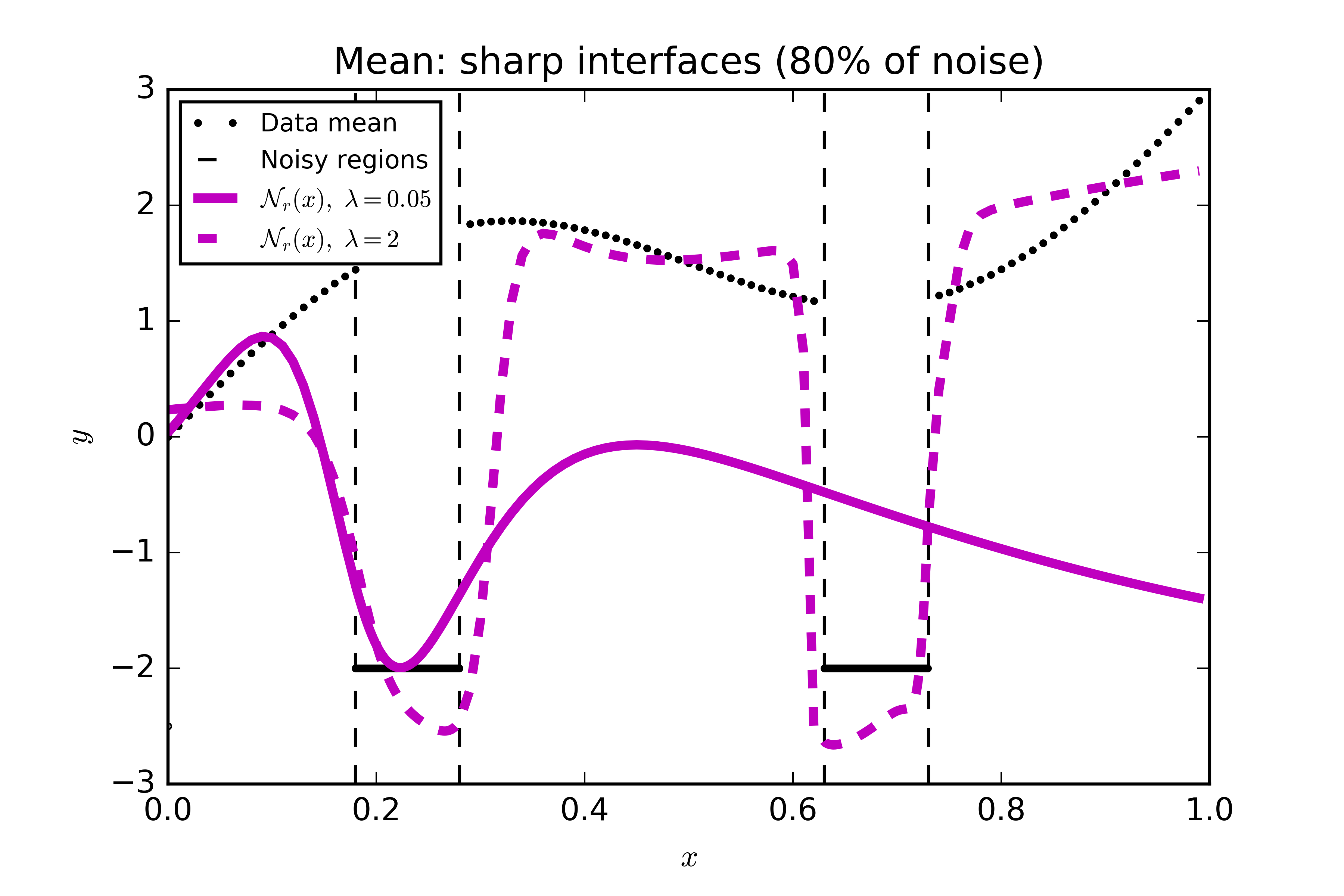}
	\end{minipage}
	\hfill
    \begin{minipage}{0.5\textwidth}
       \includegraphics[width=\textwidth]{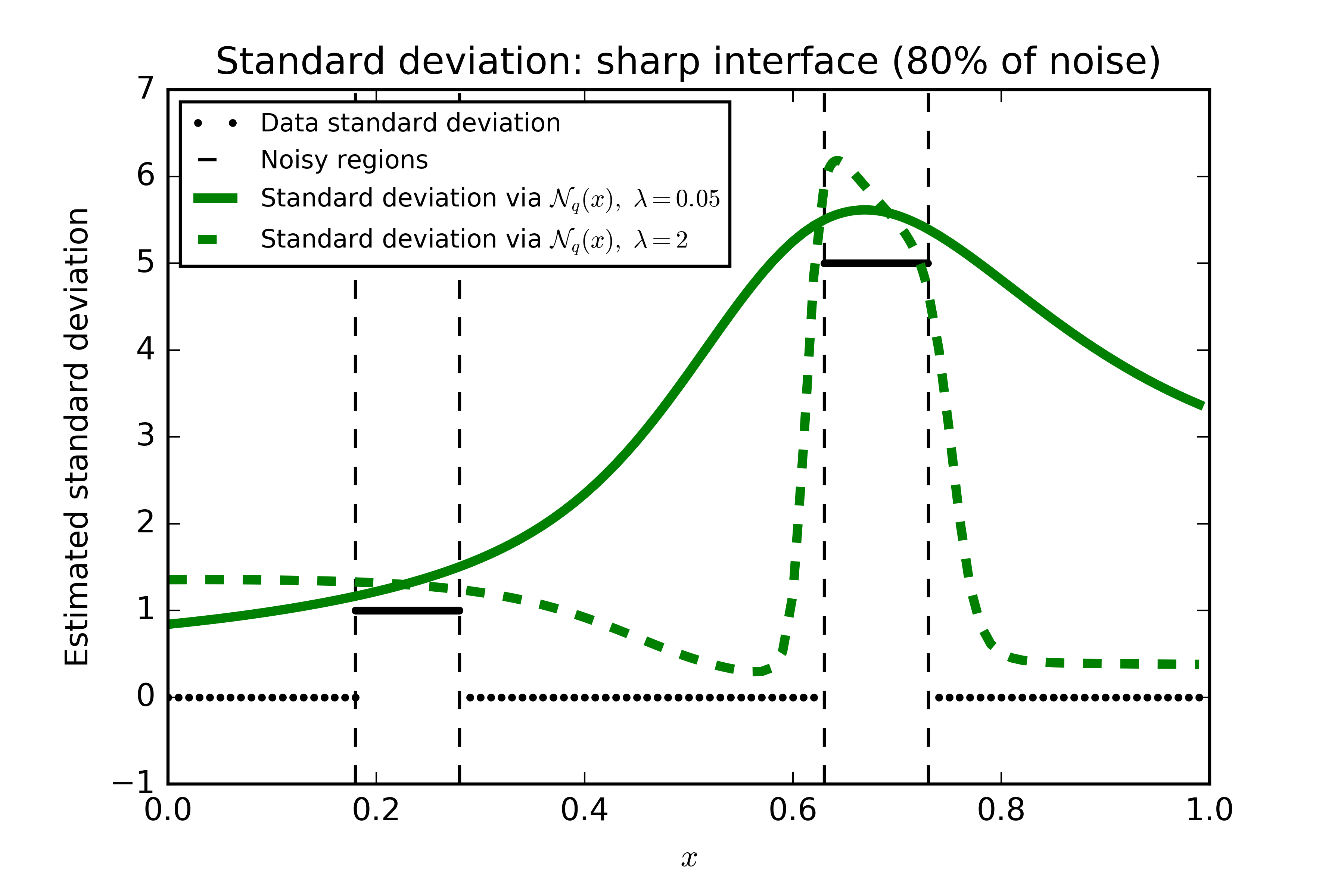}
	\end{minipage}

	\begin{minipage}{0.5\textwidth}
       \includegraphics[width=\textwidth]{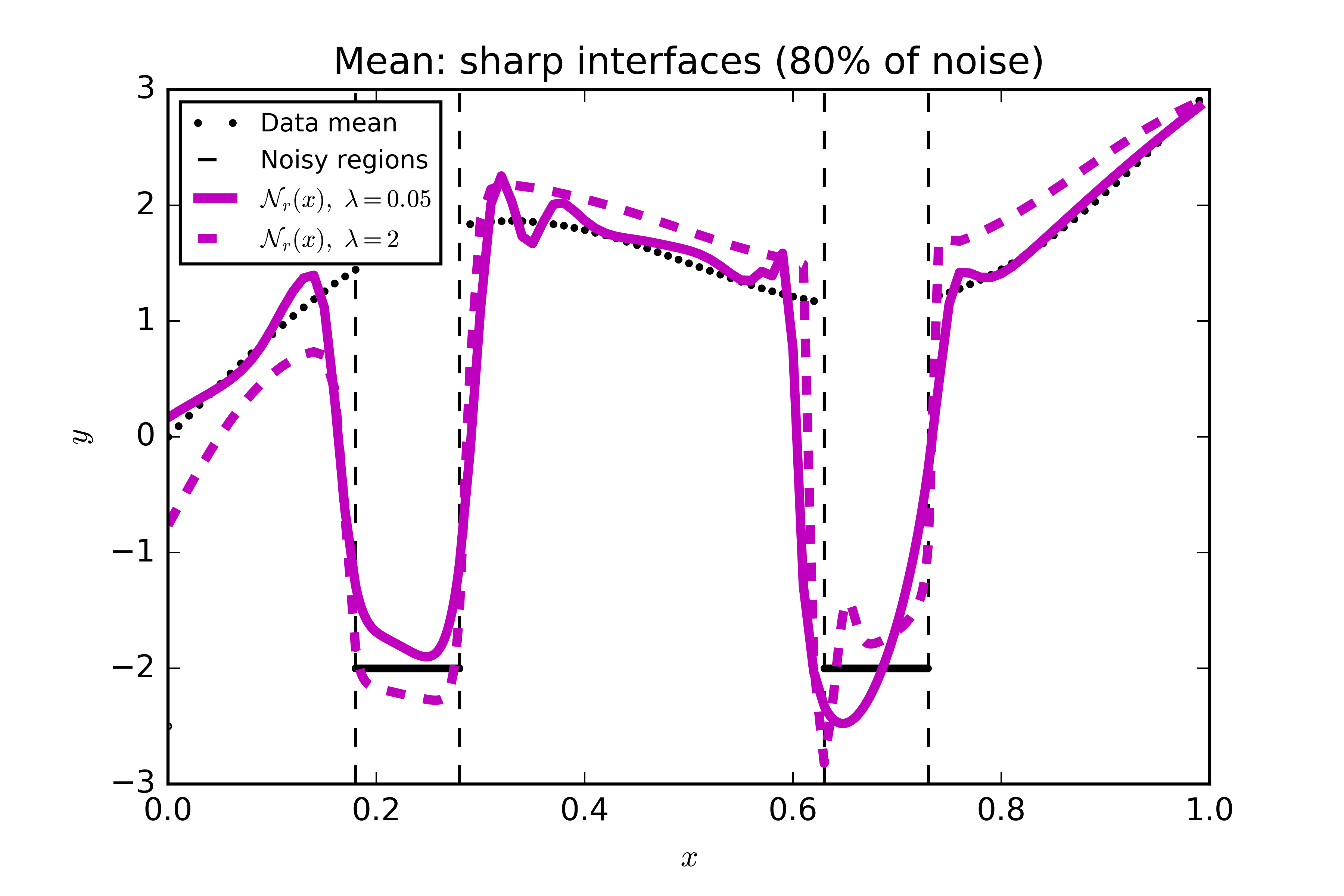}
	\end{minipage}
	\hfill
    \begin{minipage}{0.5\textwidth}
       \includegraphics[width=\textwidth]{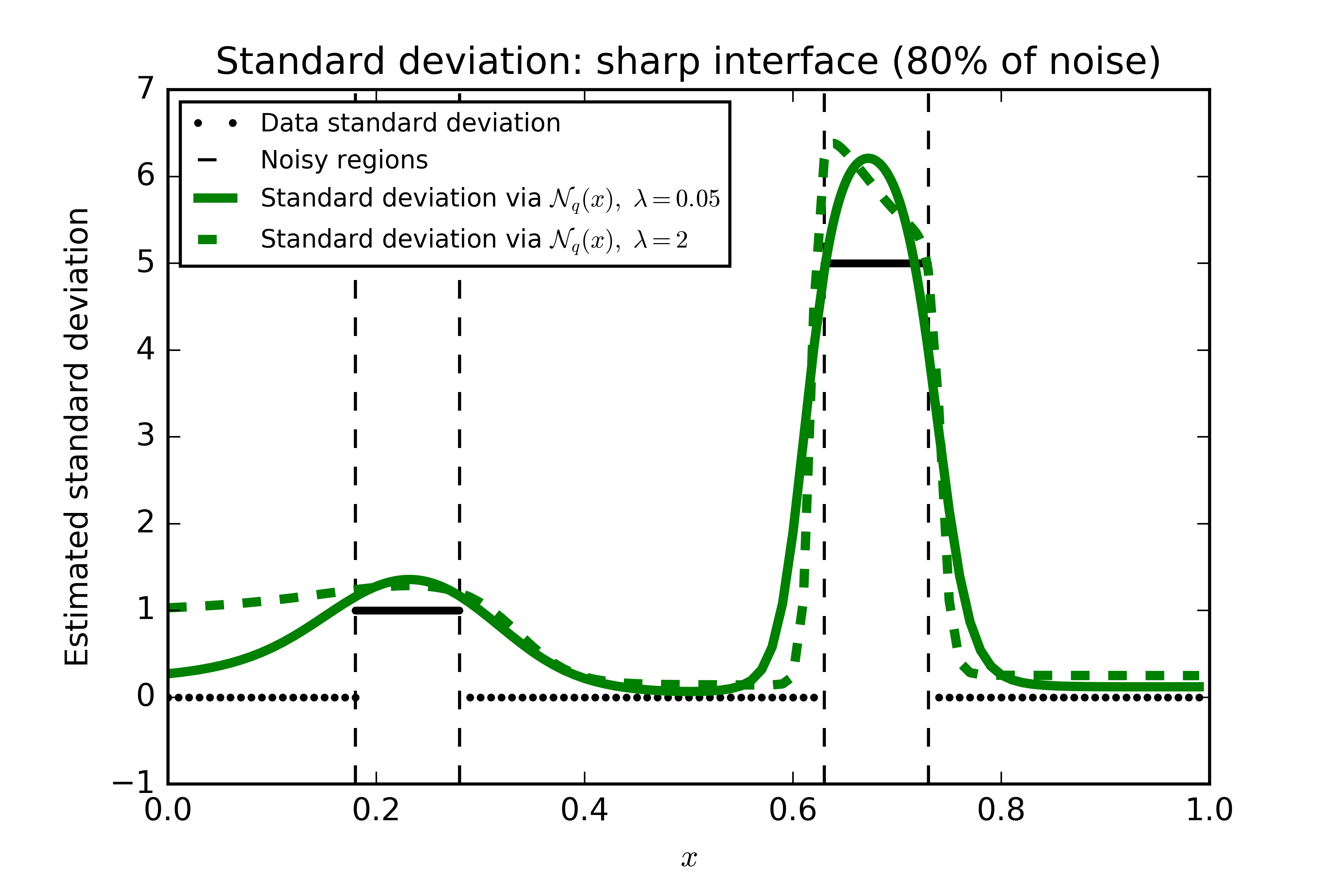}
	\end{minipage}
\caption{Fit of the networks $\cN_r$ (left) and $\cN_q$ with the sigmoid output (right). Solid lines correspond to $\lambda = 0.05$ and dashed lines to $\lambda = 2$. Top: early stage. Middle: middle stage. Bottom: final stage.}\label{figSigmoidLearningStages}
\end{figure}

Next, we compare the case  $\lambda = 0.05$ (Fig.~\ref{figSigmoidLearningStages}, the solid lines)  with the case $\lambda=2$  (Fig.~\ref{figSigmoidLearningStages}, the dashed lines) and illustrate how $\lambda$ affects the relative learning speeds in clean and noisy regions (cf. Sec.~\ref{subsecRelativeLearningSigmoid} and Fig.~\ref{figTrainingSpeedSigmoid}, right). We fill in the two noisy strips with 80\% of the data and plot the fit of the network~$\cN_r$ and the standard deviation via $\cN_q$ according to~\eqref{eqEVVar} on three different learning stages.
The four vertical dashed lines in Fig.~\ref{figSigmoidLearningStages} divide the interval $[0,1]$ in five subintervals. We refer to them as regions 1--5 from left to right (with regions 2 and 4 filled with the Gaussian noise).

Consider $\lambda=0.05$ (bold lines in Fig.~\ref{figSigmoidLearningStages}). In this case, the regressor $\cN_r$ learns faster in the regions that have smaller variance according to $\cN_q$. Indeed, we see that $\cN_r$ first starts to learn in regions 1 and 2, where $\cN_q$ is smaller (top figure), and then additionally in region 3, where $\cN_q$ was moderately larger (middle figure). As $\cN_q$ learns that the variance in region 5 is less than in region 4, the regressor $\cN_r$ accelerates its learning in region 5 (bottom picture).

Now consider $\lambda=2$ (dashed lines in Fig.~\ref{figSigmoidLearningStages}). In this case,  $\cN_q$ learns faster in the noisy regions. In particular, this results (bottom figure) in the domination of the noisy region 2 over the clean region 1. This prevents $\cN_q$ from learning that region 1 has a smaller variance compared with region 2. On the other hand, the learning speeds of the regressor $\cN_r$ in clean and noisy regions are closer to each other. As a result, it is not able to estimate regions 1, 3, and 5 (outside of the noisy strips) as well as it does for $\lambda=0.05$.

\begin{figure}[!t]
	\begin{minipage}{0.5\textwidth}
       \includegraphics[width=\textwidth]{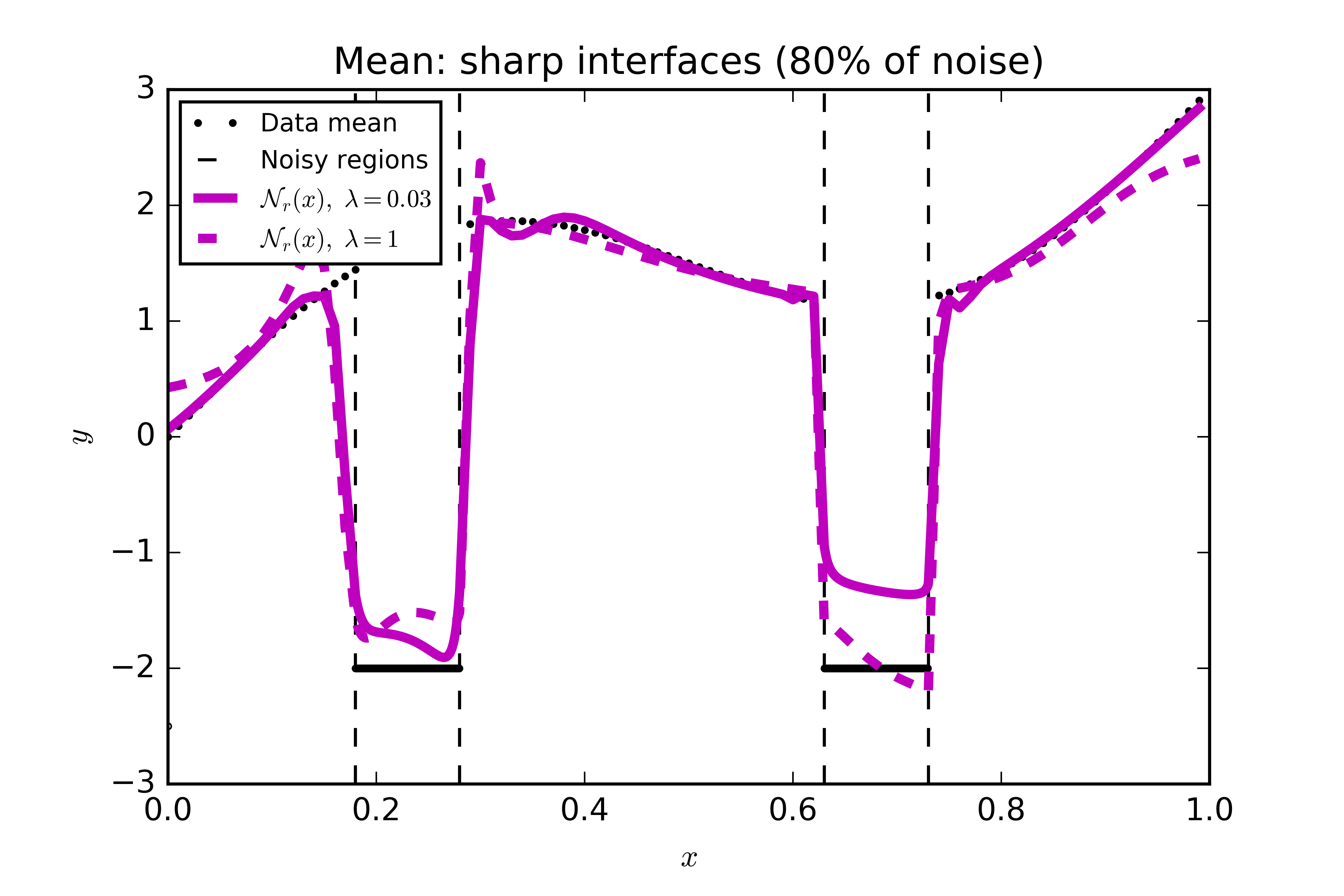}
	\end{minipage}
	\hfill
    \begin{minipage}{0.5\textwidth}
       \includegraphics[width=\textwidth]{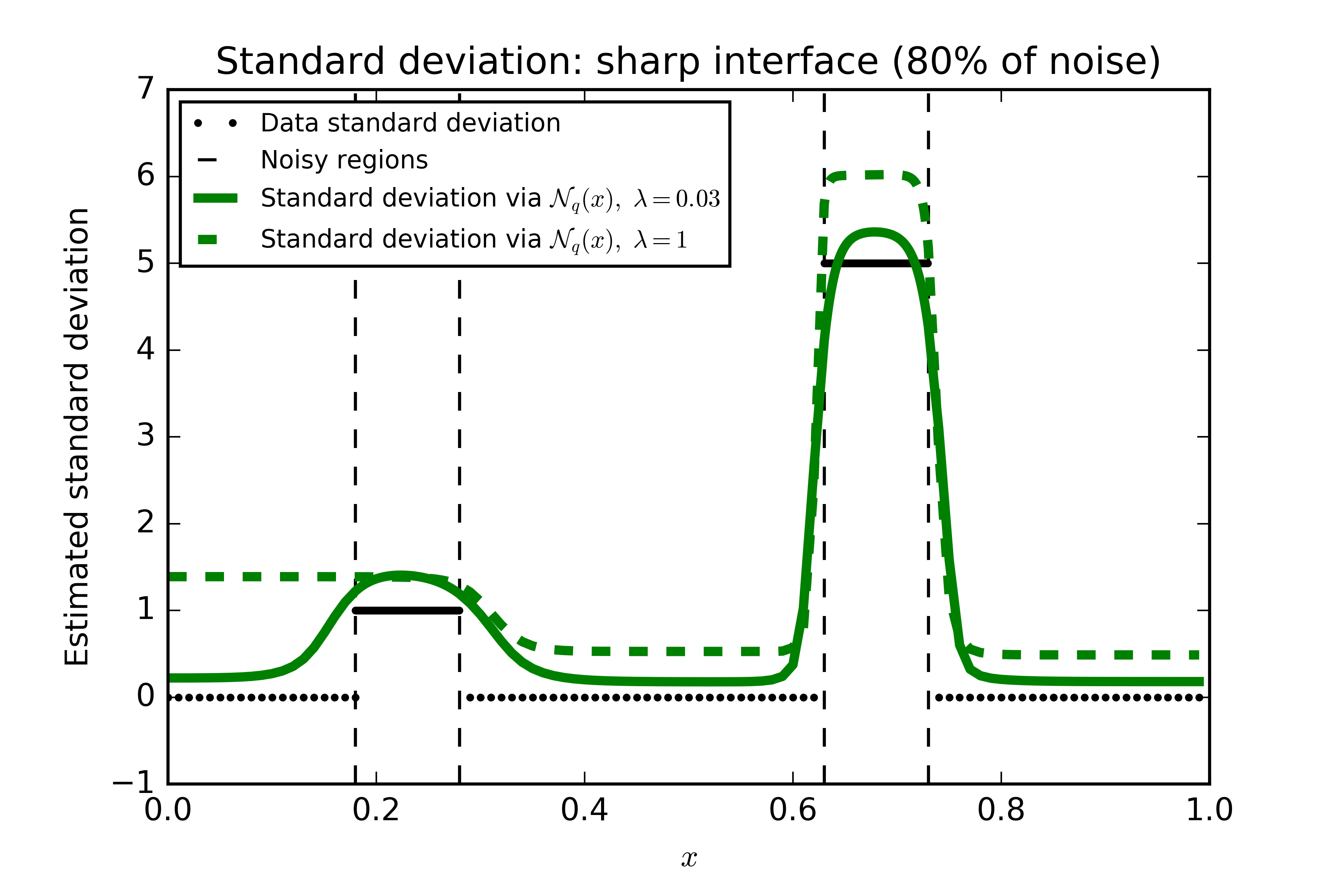}
	\end{minipage}
\caption{Fit of the networks $\cN_r$ (left) and $\cN_q$ with the softplus output (right). Solid lines correspond to $\lambda = 0.03$ and dashed lines to $\lambda = 1$.}\label{figSoftplusSharp}
\end{figure}

Finally, we illustrate the implementation of the output of $\cN_q$ via the softplus, see Sec.~\ref{subsecSoftplusProba}. The architecture of $(\cN_r,\cN_q)$ is identical to~\eqref{eqImplemArtifSigmoid}, except for the sigmoid replaced by the softplus. In Fig.~\ref{figSoftplusSharp}, we compare the cases $\lambda=0.03$ (the solid lines) and $\lambda=1$ (the dashed lines). Recall that the minimization of the joint loss~\eqref{eqJointLoss} with $\lambda=1$ is equivalent to maximization of the log-likelihood of the Gaussian distribution, see Example~\ref{exProba1}. In Fig.~\ref{figSoftplusSharp}, we see that the choice $\lambda=0.03$ is more optimal. For larger $\lambda$, $\cN_q$ learns much faster in the noisy regions compared with the clean regions (cf. Sec.~\ref{subsecRelativeLearningSoftplus} and  Fig.~\ref{figTrainingSpeedSoftplus}, right). As a result, it cannot learn properly in the clean regions 1, 3, and 5, especially in region 1.

\section{Real world data}\label{secBoston}

\textbf{Data sets.} We analyzed the following publicly available data sets: Boston House Prices~\cite{Harrison78} ($506$ samples, 13 features), Concrete Compressive Strength~\cite{ICheng98} ($1030$ samples, 8 features), Combined Cycle Power Plant~\cite{Tufekci14,Kaya12} ($9568$ samples, 4 features), Yacht Hydrodynamics~\cite{Gerritsma81,Ortigosa07} (308 samples, 6 features), Kinematics of an 8 Link Robot Arm Kin8Nm\footnote{http://mldata.org/repository/data/viewslug/regression-datasets-kin8nm/} (8192 samples, 8 feature), and Year Prediction MSD~\cite{Lichman13} ($515345$ samples, 90 features). For each data set, a one-dimensional target variable is  predicted.
Each data set, except for the year prediction MSD, is randomly split into 50 train-test folds with 95\% of samples in each train subset. All the measure values reported below are the averages of the respective measure values over 50 folds. For the year prediction MSD, we used a single split recommended in~\cite{Lichman13}.
The data are normalized so that the input features and the targets
have zero mean and unit variance in the training set.

\textbf{Methods.}
We compare the  following methods:
\begin{enumerate}

  \item\label{itemML} the maximum likelihood method (ML), in which one maximizes the likelihood of the normal distribution; like in our method, two networks (one predicting the mean and another predicting the variance) are trained simultaneously,

  \item Stein variational gradient descent (SVGD)~\cite{LiuWang2016}; Bayesian method, in which one uses a particle approximation of the posterior distribution of the weights,

  \item the probabilistic back propagation (PBP)~\cite{HernandezLobato15}; Bayesian method, in which one minimizes the KL divergence from the exact posterior to the approximating one, using assumed density filtering~\cite{Opper98} and expectation-propagation~\cite{Minka01} methods,

  \item\label{itemOurMehtod} our method (with $\lambda=0.1, 0.2, 0.5, 1, 2, 5$),

  \item\label{itemEnsML} ensemble of 5 MLs (EnsML)~\cite{Lakshminarayanan16,Lakshminarayanan17},
  \item\label{itemOurEns} ensemble of 5 pairs of networks according to our method.
\end{enumerate}

To show that our approach works for different types of regressor's loss, we optimize the above methods for the root mean squared error (RMSE) and the mean absolute error (MAE). In the case of RMSE/MAE, we use the Gaussian~\eqref{eqGaussianDistribution} / Laplacian~\eqref{eqLaplaceDistribution} likelihood in the ML, SVGD, and EnsML, and we use the MSE/MAE regressor's loss in our method. In the case of MAE for the PBP, we still use the Gaussian likelihood because changing the likelihood would require a new approximation of the normalization constant in the assumed density filtering method applied in~\cite{HernandezLobato15}.

In the case of RMSE, our method quantifies uncertainty in terms of the expected squared error (ESA) due to~\eqref{eqEVVar}. The other non-ensemble methods quantify ESA as the variance of the predictive distribution. For the ensemble methods~\ref{itemEnsML} and~\ref{itemOurEns}, we use~\eqref{eqMeanVarianceEnsemble} for the predictive mean and ESA.

In the case of MAE, our method quantifies uncertainty in terms of the expected absolute error (EAE) due to~\eqref{eqEVMeanLaplace}. The other non-ensemble methods (except the PBP) quantify EAE as $1/\tau$, where $\tau$ is the parameter in~\eqref{eqLaplaceDistribution}. For the PBP, we use its predictive Gaussian distribution $\cN(y|\mu,V)$ to calculate the expectation of $|y-\mu|$, which yields $\sqrt{2V\pi}$ as EAE.
For the ensemble methods, we use~\eqref{eqMeanEAEEnsemble} for the predictive mean and EAE.

Note that the ML and our method estimate aleatoric uncertainty, the Bayesian methods SVGD and PBP estimate epistemic uncertainty, and ensembles take into account both types of uncertainty.

\textbf{Architectures.} We use architectures proposed in~\cite{HernandezLobato15,Lakshminarayanan16,Lakshminarayanan17}, namely, 1-hidden layer regressors and 1-hidden layer uncertainty quantifiers with ReLU
nonlinearities. Each NN contains 50 hidden units for all the data sets, except for MSD, where we use 100 hidden units. For our method, we use the sigmoid output of the uncertainty quantifier (see Secs.~\ref{subsecSigmoidProba} and~\ref{subsecSigmoidLambda}). We refer to Appendix~\ref{appendixHyperparameters} for the values of hyperparameters that we used for the different methods.

\textbf{Measures.} We use two measures to estimate the quality of the fit.
\begin{enumerate}
  \item The overall error: {\em RMSE} and {\em MAE}.
  \item The area under the following curve ({\em AUC}), measuring the trade-off between properly learning the mean and estimating uncertainty. Assume the test set contains $N$ samples. If we are interested in the RMSE, then we order the samples with respect to their predicted ESA. For each $n=0,\dots,N-1$, we remove $n$ samples with the highest ESA and calculate the RMSE for the remaining $N-n$ samples. We denote the result by ${\rm RMSE}(n)$ and plot it versus~$n$ as a continuous piecewise linear curve. The AUC is the area under this curve normalized by $N-1$:
      $$
      {\rm AUC} = \frac{1}{N-1}\sum\limits_{n=0}^{N-2}\frac{{\rm RMSE}(n)+{\rm RMSE}(n+1)}{2}.
      $$

      When we use MAE instead of RMSE, the AUC is calculated similarly with MAE$(n)$ instead of RMSE$(n)$.
\end{enumerate}

\begin{figure}[!t]
	\begin{minipage}{0.16\textwidth}
	   \includegraphics[width=\textwidth]{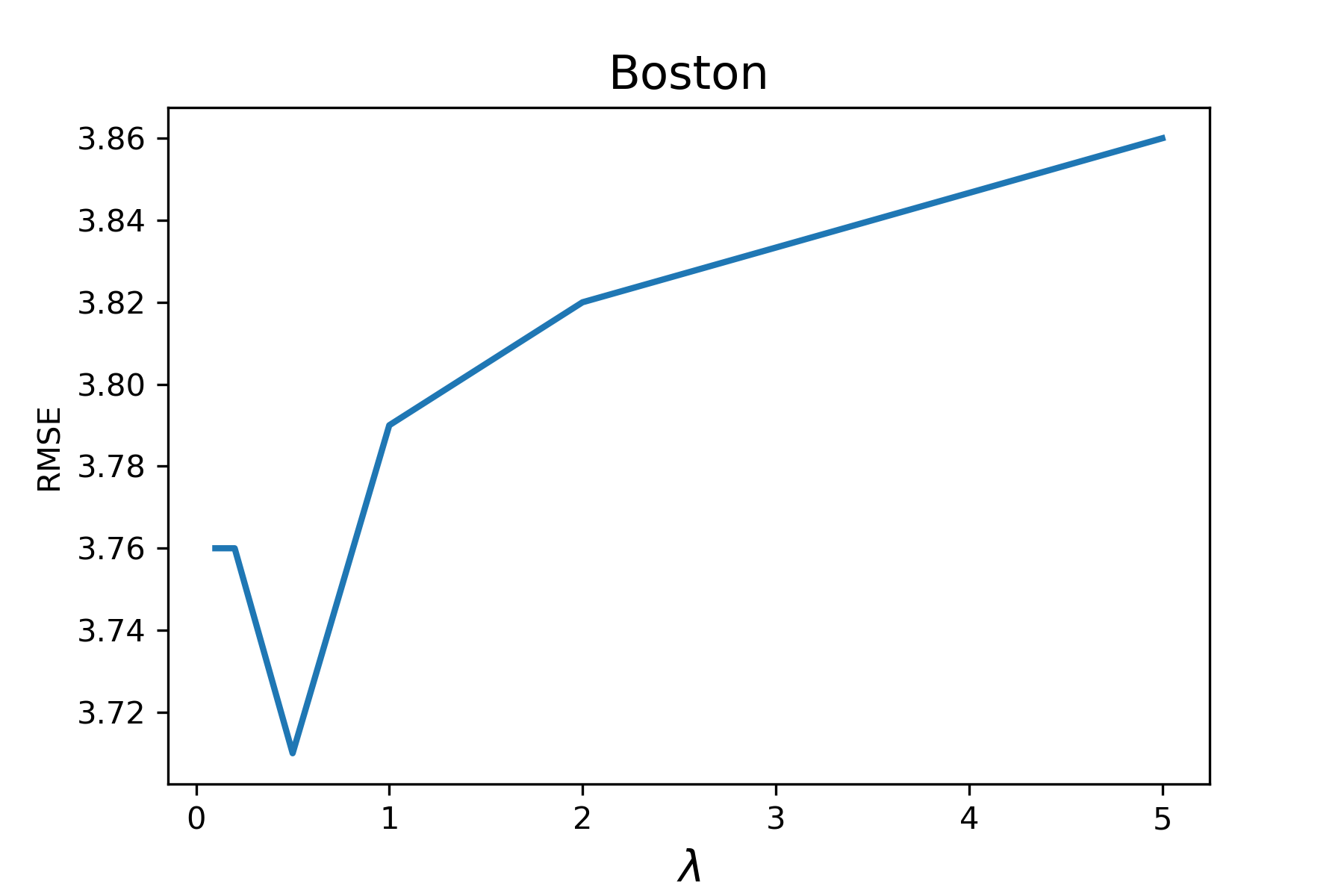}
	\end{minipage}
	\begin{minipage}{0.16\textwidth}
	   \includegraphics[width=\textwidth]{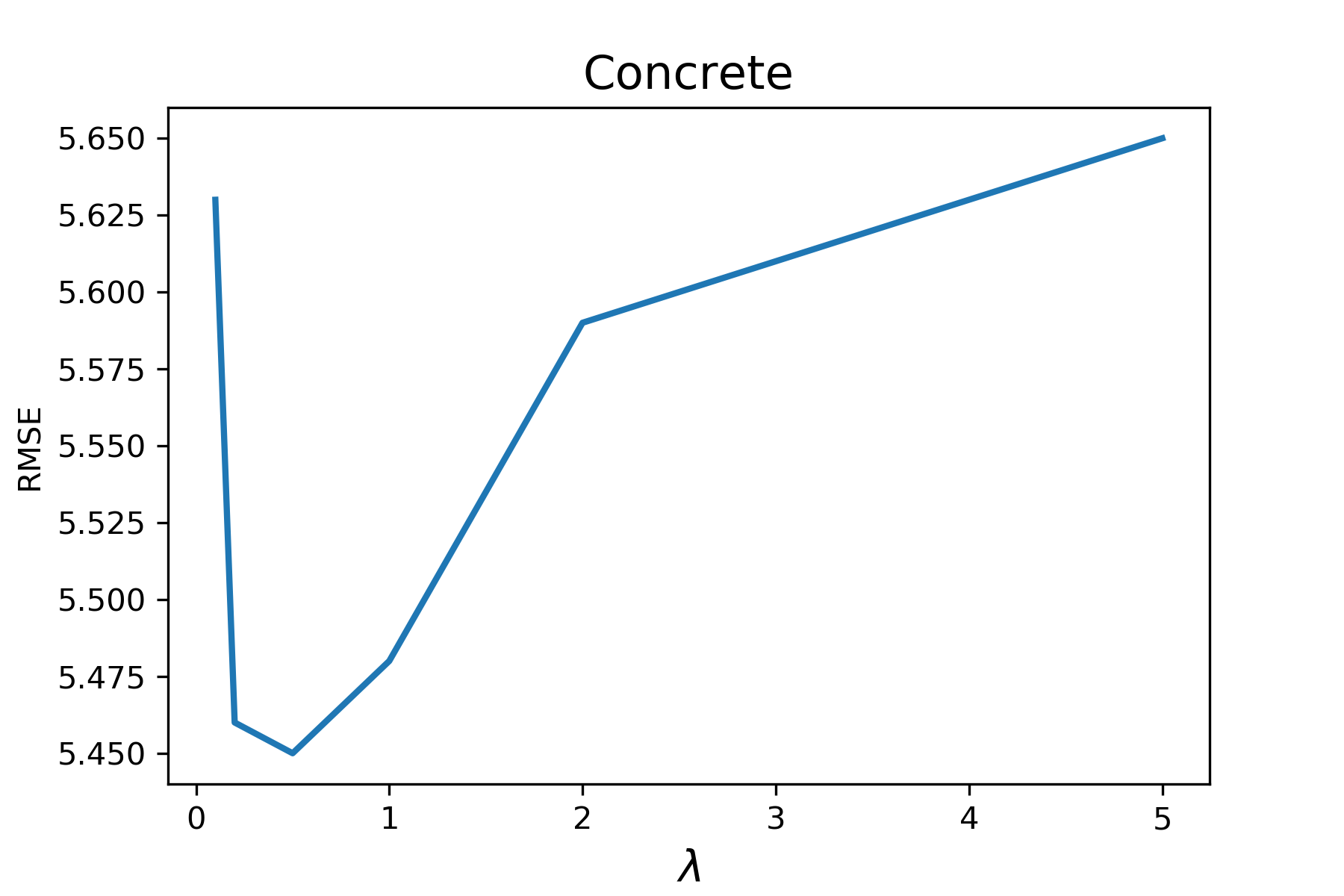}
	\end{minipage}
	\begin{minipage}{0.16\textwidth}
	   \includegraphics[width=\textwidth]{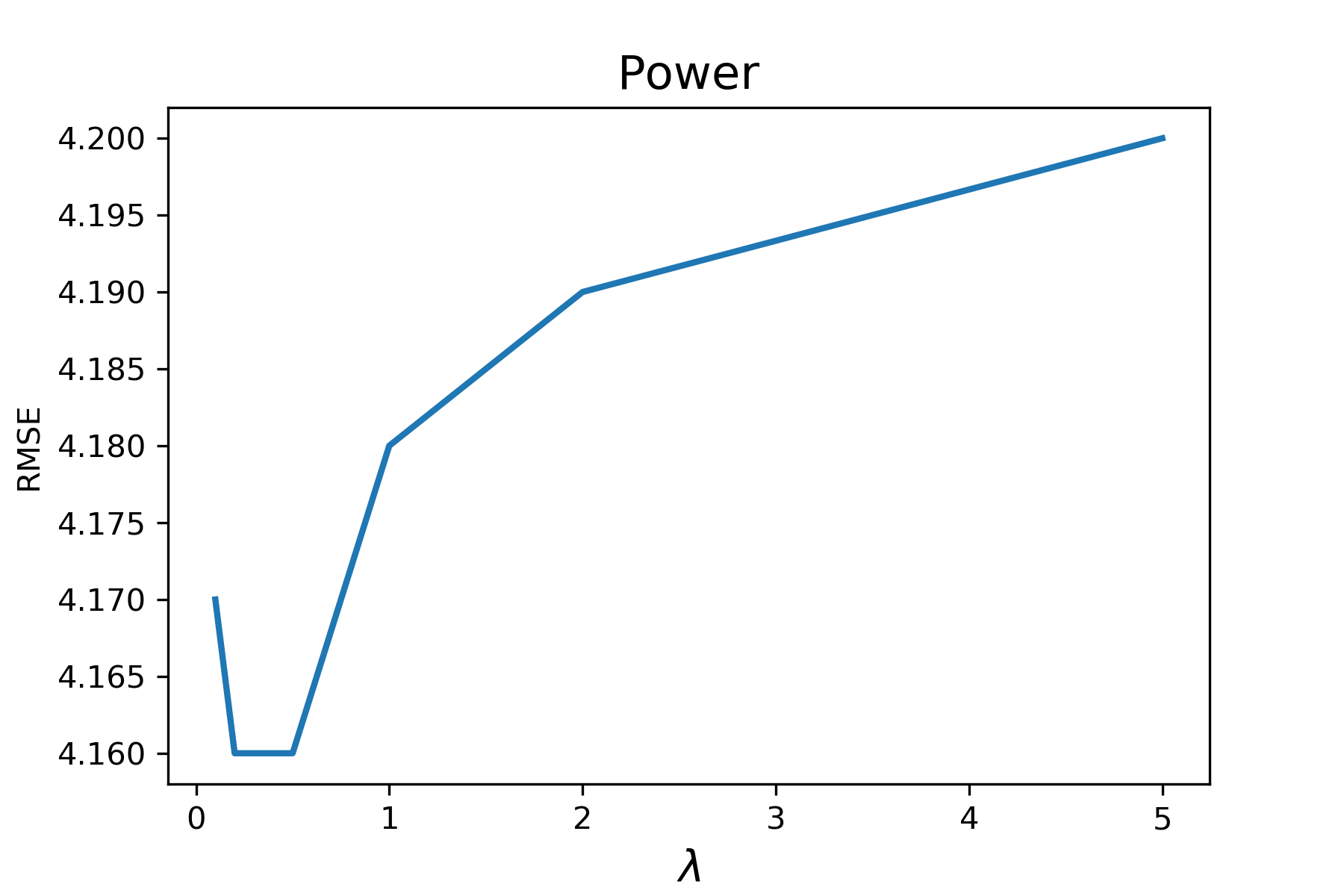}
	\end{minipage}
	\begin{minipage}{0.16\textwidth}
	   \includegraphics[width=\textwidth]{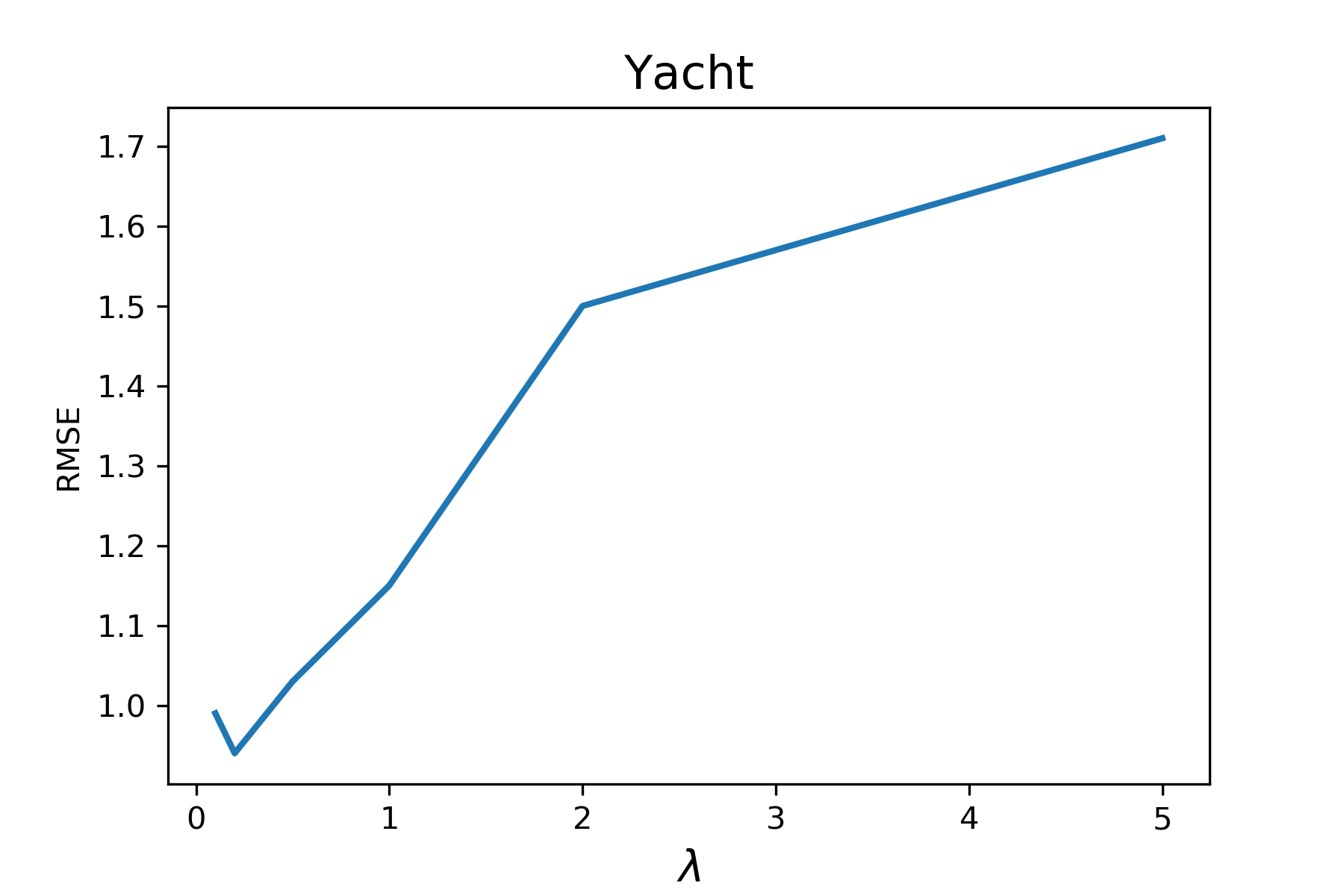}
	\end{minipage}
	\begin{minipage}{0.16\textwidth}
	   \includegraphics[width=\textwidth]{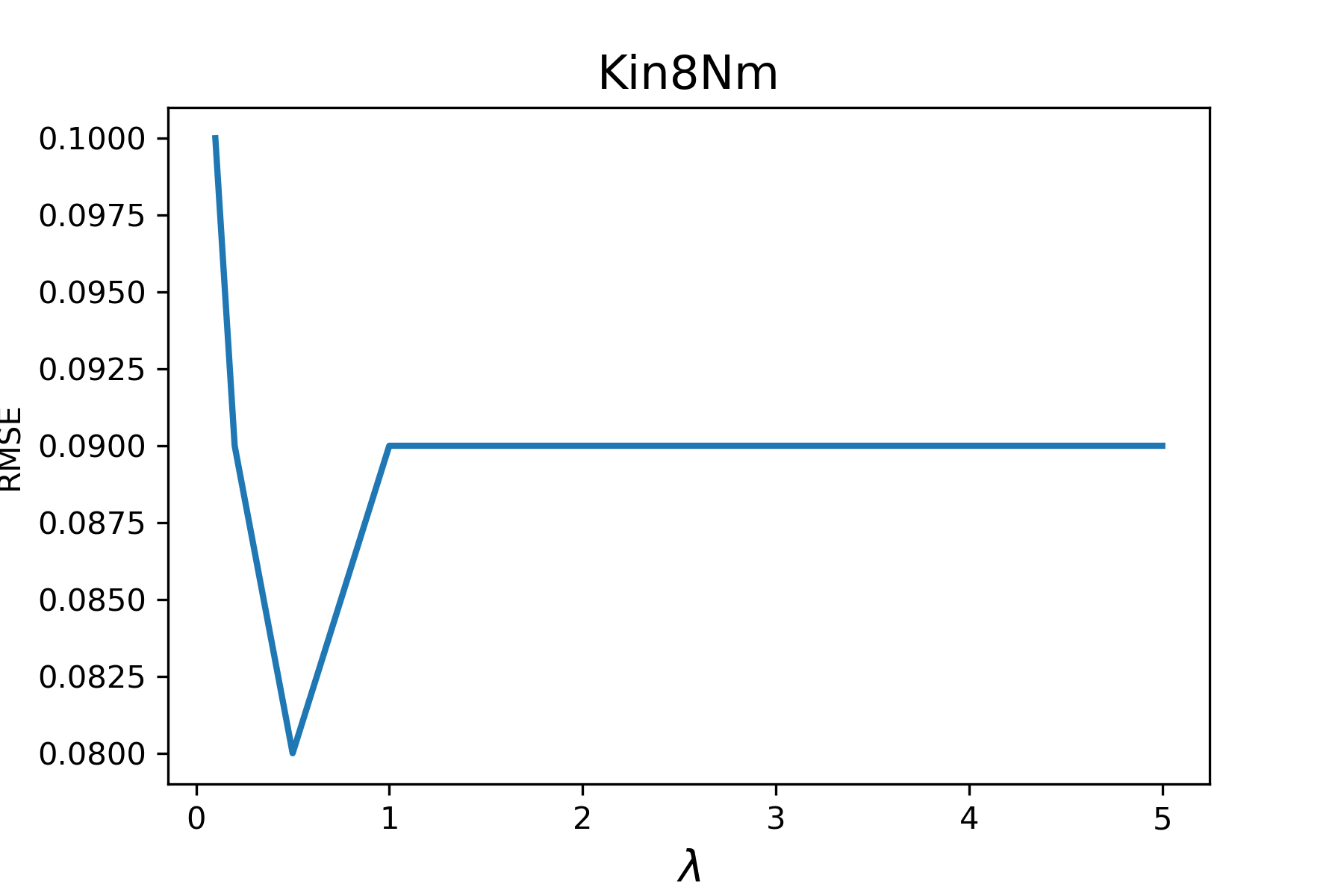}
	\end{minipage}
	\begin{minipage}{0.16\textwidth}
	   \includegraphics[width=\textwidth]{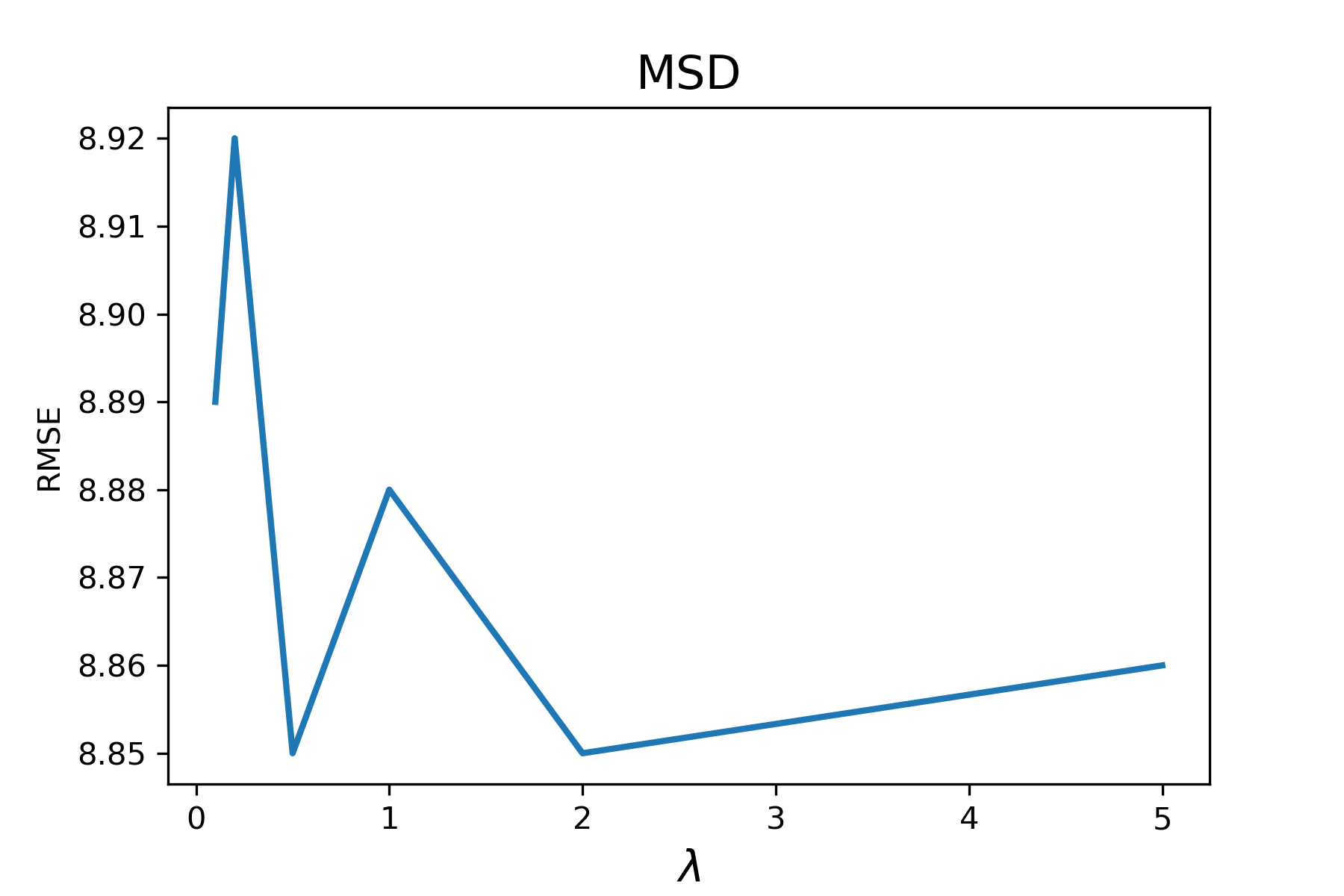}
	\end{minipage}

	\begin{minipage}{0.16\textwidth}
	   \includegraphics[width=\textwidth]{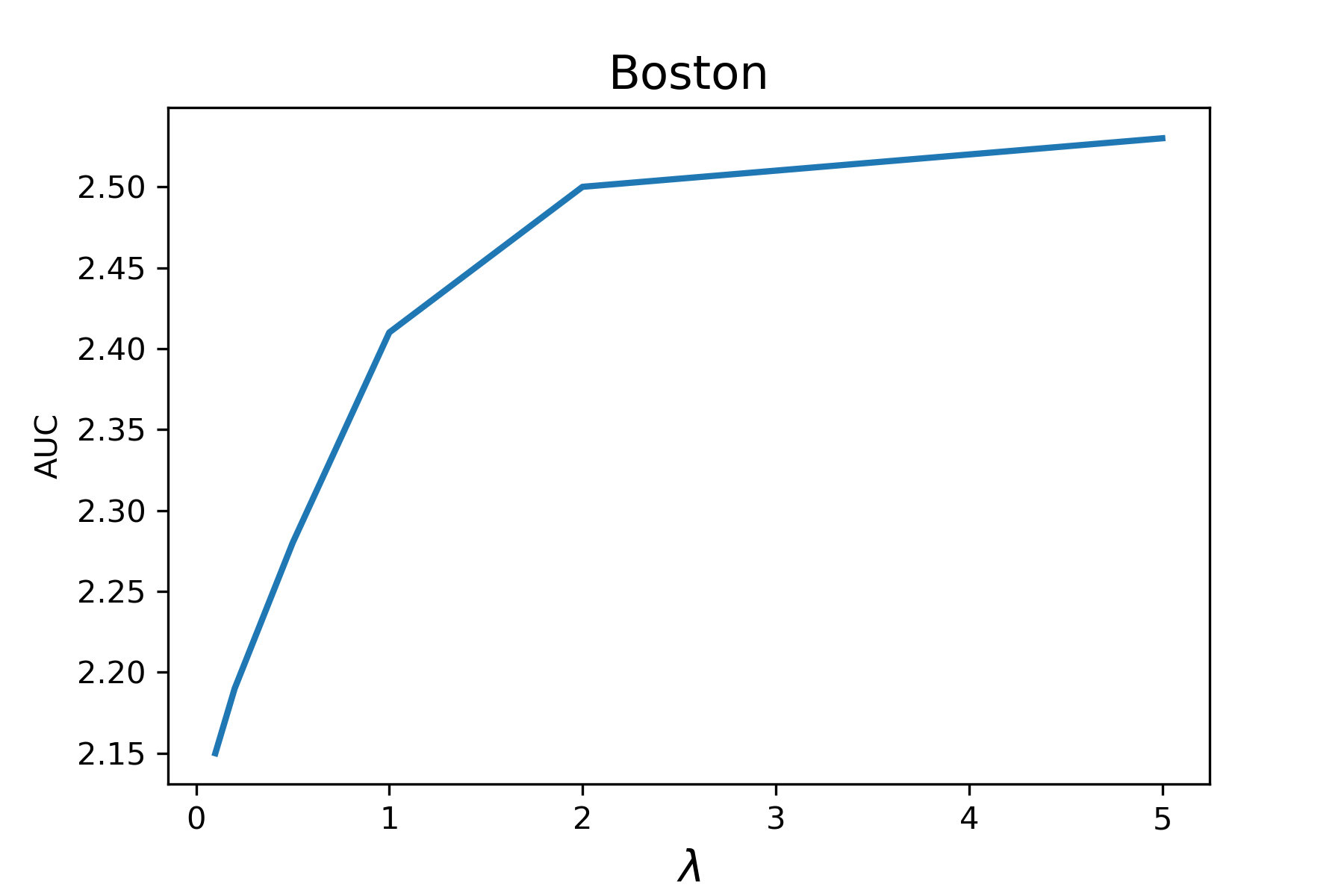}
	\end{minipage}
	\begin{minipage}{0.16\textwidth}
	   \includegraphics[width=\textwidth]{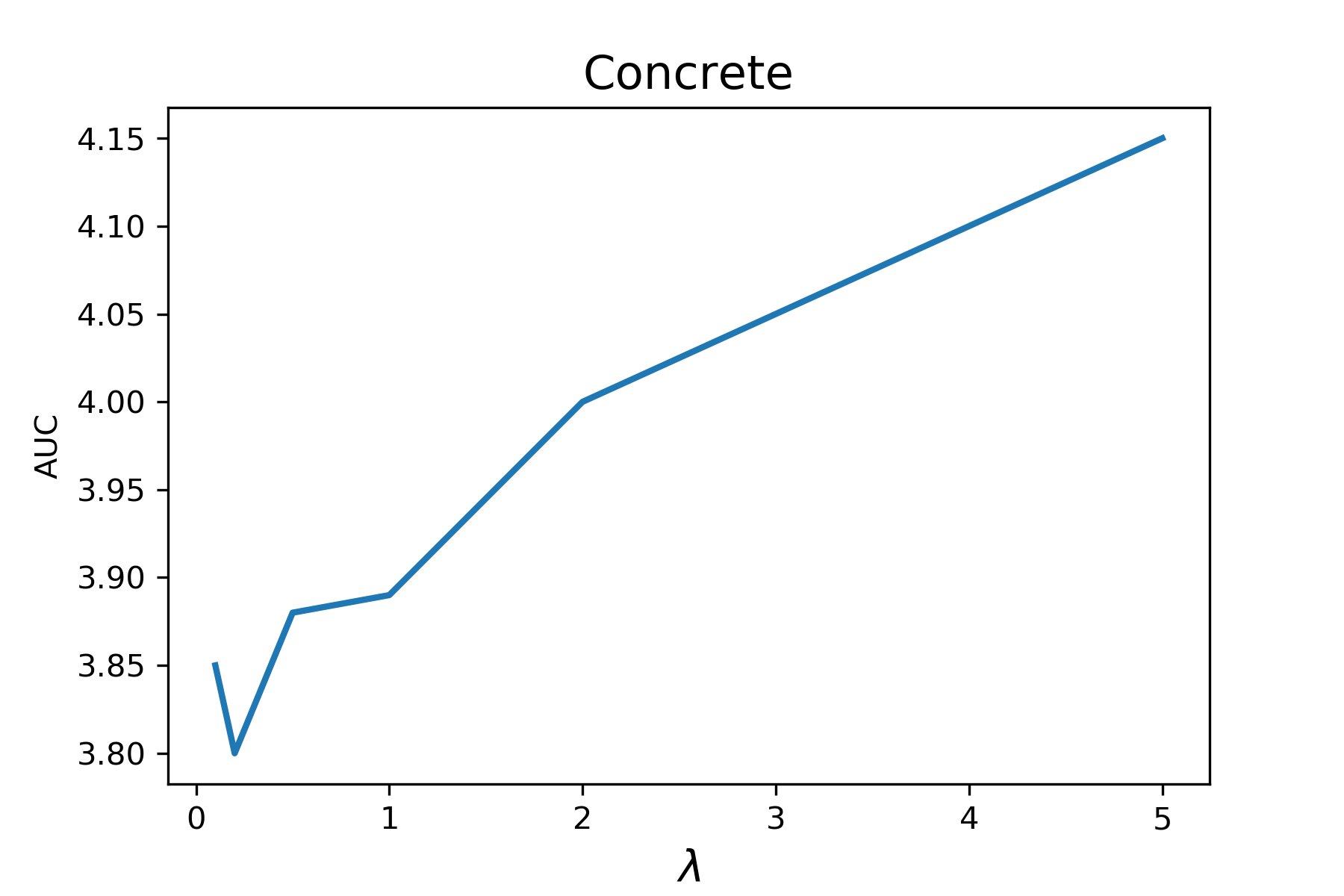}
	\end{minipage}
	\begin{minipage}{0.16\textwidth}
	   \includegraphics[width=\textwidth]{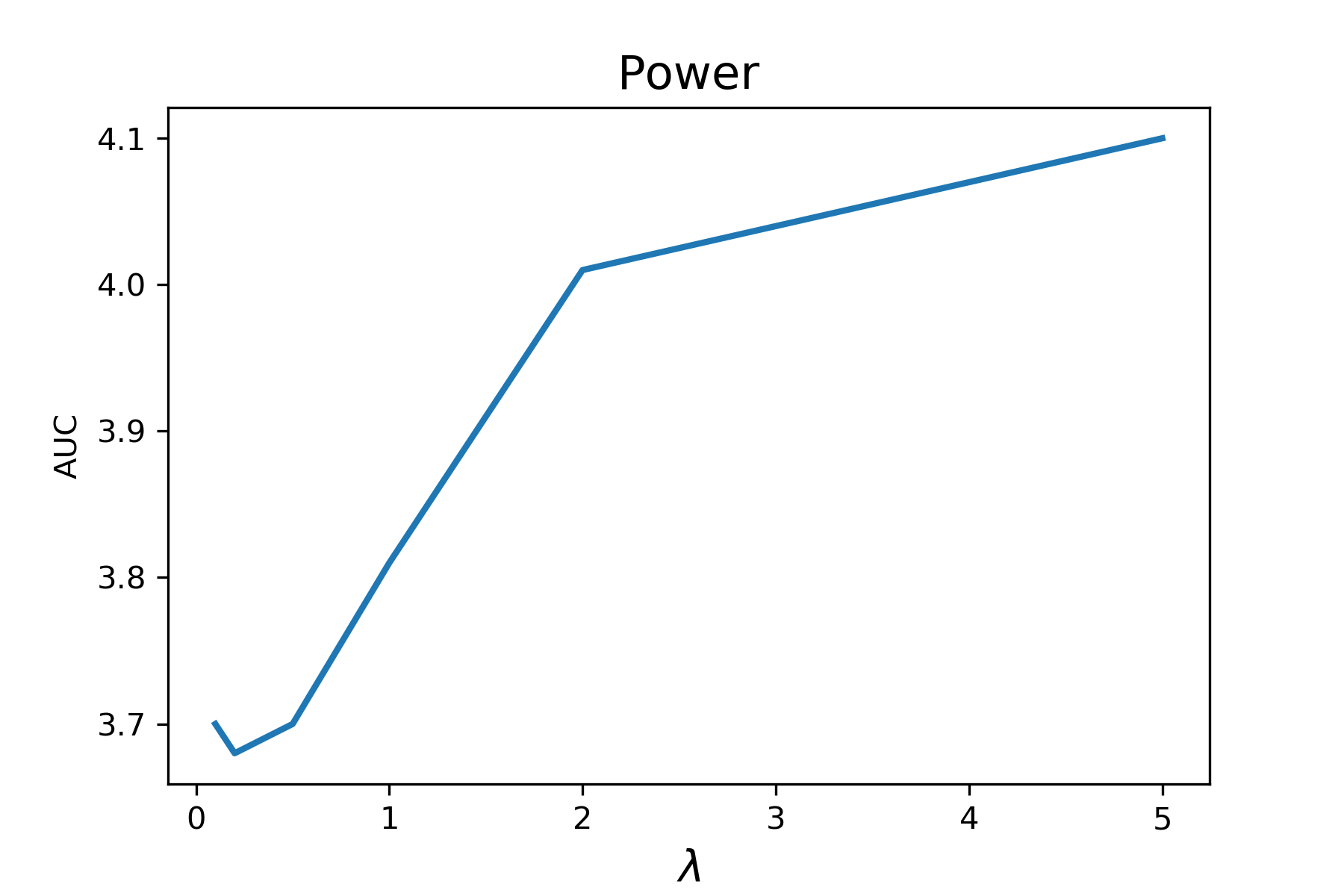}
	\end{minipage}
	\begin{minipage}{0.16\textwidth}
	   \includegraphics[width=\textwidth]{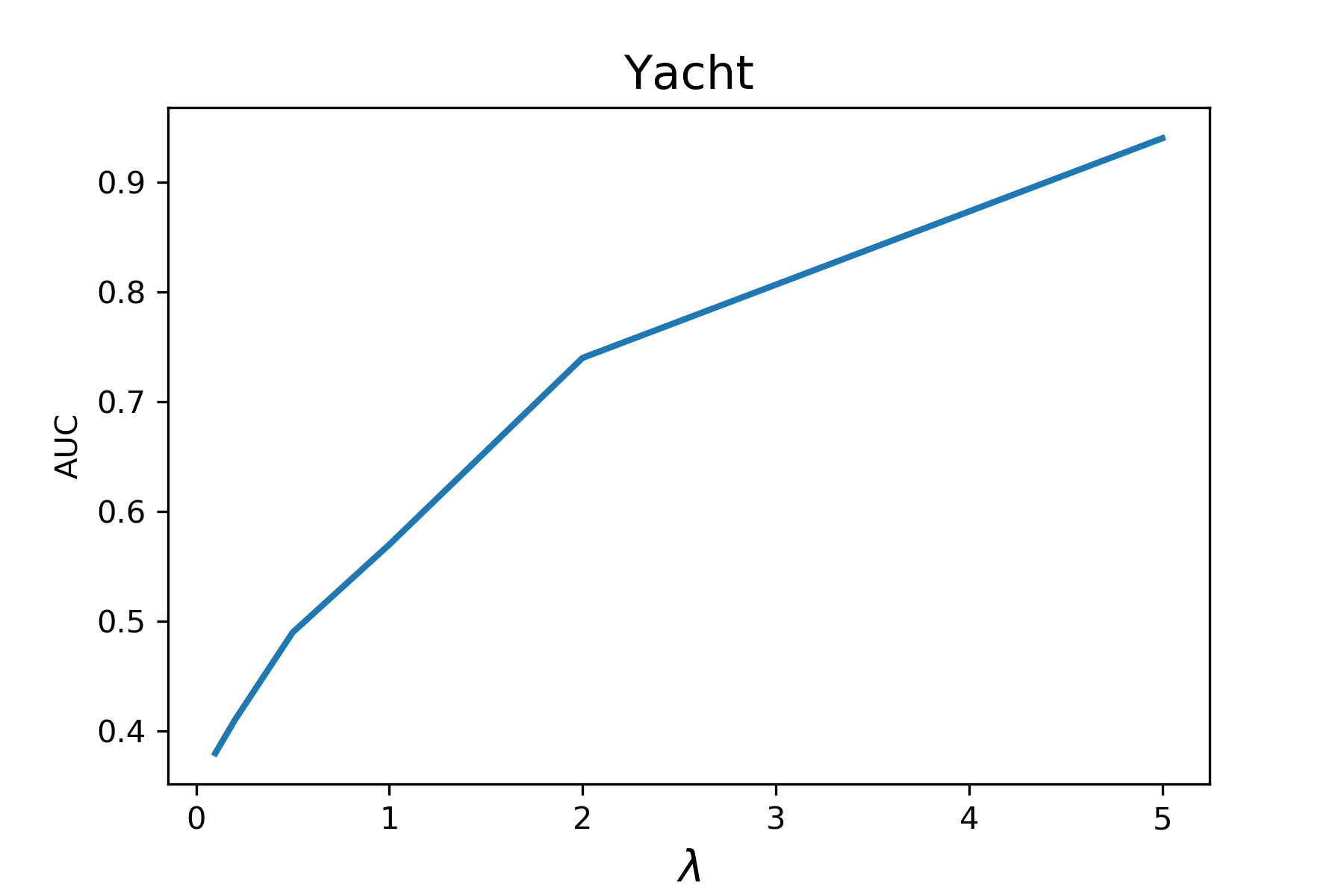}
	\end{minipage}
	\begin{minipage}{0.16\textwidth}
	   \includegraphics[width=\textwidth]{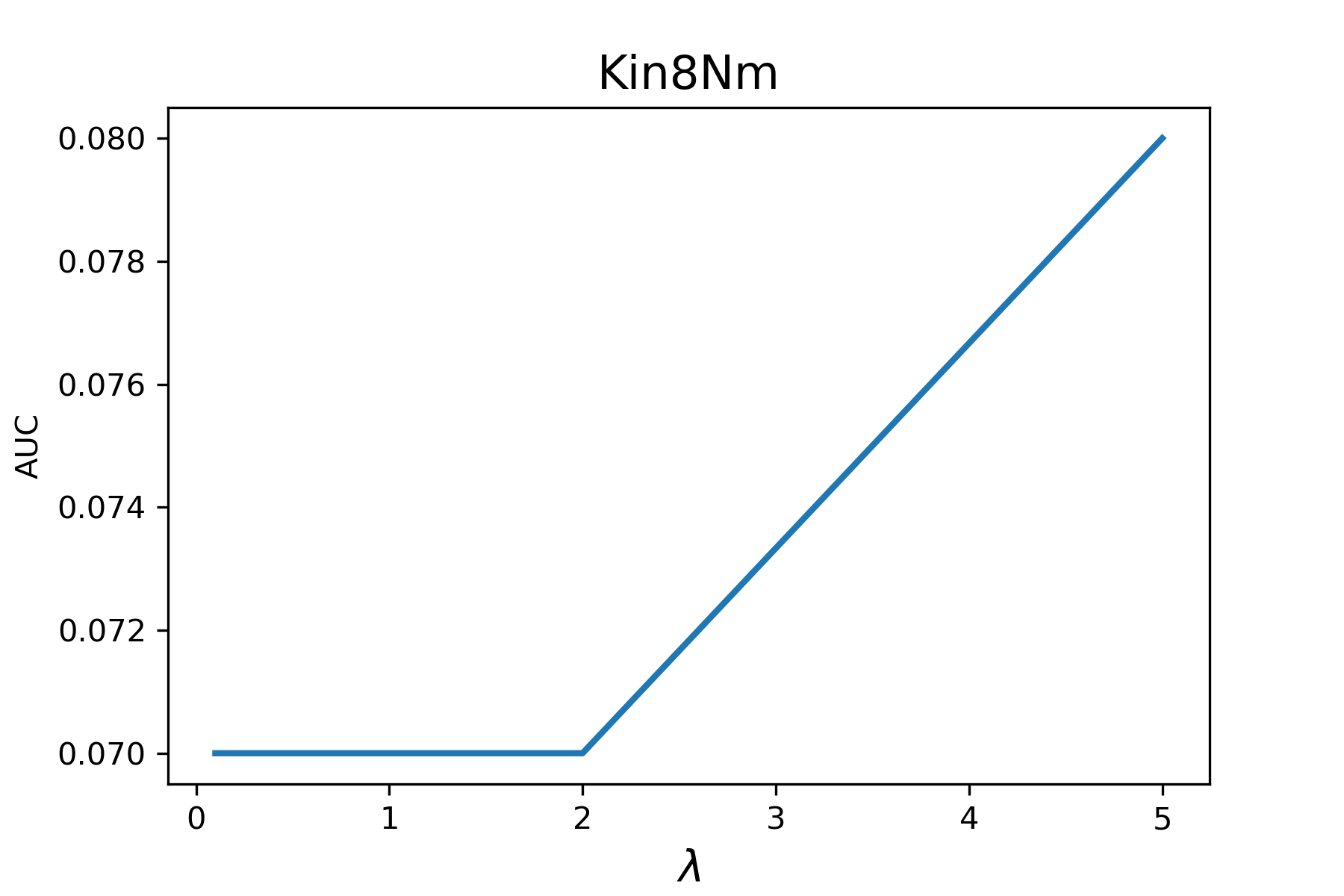}
	\end{minipage}
	\begin{minipage}{0.16\textwidth}
	   \includegraphics[width=\textwidth]{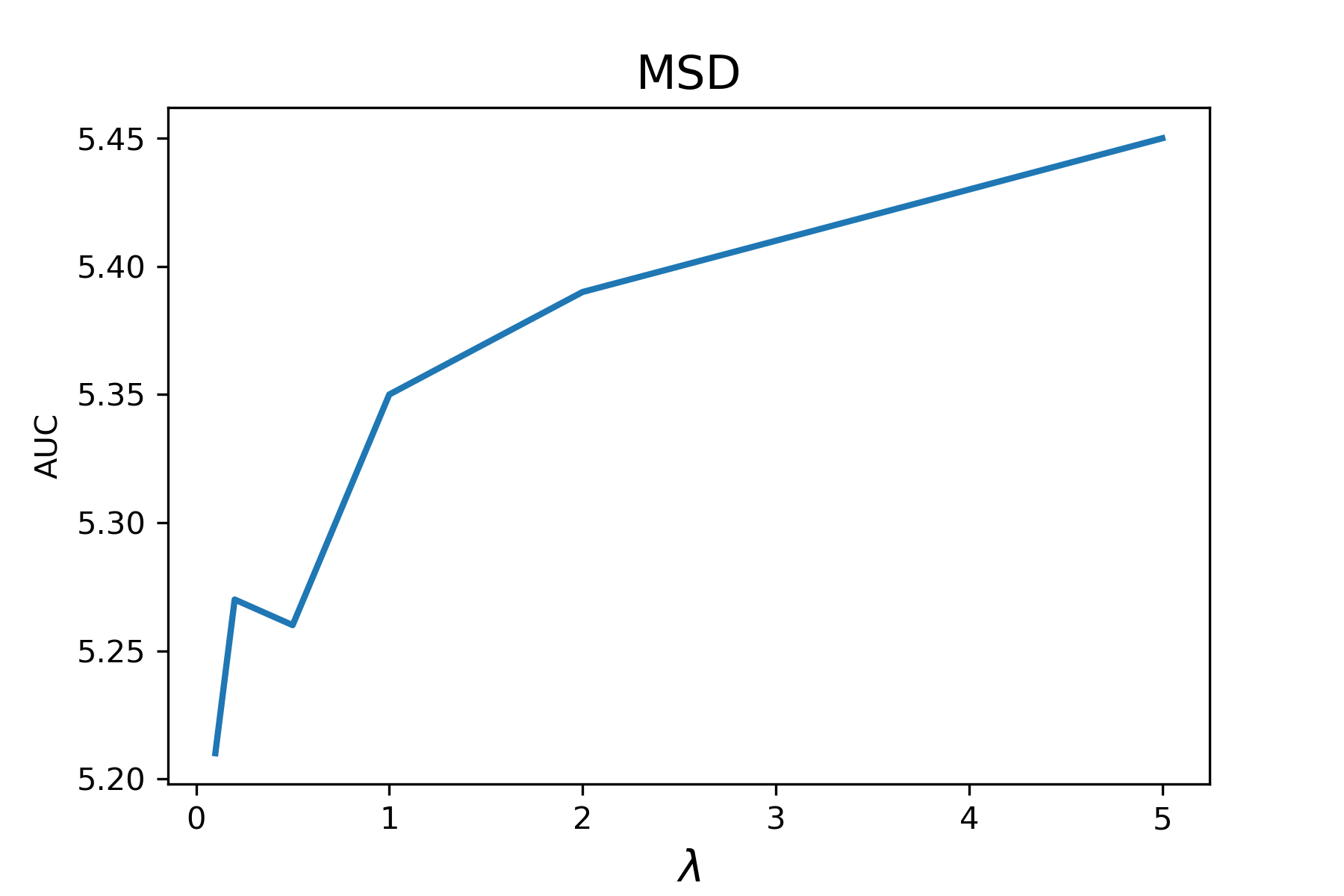}
    \end{minipage}
\caption{RMSE (upper row) and AUC (lower row) of our method with MSE regressor's loss versus $\lambda$  for different data sets.
}\label{figSqErrorMeasureVsLambda}
\end{figure}

\begin{figure}[!t]
	\begin{minipage}{0.32\textwidth}
	   \includegraphics[width=\textwidth]{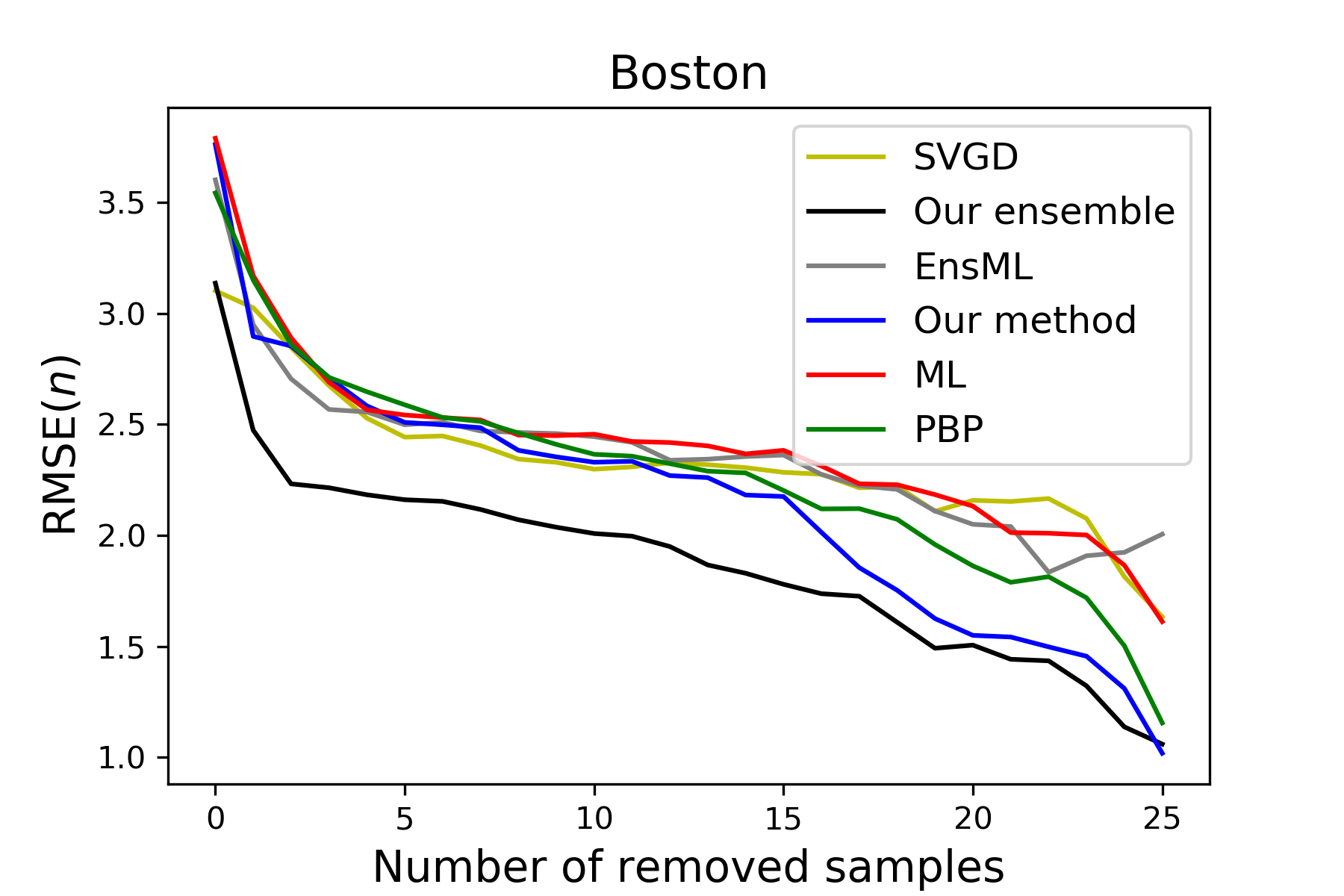}
	\end{minipage}
	\begin{minipage}{0.32\textwidth}
       \includegraphics[width=\textwidth]{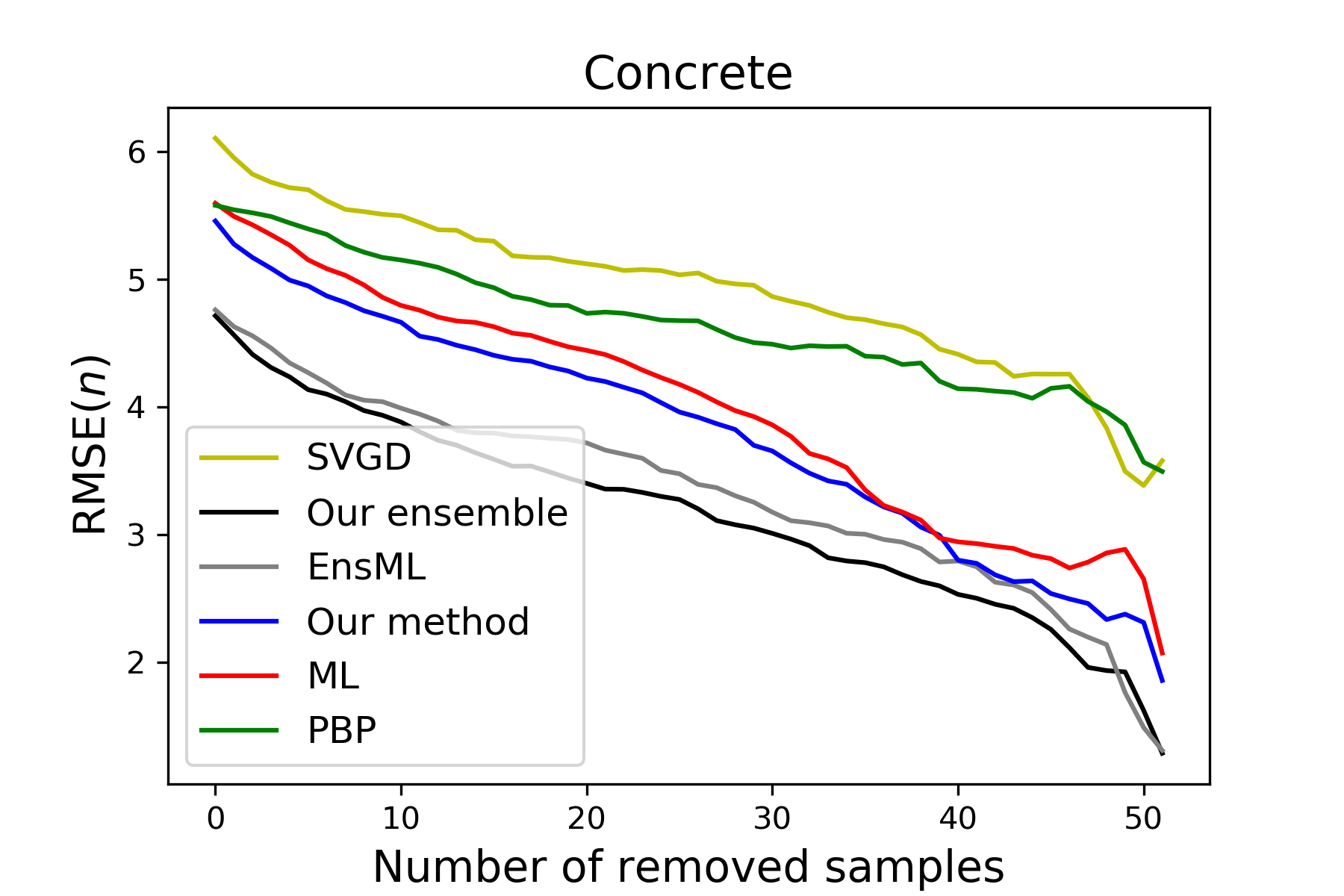}
	\end{minipage}
	\begin{minipage}{0.32\textwidth}
       \includegraphics[width=\textwidth]{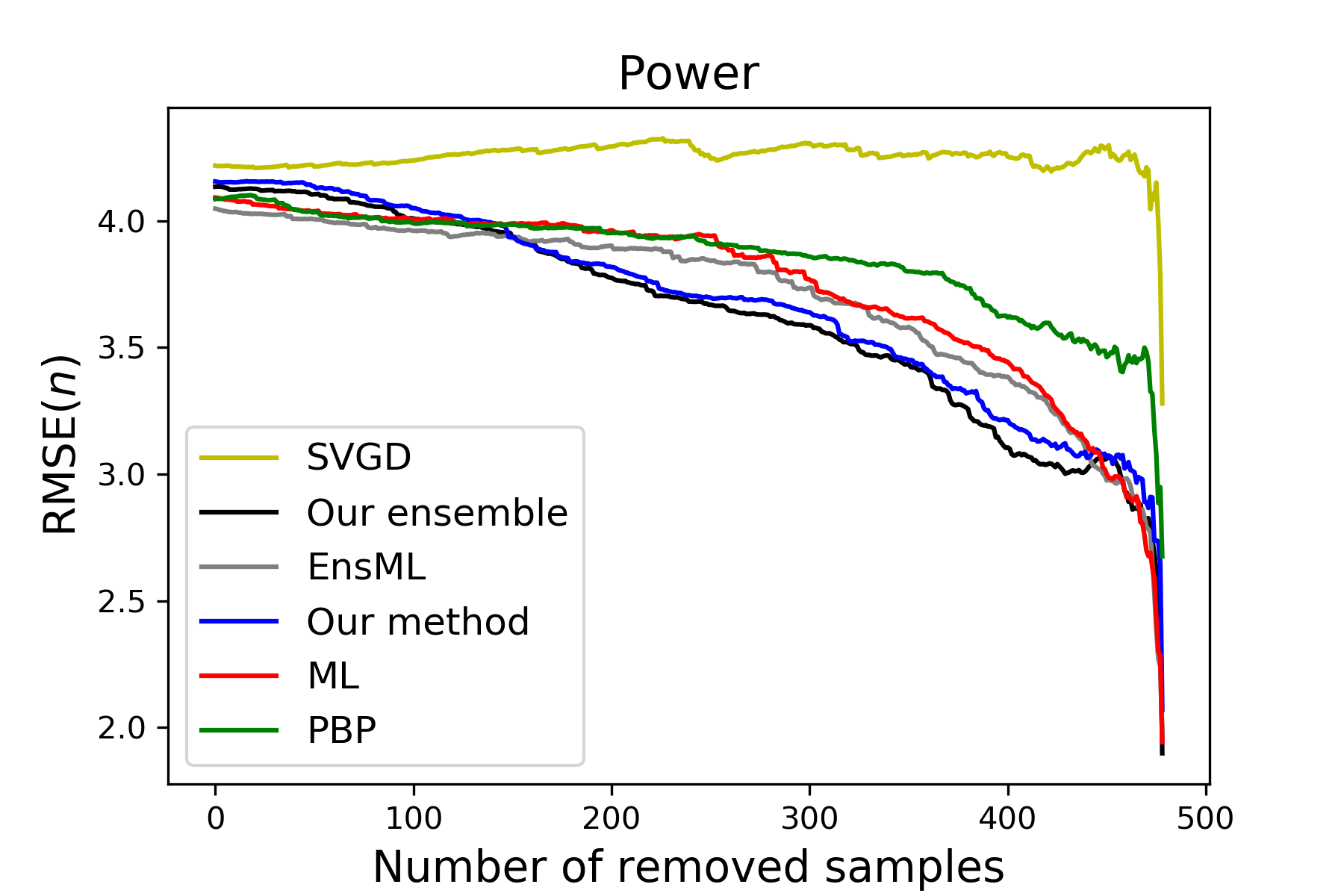}
	\end{minipage}

	\begin{minipage}{0.32\textwidth}
       \includegraphics[width=\textwidth]{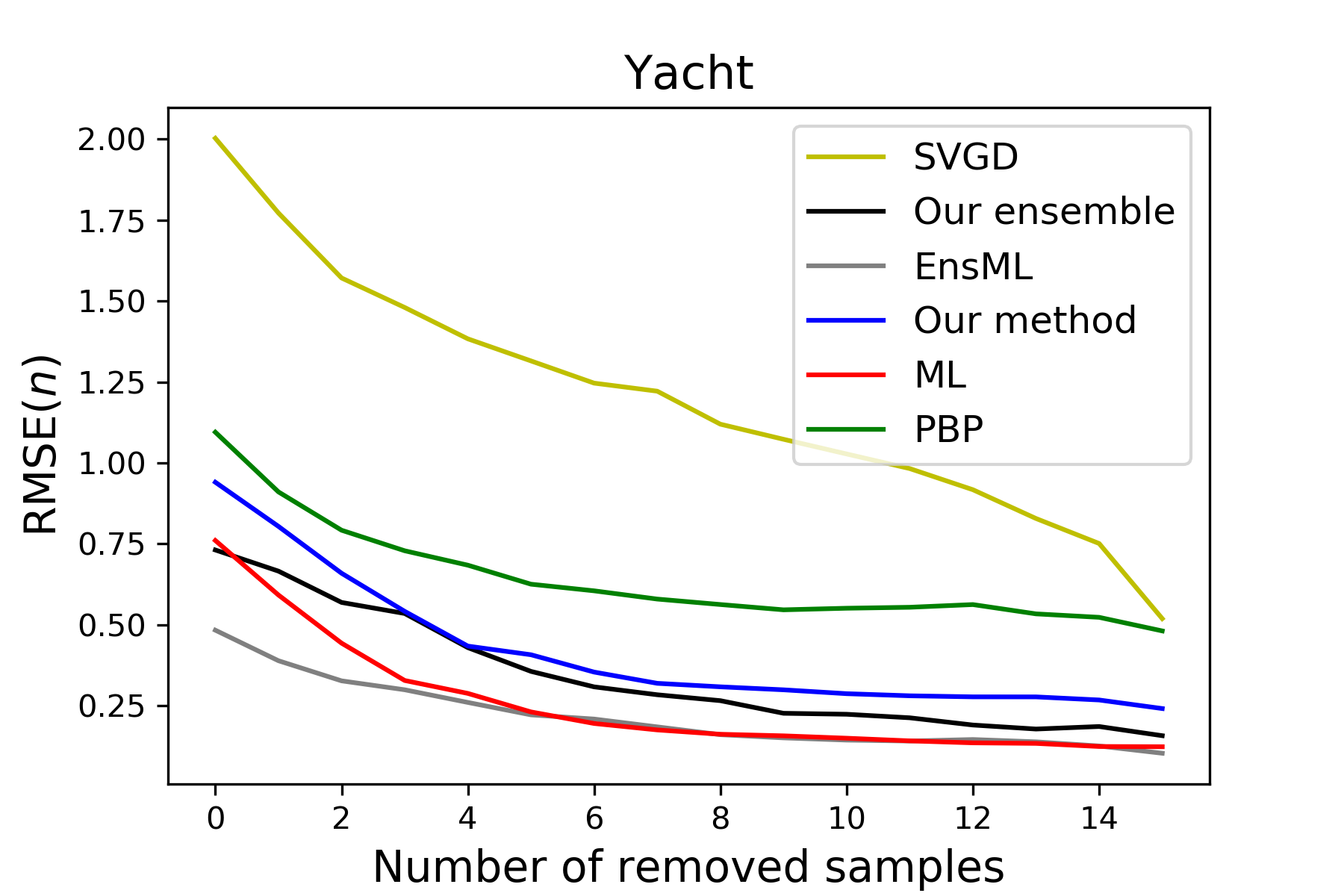}
	\end{minipage}
	\begin{minipage}{0.32\textwidth}
       \includegraphics[width=\textwidth]{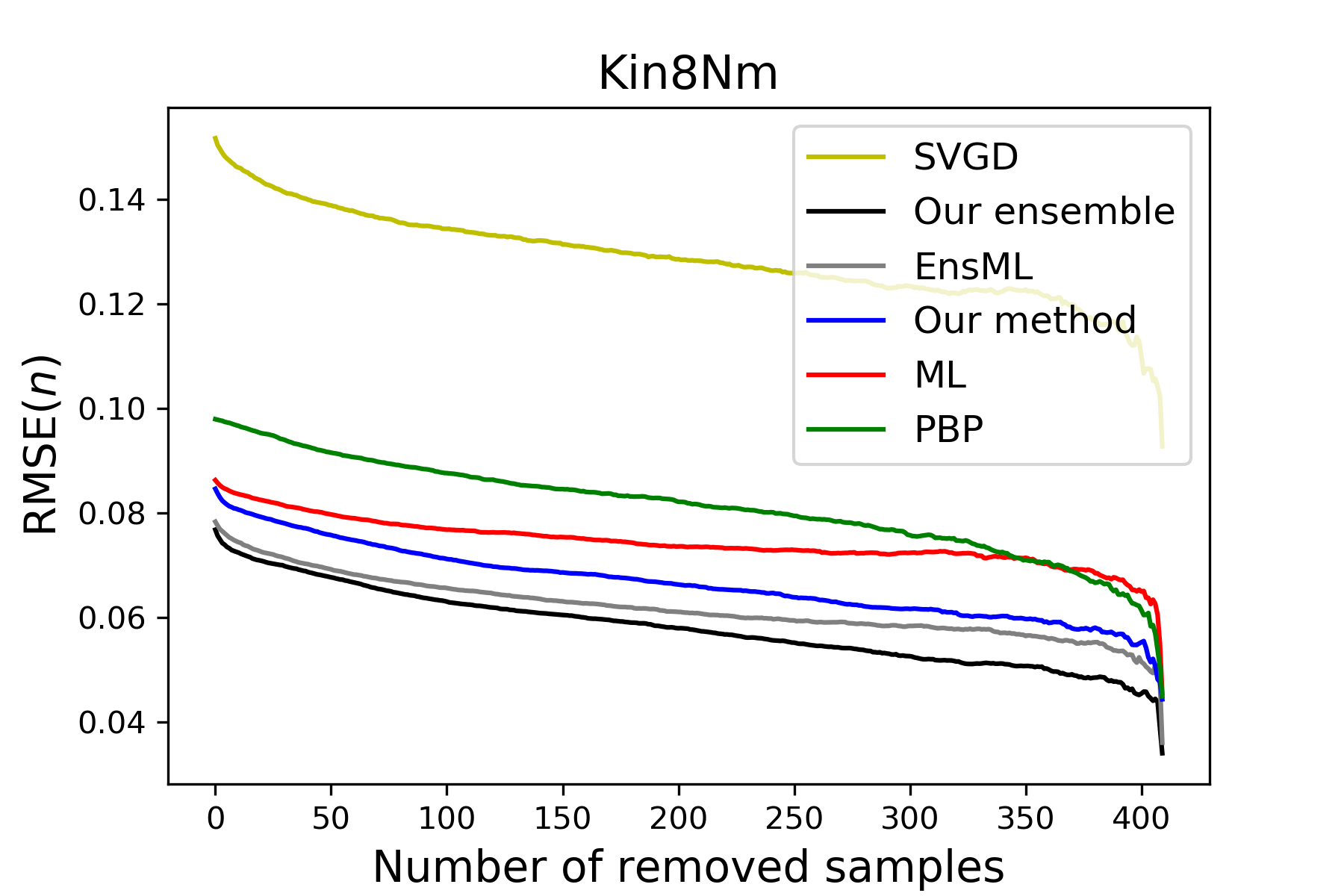}
	\end{minipage}
	\begin{minipage}{0.32\textwidth}
       \includegraphics[width=\textwidth]{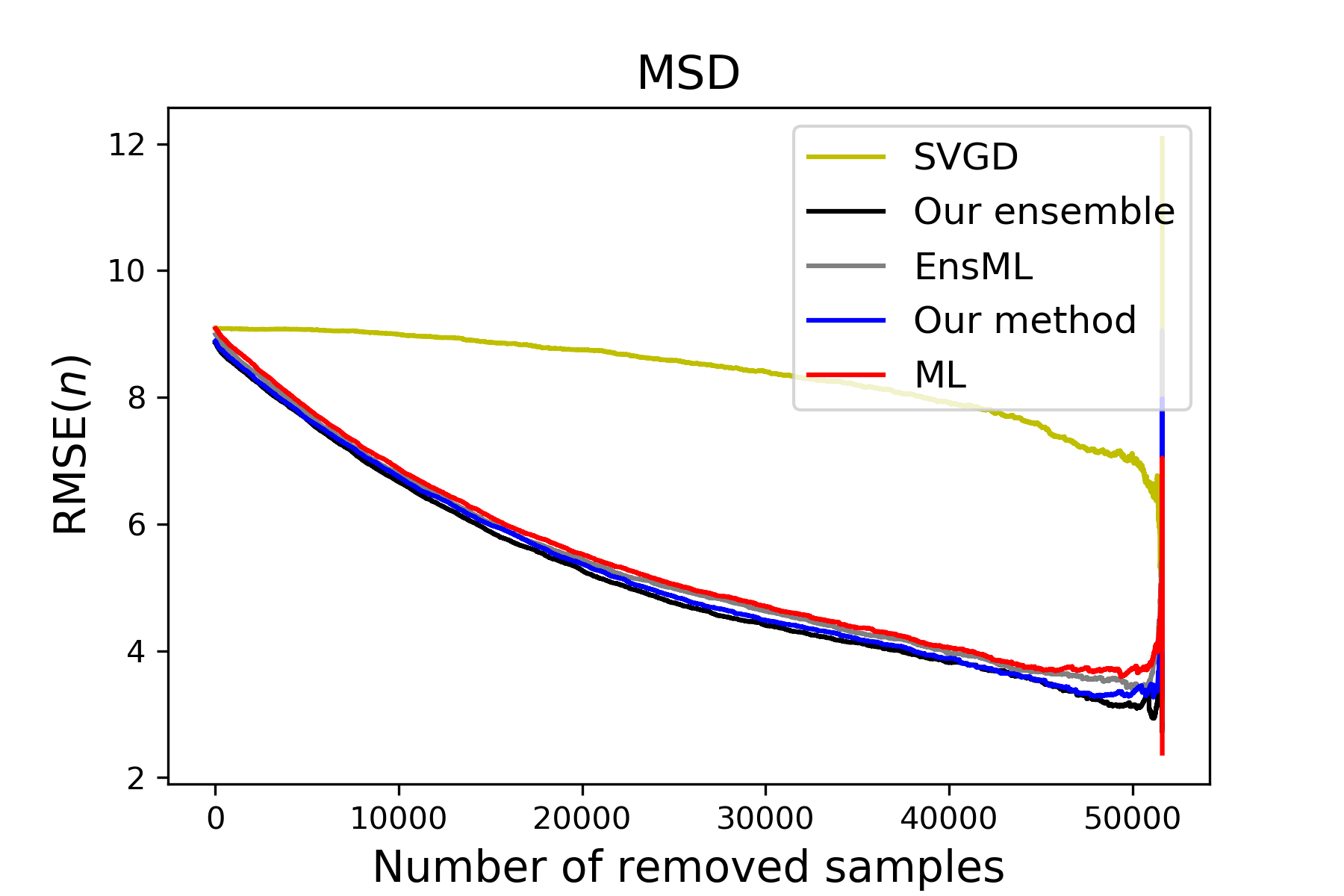}
	\end{minipage}
\caption{The curves ${\rm RMSE}(n)$ for the different methods.}\label{figSqErrorRemovedSampleCurves}
\end{figure}

\textbf{RMSE results.} Figure~\ref{figSqErrorMeasureVsLambda} shows the dependence of RMSE and AUC in our method on~$\lambda$ for different data sets.
We see that typically our method performs better when $\lambda<1$, which is due to a better fit on clean regions (see Sec.~\ref{secSigmoidSoftReLambda}).
Tables~\ref{tableSqErrorBostonConcrete1} and~\ref{tableSqErrorBostonConcrete2}  show RMSE and AUC for the different methods\footnote{We were not able to fit the PBP for the largest MSD data set.}.
The values with the best mean and the values that are not significantly different from those with the best mean (due to the two-tailed paired difference test with $p=0.05$) are marked in bold.
\begin{table}
\resizebox{\textwidth}{!}{%
\begin{tabular}{lccc}
{} & {\bf Boston} & {}\\
\toprule
{} & RMSE & AUC  \\
\midrule
ML                      &    {3.79$\pm$1.44} &         2.40$\pm$0.59  \\
SVGD                    &     {\bf 3.10$\pm$1.14}         &       2.34$\pm$0.54 \\
PBP                     &    {\bf 3.54$\pm$1.29} &         2.28$\pm$0.47  \\
Our method &    {3.76$\pm$1.50} &         {2.15$\pm$0.51}  \\
\bottomrule
\bottomrule
EnsML                   &    {\bf 3.60$\pm$1.36} &         2.35$\pm$0.49  \\
Our ensemble            &    {\bf 3.14$\pm$1.31} &         {\bf 1.86$\pm$0.46}  \\
\end{tabular}
\
\begin{tabular}{lcc}
{\bf Concrete} & {} \\
\toprule
 RMSE & AUC  \\
\midrule
    5.60$\pm$0.63 &         4.00$\pm$0.59  \\
    6.11$\pm$0.59        &         4.93$\pm$0.86 \\
    5.58$\pm$0.60 &         4.66$\pm$0.64  \\
   5.46$\pm$0.63 &         {3.80$\pm$0.60}  \\
\bottomrule
\bottomrule
    {\bf 4.76$\pm$0.64} &  {\bf 3.34$\pm$0.44}  \\
    {\bf 4.71$\pm$0.65} &  {\bf 3.17$\pm$0.38}  \\
\end{tabular}
\
\begin{tabular}{lcc}
{\bf Power} & {}\\
\toprule
RMSE & AUC  \\
\midrule
    {4.09$\pm$0.31} &       3.75$\pm$0.24     \\
     4.22$\pm$0.30         &         4.26$\pm$0.28 \\
    {\bf 4.09$\pm$0.26} &       3.86$\pm$0.26     \\
    {4.13$\pm$0.30} &       {\bf 3.63$\pm$0.24}    \\
\bottomrule
\bottomrule
    {\bf 4.05$\pm$0.31} &  {\bf 3.71$\pm$0.24}  \\
    {\bf 4.13$\pm$0.30} &  {\bf 3.64$\pm$0.26}  \\
\end{tabular}
}
\caption{Values of RMSE and AUC for the different methods on the Boston, Concrete, and Power data sets.}\label{tableSqErrorBostonConcrete1}
\end{table}
\begin{table}
\resizebox{\textwidth}{!}{%
\begin{tabular}{lccc}
{} & {\bf Yacht} & {}\\
\toprule
{} & RMSE & AUC  \\
\midrule
ML                      &  0.76$\pm$0.38    &      0.25$\pm$0.08      \\
SVGD                     &  2.00$\pm$0.72              & 1.20$\pm$0.49 \\
PBP                     &  1.09$\pm$0.35   &       $0.64\pm$0.19    \\
Our method              &  0.94$\pm$0.42    &    0.41$\pm$0.13        \\
\bottomrule
\bottomrule
EnsML                   &    {\bf 0.48$\pm$0.24} &         {\bf 0.21$\pm$0.07} \\
Our ensemble            &    0.73$\pm$0.33 &         0.34$\pm$0.15 \\
\end{tabular}
\
\begin{tabular}{lccc}
{\bf Kin8nm} & {}\\
\toprule
RMSE & AUC  \\
\midrule
     0.086$\pm$0.006 &  0.074$\pm$0.004          \\
     0.152$\pm$0.007         &   0.129$\pm$0.008 \\
     0.098$\pm$0.005 &  0.081$\pm$0.005         \\
     0.085$\pm$0.005 &   0.067$\pm$0.003        \\
\bottomrule
\bottomrule
   0.078$\pm$0.004  &     0.062$\pm$0.003       \\
   {\bf 0.077$\pm$0.004}  &     {\bf 0.058$\pm$0.003}       \\
\end{tabular}
\
\begin{tabular}{lcc}
{\bf MSD} & {}  \\
\toprule
RMSE & AUC  \\
\midrule
    9.09$\pm$NA &         5.40$\pm$NA  \\
    9.09$\pm$NA            &     8.38$\pm$NA \\
    NA                    &         NA            \\
    {8.87$\pm$NA} &         {5.23$\pm$NA}  \\
\bottomrule
\bottomrule
    8.99$\pm$NA         &         5.32$\pm$NA            \\
    {\bf 8.86$\pm$NA}           &         {\bf 5.13$\pm$NA}            \\
\end{tabular}
}
\caption{Values of RMSE and AUC for the different methods on the Yacht, Kin8nm, and MSD data sets.}\label{tableSqErrorBostonConcrete2}
\end{table}
Our ensemble achieves the best or not significantly different from the best RMSE and AUC on all the data sets (except Yacht, where it is second best after the EnsML). Further, note that our non-ensemble networks are the best among the other non-ensemble methods, again except for the Yacht data set. Figure~\ref{figSqErrorRemovedSampleCurves} shows the curves ${\rm RMSE}(n)$ for the different methods and data sets.
The curve corresponding to our ensemble is typically located below the others.


\textbf{MAE results.}

\begin{figure}[!t]
	\begin{minipage}{0.32\textwidth}
	   \includegraphics[width=\textwidth]{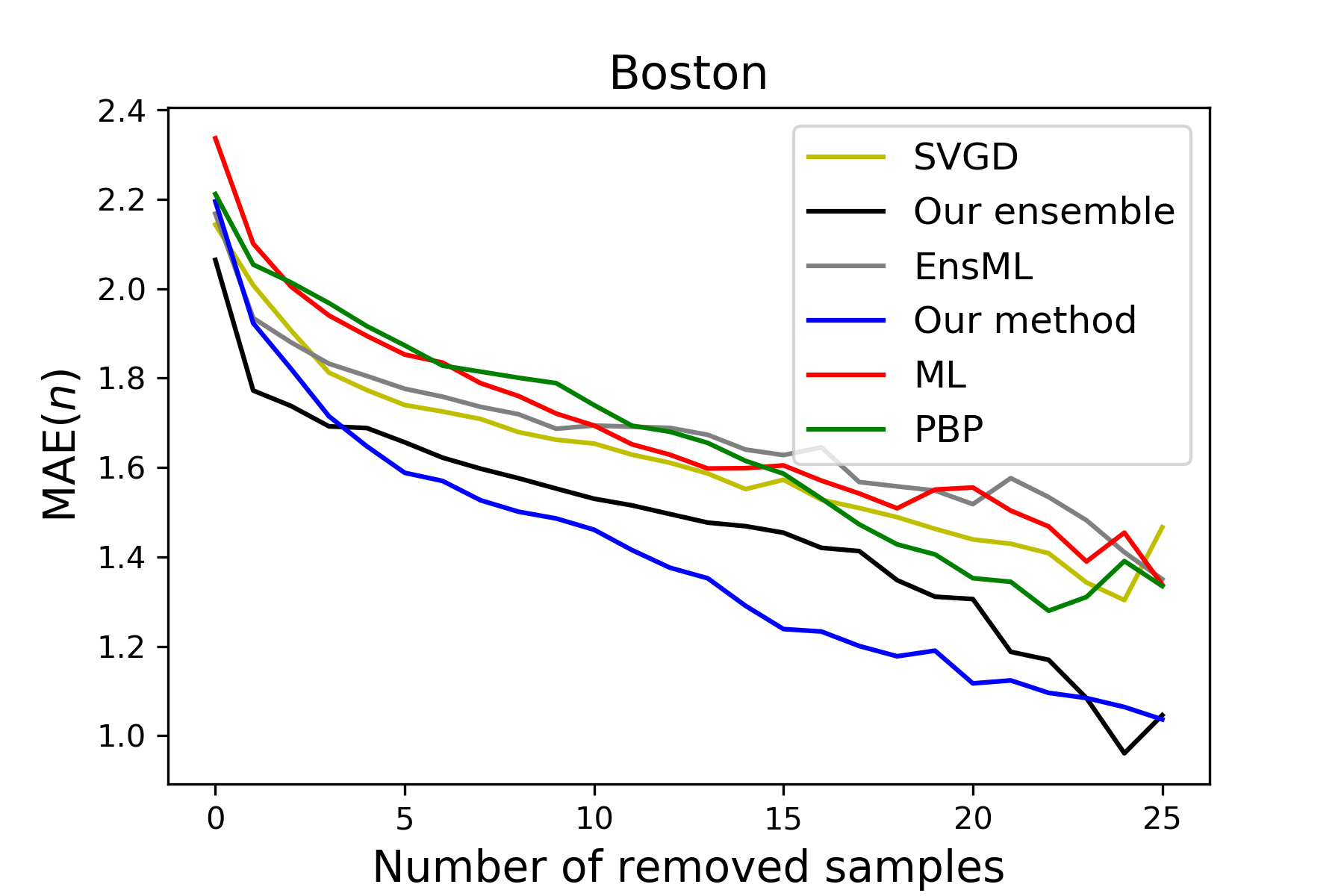}
	\end{minipage}
	\begin{minipage}{0.32\textwidth}
       \includegraphics[width=\textwidth]{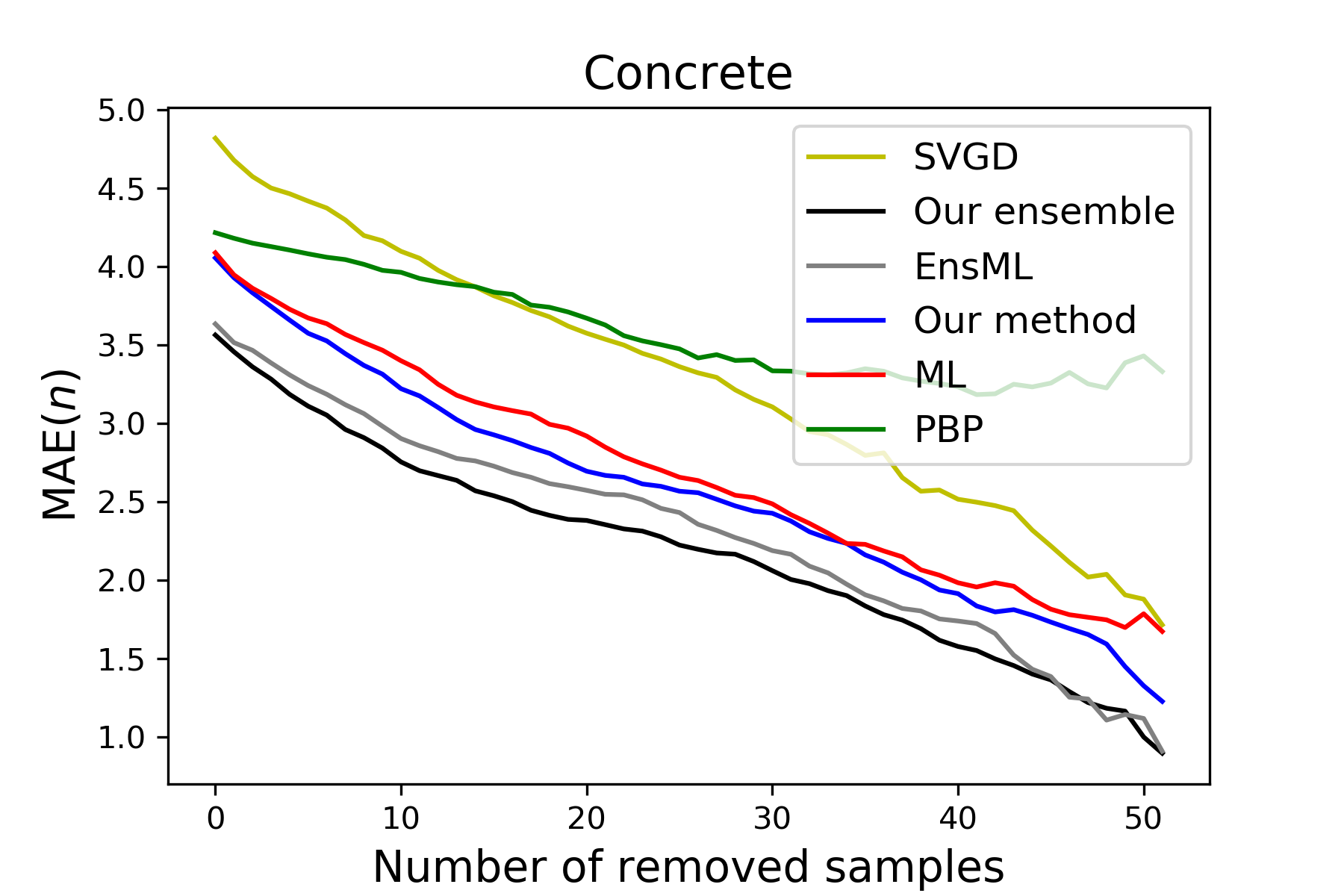}
	\end{minipage}
	\begin{minipage}{0.32\textwidth}
       \includegraphics[width=\textwidth]{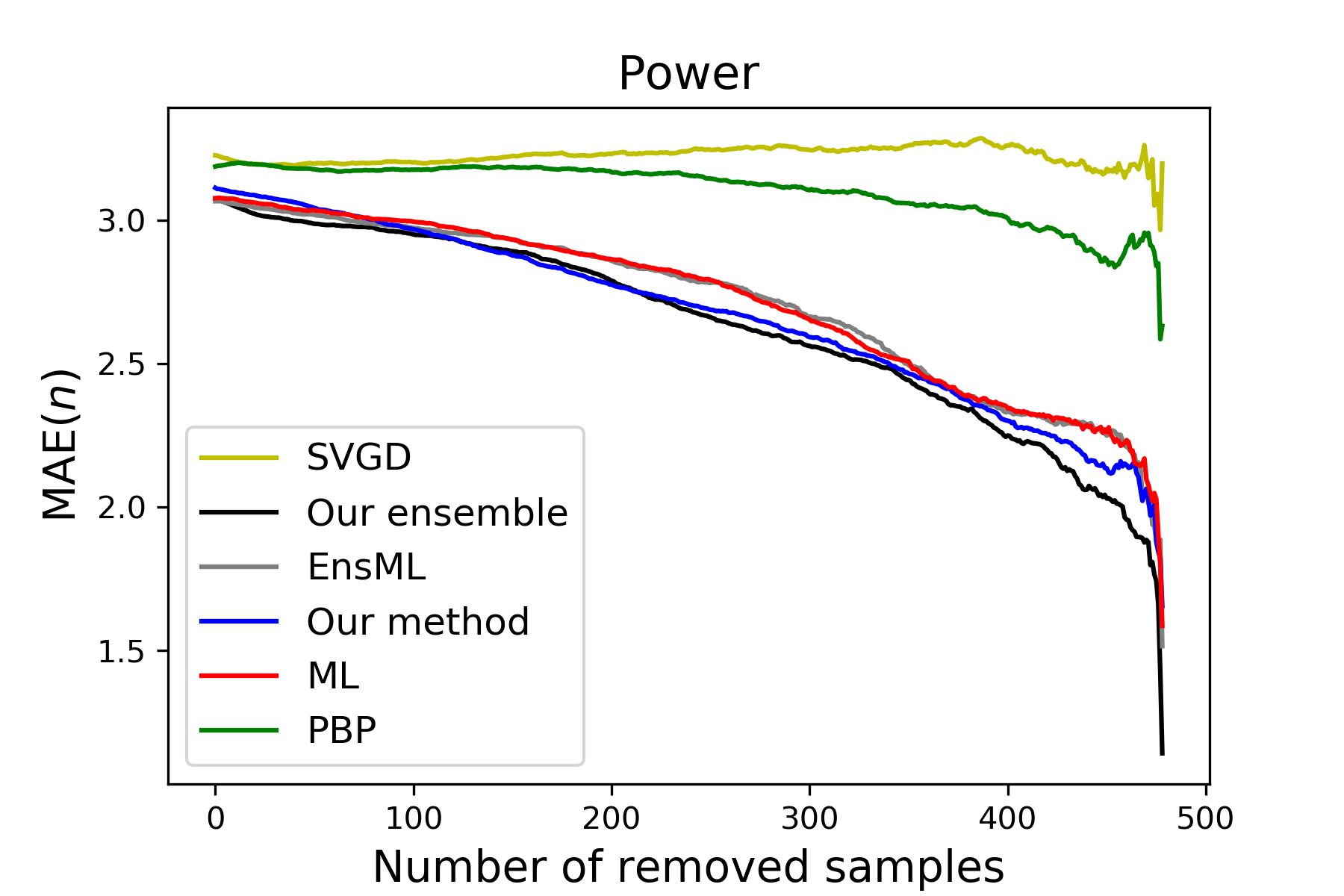}
	\end{minipage}

	\begin{minipage}{0.32\textwidth}
       \includegraphics[width=\textwidth]{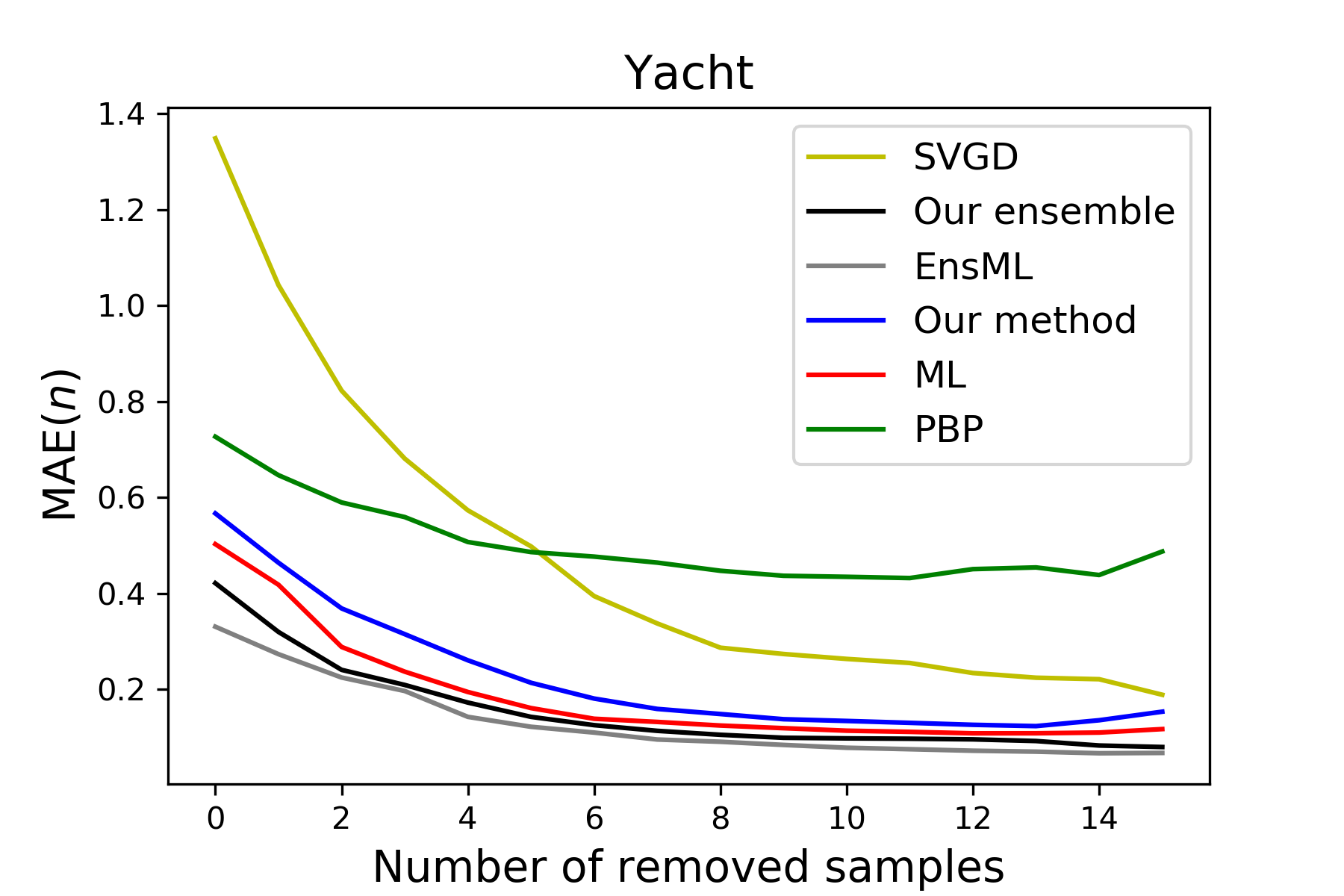}
	\end{minipage}
	\begin{minipage}{0.32\textwidth}
       \includegraphics[width=\textwidth]{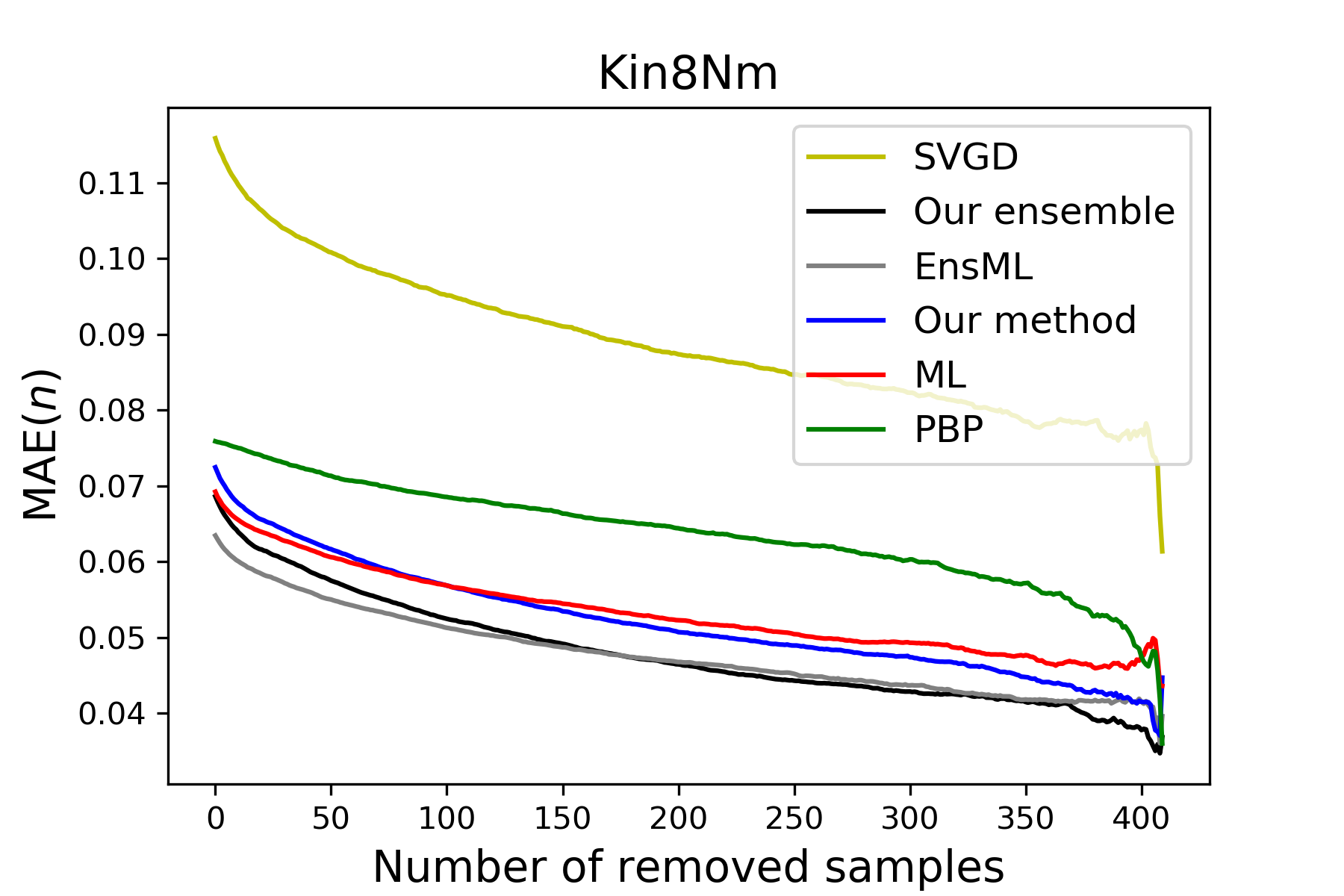}
	\end{minipage}
	\begin{minipage}{0.32\textwidth}
       \includegraphics[width=\textwidth]{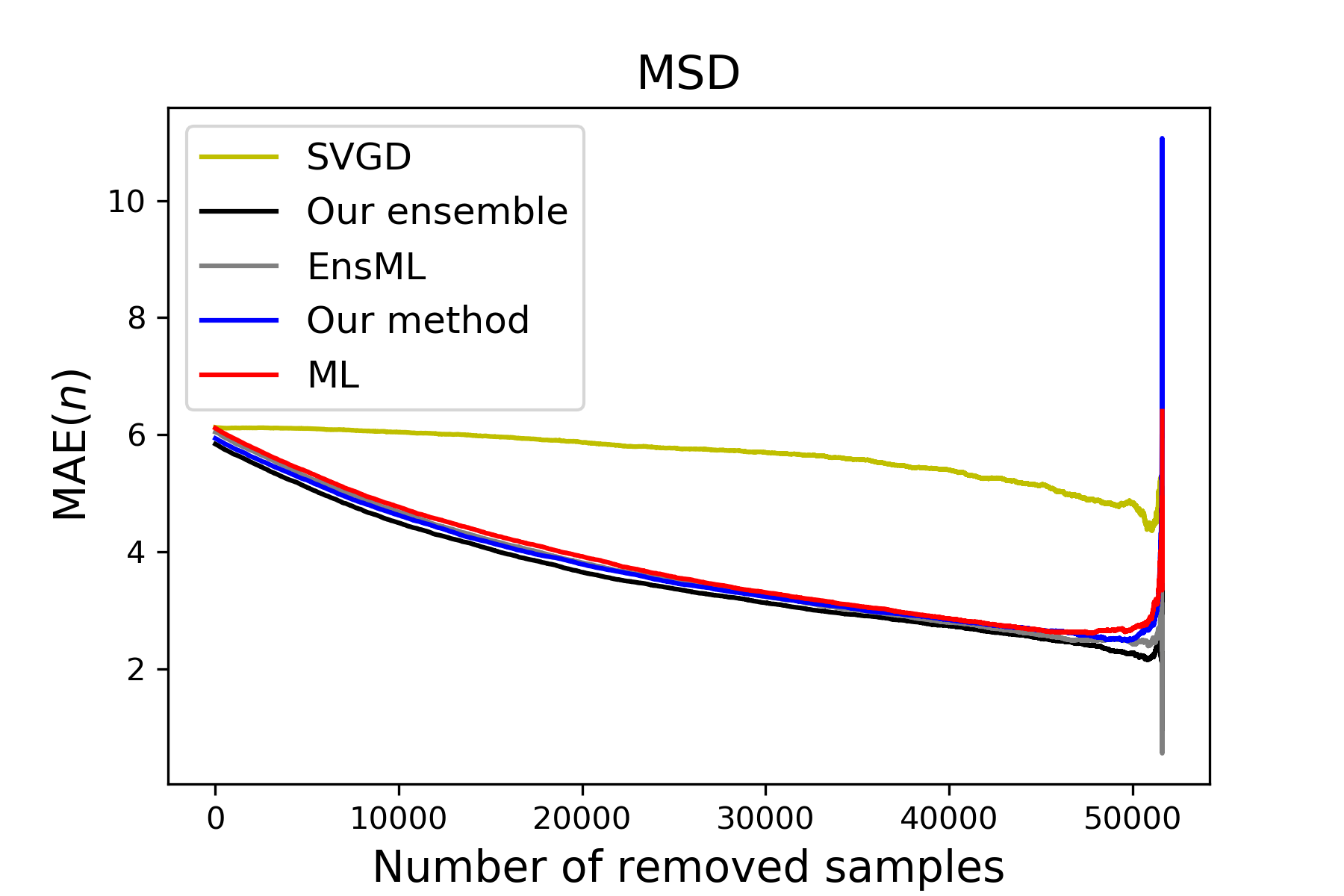}
	\end{minipage}
\caption{The curves ${\rm MAE}(n)$ for the different methods.
}\label{figAbsErrorRemovedSampleCurves}
\end{figure}

Tables~\ref{tableAbsErrorBostonConcrete} and~\ref{tableAbsErrorYachtKin8nmMSD}  show MAE and AUC for the different methods.
\begin{table}
\resizebox{\textwidth}{!}{%
\begin{tabular}{lccc}
{} & {\bf Boston} & {} \\
\toprule
{} & MAE & AUC  \\
\midrule
ML                        &    2.34$\pm$0.57 &         1.68$\pm$0.46  \\
SVGD                      &    2.14$\pm$0.53 &          1.61$\pm$0.33 \\
PBP                       &    {2.21$\pm$0.44} &         1.65$\pm$0.32  \\
Our method                &    {\bf 2.20$\pm$0.55} &         {\bf 1.39$\pm$0.39}  \\
\bottomrule
\bottomrule
EnsML                   &    {2.17$\pm$0.55} &         1.67$\pm$0.35  \\
Our ensemble            &    {\bf 2.06$\pm$0.57} &         {\bf 1.46$\pm$0.38}  \\
\end{tabular}
\
\begin{tabular}{lcc}
{\bf Concrete} & {}  \\
\toprule
 MAE & AUC  \\
\midrule
   4.09$\pm$0.49       &         2.69$\pm$0.38        \\
   4.82$\pm$0.70       &         3.29$\pm$0.59        \\
   4.21$\pm$0.64       &         3.59$\pm$0.60        \\
   4.05$\pm$0.48       &         2.57$\pm$0.38        \\
\bottomrule
\bottomrule
    3.63$\pm$0.61 &         2.32$\pm$0.39  \\
    {\bf 3.56$\pm$0.63} &         {\bf 2.19$\pm$0.40}  \\
\end{tabular}
\
\begin{tabular}{lccc}
 {\bf Power} & {} \\
\toprule
 MAE & AUC  \\
\midrule
    {3.08$\pm$0.12} &         2.72$\pm$0.15        \\
   3.23$\pm$0.12 &              3.22$\pm$0.18   &                             \\
    3.19$\pm$0.15 &         3.11$\pm$0.19        \\
    {\bf 3.09$\pm$0.13} &         {\bf 2.66$\pm$0.15}  \\
\bottomrule
\bottomrule
    {\bf 3.07$\pm$0.13} &         2.71$\pm$0.15  \\
    {\bf 3.07$\pm$0.13} &         {\bf 2.63$\pm$0.16}  \\
\end{tabular}
}
\caption{Values of MAE and AUC for the different methods on the Boston, Concrete, and Power data sets.}\label{tableAbsErrorBostonConcrete}
\end{table}
\begin{table}
\resizebox{\textwidth}{!}{%
\begin{tabular}{lccc}
{} & {\bf Yacht} & {} \\
\toprule
{} & MAE & AUC  \\
\midrule
ML                        &  0.50$\pm$0.25    &    0.18$\pm$0.06        \\
SVGD                      &  1.35$\pm$0.54    &    0.46$\pm$0.14         \\
PBP                       &  0.73$\pm$0.19    &    0.50$\pm$0.15        \\
Our method                &  0.57$\pm$0.20    &    0.22$\pm$0.07        \\
\bottomrule
\bottomrule
EnsML                   &    {\bf 0.33$\pm$0.14} &         {\bf 0.13$\pm$0.05}  \\
Our ensemble            &    0.42$\pm$0.17 &         0.15$\pm$0.04  \\
\end{tabular}
\
\begin{tabular}{lcc}
{\bf Kin8nm} & {}  \\
\toprule
 MAE & AUC  \\
\midrule
   0.069$\pm$0.005       &         0.053$\pm$0.003       \\
   0.116$\pm$0.006       &         0.089$\pm$0.005        \\
   0.076$\pm$0.004       &         0.064$\pm$0.004         \\
   0.072$\pm$0.00       &         0.052$\pm$0.003  \\
\bottomrule
\bottomrule
 {\bf 0.063$\pm$0.003} &        {\bf 0.048$\pm$0.003}  \\
 0.069$\pm$0.004 &         {\bf 0.048$\pm$0.002}  \\
\end{tabular}
\
\begin{tabular}{lcc}
{\bf MSD} & {}  \\
\toprule
MAE & AUC \\
\midrule
  6.11$\pm$NA     &         3.78$\pm$NA   \\
  6.12$\pm$NA             &    5.66$\pm$NA        \\
    NA             &         NA             \\
  {5.89$\pm$NA}   &       {  3.65$\pm$NA}  \\
\bottomrule
\bottomrule
  6.04$\pm$NA &     3.69$\pm$NA       \\
  {\bf 5.84$\pm$NA} &         {\bf 3.55$\pm$NA}  \\
\end{tabular}
}
\caption{Values of MAE and AUC for the different methods on the Yacht, Kin8nm, and MSD data sets.}\label{tableAbsErrorYachtKin8nmMSD}
\end{table}
Our ensemble achieves the best or not significantly different from the best MAE and AUC on all the data sets (except for MAE and AUC on Yacht and MAE on Kin8nm, where it is second best after the EnsML). Our non-ensemble networks are the best among the other non-ensemble methods, with the same exceptions as for the ensembles.
Figure~\ref{figAbsErrorRemovedSampleCurves} shows the curves ${\rm MAE}(n)$ for the different methods.

\bigskip

{\bf Acknowledgements.}  Both authors would like to thank the DFG project SFB 910.  The research of
the first author was also supported by the DFG Heisenberg Programme and the RUDN University Program 5-100.

\section{Conclusion}\label{secDiscussion}

We introduced a general approach to uncertainty quantification in artificial NNs, based on a specific joint loss~\eqref{eqJointLoss} for two NNs: one for the regression and another for the uncertainty quantification. We analyzed in detail how the functions $f$ and $g$ and the hyperparameter $\lambda$ in the loss affect the learning process. We showed that  the uncertainty quantifier provides an estimate of how certain the predictions are in terms of {\em any} regressor's loss and without the knowledge of the underlying distribution (whose form may even vary in different regions of the input space). Moreover, we explained how the presence of the uncertainty quantifier improves the predictions of the regressor and of NNs based on the classical likelihood maximization. We showed that the crucial role here is played by the hyperparameter $\lambda$, which allows for better fits on clean regions. Finally, we compared our NNs and their ensemble counterparts  with the other NN methods, using two measures: overall error (RMSE and MAE) and AUC. We showed that our approach typically yields the best or not significantly different from the best results.

It is also worth mentioning that one could fit a regressor first and then quantify the uncertainty of its predictions by training only the neural network $\cN_q$ with the loss~\eqref{eqJointLoss}. Now the function $\cL_r(y^i,y_r^i)$ need not coincide with the loss function that was used for training the regressor, but can represent {\em any} error whose local average we want to estimate by $\cN_q$. With this modification, one loses the benefit of the joined training of $\cN_r$ and $\cN_q$ that may improve regressor's predictions in the clean regions, but, on the other hand, one  can choose any type of regressor, not only a neural network. For the uncertainty quantification, one still has the full freedom in the choice of the functions $f$ and $g$, while the parameter $\lambda$ still has the same influence on the learning dynamics of $\cN_q$ as in Sec.~\ref{secSigmoidSoftReLambda}. 
In both settings, it would be interesting to analyze other choices of the functions $f$ and $g$ and to develop an automatic procedure that could choose an optimal $\lambda$ and properly adjust it during the learning process.

\appendix

\section{Hyperparameters}\label{appendixHyperparameters}

When we fit different methods on the real world data sets, we normalize them so that the input features and the targets
have zero mean and unit variance in the training set. We used minibatch~5 on Boston, Concrete, and Yacht, minibatch 10 on Power and Kin8nm, and minibatch 5000 on MSD. For the ML and for our method, we used the dropout regularization~\cite{Hinton12} during training, but not for the prediction, and we did not regularize the ensembles. We used Nesterov momentum (with momentum $0.9$) optimizer for fitting the ML, our method and the ensembles. We performed a grid search for the learning rate and the dropout rate for the non-ensemble methods. For the ensemble methods, we used the same learning rate as for the corresponding individual predictors. The parameters yielding the best AUC in case of RMSE are presented in Tables~\ref{tableHyperparametersBostonConcrete} and~\ref{tableHyperparametersYacht}, and the parameters yielding the best AUC in case of MAE are presented in Tables~\ref{tableHyperparametersBostonConcreteMAE} and~\ref{tableHyperparametersYachtMAE}. For the SVGD and PBP, we used the hyperparameters default settings in the authors' code\footnote{See https://github.com/DartML/Stein-Variational-Gradient-Descent for the SVGD and https://github.com/HIPS/Probabilistic-Backpropagation for the PBP.}.

\begin{table}
\resizebox{\textwidth}{!}{%
\begin{tabular}{lcccc}
{} & {\bf Boston} & {} & {} & {}\\
\toprule
{} & {$\lambda$} & \begin{tabular}{@{}c} {\bf Learning} \\ {\bf rate} \end{tabular} & {\bf Dropout} & \begin{tabular}{@{}c} {\bf Number}\\ {\bf of epochs} \end{tabular}  \\
\midrule
{\bf ML}                       &    {}  &         0.00004  & 0.4 & 500   \\
{\bf Our method}               & 0.1   &         0.0008  & 0.4 & 500  \\
\bottomrule
\end{tabular}
\
\begin{tabular}{cccc}
 {\bf Concrete} & {} & {} & {}\\
\toprule
 {$\lambda$} & \begin{tabular}{@{}c} {\bf Learning} \\ {\bf rate} \end{tabular} & {\bf Dropout} & \begin{tabular}{@{}c} {\bf Number}\\ {\bf of epochs} \end{tabular}  \\
\midrule
    {} &     0.0001  & 0.1 & 700  \\
    0.2 &     0.0002 & 0.15 & 700 \\
\bottomrule
\end{tabular}
\
\begin{tabular}{cccc}
 {\bf Power} & {} & {} & {}\\
\toprule
 {$\lambda$} & \begin{tabular}{@{}c} {\bf Learning} \\ {\bf rate} \end{tabular} & {\bf Dropout} & \begin{tabular}{@{}c} {\bf Number}\\ {\bf of epochs} \end{tabular}  \\
\midrule
     {} & 0.00005 & 0 & 150 \\
     0.2 & 0.0003  & 0 & 80  \\
\bottomrule
\end{tabular}
}
\caption{Hyperparameters for the ML, our method, and the respective ensembles optimising RMSE on the Boston, Concrete, and Power data sets.}\label{tableHyperparametersBostonConcrete}
\end{table}
\begin{table}
\resizebox{\textwidth}{!}{%
\begin{tabular}{lcccc}
{} & {\bf Yacht} & {} & {} & {}\\
\toprule
{} & {$\lambda$} & \begin{tabular}{@{}c} {\bf Learning} \\ {\bf rate} \end{tabular} & {\bf Dropout} & \begin{tabular}{@{}c} {\bf Number}\\ {\bf of epochs} \end{tabular}  \\
\midrule
{\bf ML}                       &    {} &         0.0001 & 0.1 & 2000   \\
{\bf Our method}               &    0.2     &         0.0004 & 0 & 500 \\
\bottomrule
\end{tabular}
\
\begin{tabular}{cccc}
 {\bf Kin8nm} & {} & {} & {}\\
\toprule
 {$\lambda$} & \begin{tabular}{@{}c} {\bf Learning} \\ {\bf rate} \end{tabular} & {\bf Dropout} & \begin{tabular}{@{}c} {\bf Number}\\ {\bf of epochs} \end{tabular}  \\
\midrule
    {} & 0.00005 & 0 & 200  \\
    0.5 & 0.0002 & 0 & 300 \\
\bottomrule
\end{tabular}
\
\begin{tabular}{cccc}
 {\bf MSD} & {} & {} & {}\\
\toprule
 {$\lambda$} & \begin{tabular}{@{}c} {\bf Learning} \\ {\bf rate} \end{tabular} & {\bf Dropout} & \begin{tabular}{@{}c} {\bf Number}\\ {\bf of epochs} \end{tabular}  \\
\midrule
     {} & 0.005 & 0.1 & 50  \\
     0.1 & 0.2 & 0.1 & 50 \\
\bottomrule
\end{tabular}
}
\caption{Hyperparameters for the ML, our method, and the respective ensembles optimising RMSE on the Yacht, Kin8nm, and MSD data sets.}\label{tableHyperparametersYacht}
\end{table}

\begin{table}
\resizebox{\textwidth}{!}{%
\begin{tabular}{lcccc}
{} & {\bf Boston} & {} & {} & {}\\
\toprule
{} & {$\lambda$} & \begin{tabular}{@{}c} {\bf Learning} \\ {\bf rate} \end{tabular} & {\bf Dropout} & \begin{tabular}{@{}c} {\bf Number}\\ {\bf of epochs} \end{tabular}  \\
\midrule
{\bf ML}                       &    {}  &         0.00003  & 0.4 & 600   \\
{\bf Our method}               & 0.2   &         0.0002 & 0.4 & 2000  \\
\bottomrule
\end{tabular}
\
\begin{tabular}{cccc}
 {\bf Concrete} & {} & {} & {}\\
\toprule
 {$\lambda$} & \begin{tabular}{@{}c} {\bf Learning} \\ {\bf rate} \end{tabular} & {\bf Dropout} & \begin{tabular}{@{}c} {\bf Number}\\ {\bf of epochs} \end{tabular}  \\
\midrule
    {} &     0.00003  & 0.1 & 800  \\
    0.2 &     0.0003 & 0.1 & 600 \\
\bottomrule
\end{tabular}
\
\begin{tabular}{cccc}
 {\bf Power} & {} & {} & {}\\
\toprule
 {$\lambda$} & \begin{tabular}{@{}c} {\bf Learning} \\ {\bf rate} \end{tabular} & {\bf Dropout} & \begin{tabular}{@{}c} {\bf Number}\\ {\bf of epochs} \end{tabular}  \\
\midrule
     {} & 0.00005 & 0 & 150 \\
     1 & 0.0001  & 0 & 200  \\
\bottomrule
\end{tabular}
}
\caption{Hyperparameters for the ML, our method, and the respective ensembles optimising MAE on the Boston, Concrete, and Power data sets.}\label{tableHyperparametersBostonConcreteMAE}
\end{table}
\begin{table}
\resizebox{\textwidth}{!}{%
\begin{tabular}{lcccc}
{} & {\bf Yacht} & {} & {} & {}\\
\toprule
{} & {$\lambda$} & \begin{tabular}{@{}c} {\bf Learning} \\ {\bf rate} \end{tabular} & {\bf Dropout} & \begin{tabular}{@{}c} {\bf Number}\\ {\bf of epochs} \end{tabular}  \\
\midrule
{\bf ML}                       &    {} &         0.0001 & 0.1 & 1000   \\
{\bf Our method}               &    0.2     &         0.0004 & 0 & 500 \\
\bottomrule
\end{tabular}
\
\begin{tabular}{cccc}
 {\bf Kin8nm} & {} & {} & {}\\
\toprule
 {$\lambda$} & \begin{tabular}{@{}c} {\bf Learning} \\ {\bf rate} \end{tabular} & {\bf Dropout} & \begin{tabular}{@{}c} {\bf Number}\\ {\bf of epochs} \end{tabular}  \\
\midrule
    {} & 0.00005 & 0 & 400  \\
    0.2 & 0.0002 & 0 & 400 \\
\bottomrule
\end{tabular}
\
\begin{tabular}{cccc}
 {\bf MSD} & {} & {} & {}\\
\toprule
 {$\lambda$} & \begin{tabular}{@{}c} {\bf Learning} \\ {\bf rate} \end{tabular} & {\bf Dropout} & \begin{tabular}{@{}c} {\bf Number}\\ {\bf of epochs} \end{tabular}  \\
\midrule
     {} & 0.01 & 0.1 & 50  \\
     0.5 & 0.1 & 0.1 & 70 \\
\bottomrule
\end{tabular}
}
\caption{Hyperparameters for the ML, our method, and the respective ensembles optimising MAE on the Yacht, Kin8nm, and MSD data sets.}\label{tableHyperparametersYachtMAE}
\end{table}

\end{document}